\documentclass[10pt,journal,compsoc]{IEEEtran}

\ifCLASSOPTIONcompsoc
  \usepackage[nocompress]{cite}
\else
  \usepackage{cite}
\fi

\usepackage{times}
\usepackage{epsfig}
\usepackage{graphicx}
\usepackage{amsmath}
\usepackage{amssymb}
\usepackage{paralist}
\usepackage{caption}
\usepackage{subcaption}
\usepackage{color,soul}
\usepackage[dvipsnames]{xcolor,colortbl}
\usepackage{multirow}
\usepackage{siunitx}
\usepackage{pifont}
\usepackage{dsfont}
\usepackage{mathtools}
\usepackage{bbm}
\usepackage{adjustbox}
\usepackage{float}
\usepackage[pagebackref=true,breaklinks=true,colorlinks,bookmarks=false]{hyperref}

\graphicspath{ {./fig/} }

\renewcommand{\vec}[1]{\mathbf{#1}}
\newcommand{\set}[1]{\mathcal{#1}}

\DeclareMathOperator*{\argmax}{arg\,max}
\DeclareMathOperator*{\argmin}{arg\,min}
\newcommand{\ie}{i.e.\ }
\newcommand{\eg}{e.g.\ }
\newcommand{\wrt}{w.r.t.\ }
\newcommand*{\tran}{^{\mkern-1.5mu\mathsf{T}}}

\newcommand*{\inverse}{^{\mkern-1.5mu{-1}}}
\newcommand*{\pinv}{^{\mkern-1.5mu{+}}}

\definecolor{nicegreen}{HTML}{99C000}
\definecolor{nicegreen2}{HTML}{81d41a}
\definecolor{niceyellow}{HTML}{FDCA00}
\definecolor{niceorange}{HTML}{f5a300}
\definecolor{niceblue}{HTML}{0083cc}
\definecolor{nicepurple}{HTML}{a60084}

\definecolor{c_red}{HTML}{e6194b}
\definecolor{c_blue}{HTML}{4363d8}
\definecolor{c_green}{HTML}{aaffc3}
\definecolor{c_purple}{HTML}{911eb4}
\definecolor{c_cyan}{HTML}{46f0f0}
\definecolor{c_orange}{HTML}{f58231}

\newcommand{\cmark}{\ding{51}}%
\newcommand{\xmark}{\ding{55}}%

\newcommand{\new}[1]{#1}

\newcommand{\best}[1]{{\colorbox[HTML]{CCDDAA}{$\mathbf{#1}$}}}
\newcommand{\second}[1]{{\colorbox[HTML]{EEEEBB}{\underline{$#1$}}}}
\newcommand{\third}[1]{{\colorbox[HTML]{cceeff}{$\mathit{#1}$}}}

\renewcommand{\paragraph}[1]{\vspace{.25em}\noindent\textbf{#1.}}

\newsavebox{\twosubbox}

\newsavebox{\subboxB}

\definecolor{ours}{HTML}{e1e0ff}
\definecolor{baseline}{HTML}{fff7e0}
\newcolumntype{o}{>{\columncolor{ours}[\dimexpr\tabcolsep+0.1pt\relax][\dimexpr\tabcolsep+0.1pt\relax]}c}
\newcolumntype{b}{>{\columncolor{baseline}[\dimexpr\tabcolsep+0.1pt\relax][\dimexpr\tabcolsep+0.1pt\relax]}c}

\begin{document}

\title{Robust Shape Fitting for 3D Scene Abstraction}

\author{Florian Kluger,
        Eric Brachmann,
        Michael Ying Yang,
        Bodo Rosenhahn
\IEEEcompsocitemizethanks{\IEEEcompsocthanksitem F. Kluger and B. Rosenhahn are with the Institute of Information Processing, Leibniz University Hannover, Germany. kluger@tnt.uni-hannover.de\protect
\IEEEcompsocthanksitem E. Brachmann is with Niantic, Inc.
\IEEEcompsocthanksitem M.Y. Yang is with the Scene Understanding Group, University of Bath, UK.
}
\thanks{
© 2024 IEEE. Personal use of this material is permitted. Permission
from IEEE must be obtained for all other uses, in any current or future
media, including reprinting/republishing this material for advertising or
promotional purposes, creating new collective works, for resale or
redistribution to servers or lists, or reuse of any copyrighted
component of this work in other works.}
}

\IEEEtitleabstractindextext{%
\begin{abstract}
Humans perceive and construct the world as an arrangement of simple parametric models. 
In particular, we can often describe man-made environments using volumetric primitives such as cuboids or cylinders.
Inferring these primitives is important for attaining high-level, abstract scene descriptions. 
Previous approaches for primitive-based abstraction estimate shape parameters directly and are only able to reproduce simple objects.
In contrast, we propose a robust estimator for primitive fitting, which meaningfully abstracts complex real-world environments using cuboids. 
A RANSAC estimator guided by a neural network fits these primitives to a depth map. 
We condition the network on previously detected parts of the scene, parsing it one-by-one. 
To obtain cuboids from single RGB images, we additionally optimise a depth estimation CNN end-to-end.
Naively minimising point-to-primitive distances leads to large or spurious cuboids occluding parts of the scene. 
We thus propose an improved occlusion-aware distance metric correctly handling opaque scenes.
Furthermore, we present a neural network based cuboid solver which provides more parsimonious scene abstractions while also reducing inference time.
The proposed algorithm does not require labour-intensive labels, such as cuboid annotations, for training.
Results on the NYU Depth v2 dataset demonstrate that the proposed algorithm successfully abstracts cluttered real-world 3D scene layouts.
\end{abstract}

}

\maketitle

\IEEEdisplaynontitleabstractindextext

%
\IEEEpeerreviewmaketitle

\begin{figure}
		\centering
		\begin{subfigure}[t]{\linewidth}
			\centering
   
			\includegraphics[width=0.49\linewidth]{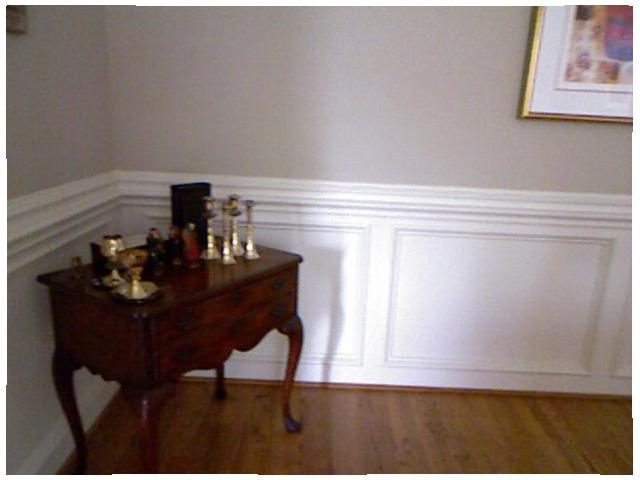}
			\includegraphics[width=0.49\linewidth]{fig/teaser/1171_original.jpg}
			\caption{Input images
                \vspace{1em}}
		\end{subfigure}
		\begin{subfigure}[t]{\linewidth}
			\centering
			\includegraphics[width=0.49\linewidth]{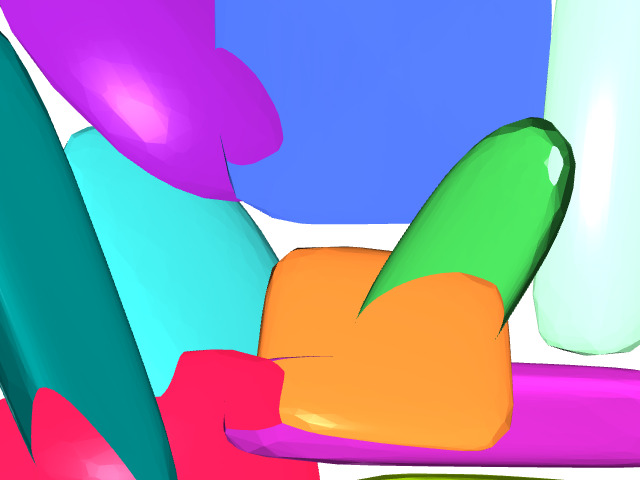}
			\includegraphics[width=0.49\linewidth]{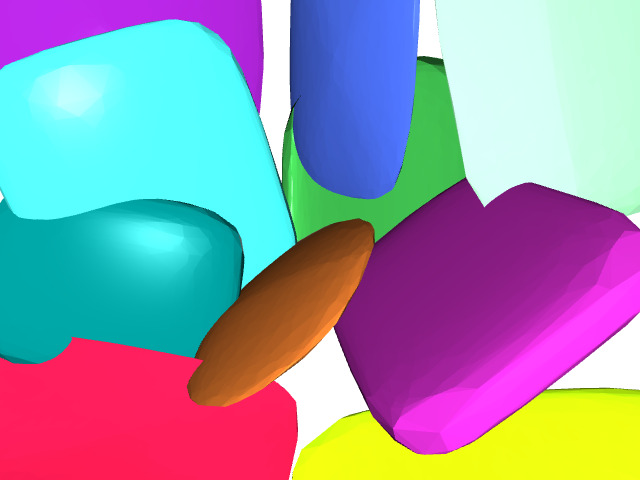}
			\caption{Superquadrics Revisited~\cite{paschalidou2019superquadrics}
                \vspace{1em}}
		\end{subfigure}
		\begin{subfigure}[t]{\linewidth}
			\centering
			\includegraphics[width=0.49\linewidth]{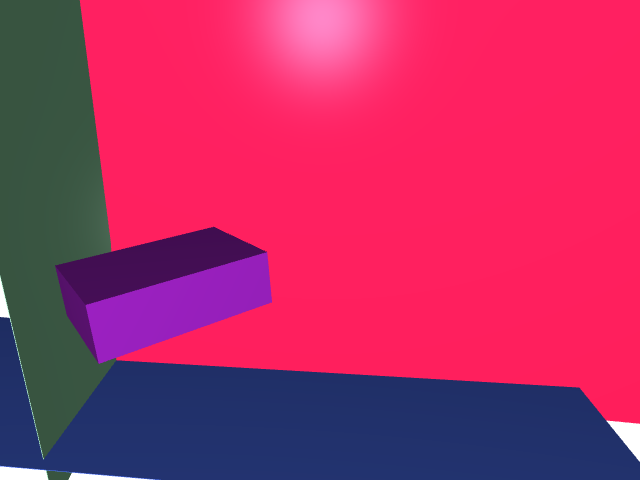}
			\includegraphics[width=0.49\linewidth]{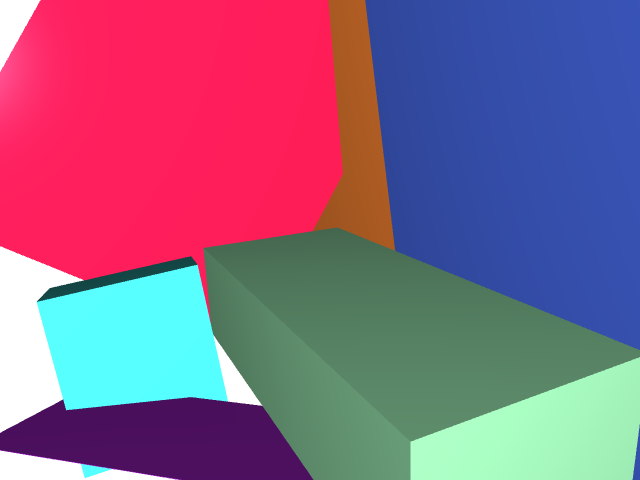}
			\caption{Our approach}
		\end{subfigure}
		\caption{\textbf{Primitive-based Scene Abstractions:} We parse images of real-world scenes (a) and generate abstractions of their 3D structure using cuboids (c). Our method captures scene structure more accurately than previous work~\cite{paschalidou2019superquadrics} using superquadrics (b). 
  }
		\label{fig:teaser}
\end{figure}

\IEEEraisesectionheading{\section{Introduction}}
\label{sec:intro}
\IEEEPARstart{H}{umans} tend to create using simple geometric forms. 
For instance, a house is made from bricks and squared timber, and a book is a cuboidal assembly of rectangles.
Thus humans also appear to visually abstract environments by decomposing them into arrangements of cuboids, cylinders, ellipsoids and other simple volumetric primitives~\cite{biederman1987recognition}. 
Such an abstraction of the 3D world is also very useful for machines with visual perception. 
Scene representation based on geometric shape primitives has been an active topic since the very beginning of Computer Vision. 
In 1963, \emph{Blocks World}~\cite{roberts1963machine} was one of the earliest approaches for qualitative 3D shape recovery from images using generic shape primitives.
In recent years, with rapid advances in deep learning, 
high-quality 3D reconstruction from single images has become feasible. 
Most approaches recover 3D information such as depth~\cite{Eigen2014:DepthMap}, meshes~\cite{WangZLFLJ18} {and implicit surfaces}~\cite{mescheder2019occupancy} from RGB images.
{These methods are capable of representing surfaces of 3D structures with high fidelity,
but the resulting detailed models contain redundant information if the underlying structures are composed of simpler geometric shapes.
In contrast, decomposing objects and scenes into their constituent primitives is not only more parsimonious, but essential for applications such as CAD based modelling~\cite{le2021cpfn} and scene layout estimation~\cite{baruch2021arkitscenes}.}
Few works consider such abstract 3D shape decompositions, \eg using cuboids~\cite{tulsiani2017learning} or superquadrics~\cite{paschalidou2019superquadrics, Paschalidou2020:Decomposition}.
These 3D shape parsers work well for isolated objects, but do not generalise to complex real-world scenes (cf.~Fig.~\ref{fig:teaser}). 

Robust model fitting algorithms such as RANSAC~\cite{Fischler1981:RANSAC} and its many  derivatives~\cite{barath2018graph, Chin2014:AccHypGen, PurChi2014a} have been used to fit low-dimensional parametric models, such as plane homographies, fundamental matrices or geometric primitives, to real-world noisy data.
Trainable variants of RANSAC~\cite{Brachmann2017:DSAC, Brachmann2019:NGRANSAC, KluBra2020} use a neural network to predict sampling weights from data. 
{They achieve state-of-the-art performance while requiring fewer samples.}

Leveraging advances in the fields of 
single image 3D reconstruction and robust multi-model fitting, 
we present a novel approach for robustly parsing real-world scenes using 3D shape primitives, such as cuboids.
A trainable RANSAC estimator fits these primitives to 3D features, such as a depth map. 
We build upon the estimator proposed in \cite{KluBra2020}, and extend it by predicting multiple sets of RANSAC sampling weights concurrently.
This enables our method to distinguish between different structures in a scene more easily.
We obtain 3D features from a single RGB image using a CNN, and show how to optimise this CNN in an end-to-end manner.
Our training objective is based on geometrical consistency with readily available 3D sensory data.

During primitive fitting, a naive maximisation of inlier counts considering point-to-primitive distances causes the algorithm to detect few but excessively large models. 
We argue that this is due to parts of a primitive surface correctly representing some parts of a scene, while other parts of the same primitive wrongly occlude other parts of the scene.
Thus, points should not be assigned to primitive surfaces which cannot be seen by the camera due to occlusion or self-occlusion. 
We therefore propose an occlusion-aware distance and a corresponding occlusion-aware inlier count.  

As no closed-form solution exists to calculate cuboid parameters from a set of surface points, {we propose two different \new{\emph{cuboid solver}} approaches to infer them}.
{The first approach is based on iterative numerical optimisation and minimises a cost function in order to find cuboid parameters which optimally fit a minimal set of points.
We call this method the \emph{numerical solver}.}
Backpropagation through this optimisation is numerically unstable and computationally costly.
We therefore analytically derive the gradient of primitive parameters \wrt the features used to compute them.
Our  gradient computation allows for end-to-end training without backpropagation through the cuboid solver itself.
{The second approach, which we call the \emph{neural solver}, uses a neural network to predict cuboid parameters from a minimal set of points in a single forward pass.
It is significantly faster than the numerical solver, fully differentiable, and still provides high-quality results.}

We demonstrate the efficacy of our method on the real-world NYU Depth v2~\cite{silberman2012indoor} and the Synthetic Metropolis Homography~\cite{kluger2024parsac} datasets. 
{For quantitative evaluation, we employ a diverse set of performance metrics measuring reconstruction accuracy, abstraction parsimony, and overall scene coverage. }
{An extensive set of ablation studies provides detailed empirical analyses of our method, of its individual components and their hyperparameters.}

{
This article is an extension of our previous work on primitive-based 3D scene abstraction published in~\cite{kluger2021cuboids}.
In summary, the contributions presented in~\cite{kluger2021cuboids}\footnote{Source code is available at: \url{https://github.com/fkluger/cuboids_revisited}} are:}
\begin{itemize}
\item A 3D scene parser which can process more complex real-word scenes than previous works on 3D scene abstraction.
\item An occlusion-aware distance metric, necessary for inferring volumetric primitives from depth maps.
\item Analytical derivation of the gradient of cuboid parameters \wrt input points in order to implement end-to-end training with a numerical cuboid solver. 
\item \new{Training supervised by depth maps, which does not require labour-intensive labels, such as cuboid or object annotations.}
\end{itemize}

\vspace{.75em}
{
\noindent In this article, we present the following additional contributions:
\begin{itemize}
\item A neural network based cuboid solver, which directly regresses cuboid parameters from a set of 3D points, reducing inference time by a factor of eight.
\item An improved occlusion-aware inlier metric which avoids strong occlusions more effectively.
\item Additional evaluation metrics for a \new{more comprehensive} assessment of scene abstraction quality.
\item An extensive empirical analysis of our method via several ablation studies.
\end{itemize}
}

\section{Related Work}
\label{sec:related}
{
In Computer Vision literature, a large body of works deals with acquisition, processing and representation of 3D models of objects and spatial environments. 
In the following, we review a number of works and discuss their similarities and differences \new{with} our proposed method in the following aspects: type of input data, 3D shape representation, and ground truth data requirements.
}

\vspace{-0.1em}
\subsection{3D Bounding Box Regression.}
For predefined object classes, 3D bounding box regression is a well investigated topic in 3D object detection for RGB images~\cite{Ren2016:3dObjectDetection, Dwibedi2016:Beyond2dBBs,Kehl2018:3dDetection,Mousavian2017:3dBB},
RGB-D images~\cite{Qi2018:FrustumPNets, Zhang2017:DeepContext}, and 3D point clouds~\cite{Shi2019:PRCNN,Thomas2019ICCV}. 
These works not only detect bounding boxes but also classify objects. 
{However, these bounding boxes are not intended to approximate the shapes of objects, but instead describe pose and extent of arbitrarily shaped objects.}
During training, target 3D boxes as well as category labels are needed. 
In contrast, our proposed method does not require such annotations.

\vspace{-0.1em}
\subsection{Monocular Depth Estimation.}
\new{Monocular depth estimation has been an active topic of research for around two decades~\cite{saxena2005learning}, but has become significantly more accurate with advances in deep learning over recent years. 
We categorise learning based depth estimation methods based on whether they utilise supervised learning~\cite{Eigen2014:DepthMap, Liu2015:Depth, lee2019big}, }
semi-supervised learning~\cite{Kuznietsov2017:Semi}, unsupervised learning~\cite{Garg0216:UnsupervisedDepth, Mahjourian2018:Unsupervised, Godard2017:DepthEstimation, Yin2018:GeoNet} or self-supervised learning~\cite{Godard2019:SelfSupervised, Ma2019:SelfSupervised,yuan2022new,bhat2021adabins}.
These methods predict dense 3D information instead of the more parsimonious primitive based descriptions we are interested in.
However, we show how to leverage monocular depth estimation to this end.

\vspace{-0.1em}
\subsection{Single Image 3D Reconstruction.}
{
\paragraph{Planar Surface Reconstruction}
The surfaces of scenes in human-made environments can often be well approximated by decomposing them into a set of planes.
Early works~\cite{criminisi2000single, delage2007automatic, barinova2008fast} rely on geometric image features, such as vanishing points~\cite{kluger2017deep} or horizon lines~\cite{kluger2020temporally}, in order to reconstruct planar surfaces from single images.
More recently, the authors of~\cite{Liu2018:PlaneNet, liu2019planercnn, yang2018recovering} directly estimate 3D planes from RGB images using convolutional neural networks.
}
Like depth estimation, however, these methods can only describe the visible surfaces of a scene and do not {reason about its} volumetric characteristics.

{
\paragraph{Single Object Reconstruction}
}
{
Other methods predict the full 3D shape of objects, either implicitly as a signed distance function~\cite{xu2019disn} or an occupancy function~\cite{mescheder2019occupancy,bian2022ray,xian2022any}, or explicitly as a voxel volume~\cite{xie2019pix2vox} or a mesh~\cite{Pan2019:Mesh,groueix2018papier,WangZLFLJ18,shi2021geometric}.
GSIR~\cite{wang2020gsir} combines shape reconstruction with cuboid based shape abstraction. }
{However, these works are limited to single objects in isolation and cannot be applied directly to more complex scenes. }
{Furthermore, they} require ground-truth object meshes for training, which are costly to obtain.

{
\paragraph{Scene Reconstruction}
}
{
The authors of~\cite{gkioxari2019mesh} combine the Mask R-CNN~\cite{he2017mask} object detector with a voxel-based single object reconstruction branch in order to extract 3D models of multiple objects from images of real-world scenes.
These objects, however, are reconstructed with a canonical pose, so their position within the scene is not recovered.
Other works additionally predict object poses~\cite{weng2021holistic,zhang2021holistic,Nie2020:Total3dUnderstanding,liu2022towards,liu2021voxel}, room layout~\cite{Nie2020:Total3dUnderstanding,weng2021holistic,zhang2021holistic}, background surfaces~\cite{liu2022towards} or humans~\cite{weng2021holistic} in order to attain a more holistic reconstruction of a scene. 
As these methods are trained in a supervised manner, they require additional ground truth annotations.
In~\cite{gkioxari2022learning}, the authors forego the need for any 3D ground truth by relying on photometric consistency of multiple views of a scene, utilising a differentiable rendering technique during training. 
However, they still require 2D object bounding box annotations.
In contrast, our approach \new{relies on} ground truth depth maps for training.
\new{These can be acquired, for example, with commercially available RGB-D cameras. Additional, labour intensive manual labelling is not necessary.}
\new{Apart from that,} the mutual goal of single object and scene reconstruction methods is to attain high-fidelity 3D models.
In this work, however, we focus on parsimonious primitive-based shape descriptions instead.

}

{
\subsection{3D Shape Decomposition}
}
{
The decomposition of objects and scenes into smaller parts is a field of research with a variety of applications, such as CAD-based modelling~\cite{koch2019abc}, 3D reconstruction~\cite{Genova_2020_CVPR} and layout estimation~\cite{baruch2021arkitscenes}.
One branch of works focuses on reconstructing objects with high fidelity using a set of either implicit~\cite{genova2019learning,Genova_2020_CVPR} or explicit~\cite{deng2020cvxnet, paschalidou2021neural} shapes that require relatively few parameters but can assume nearly arbitrary forms.
They can infer shapes either from either 3D inputs, such as point clouds, or RGB images.
These works, however, focus on single objects in isolation and require full 3D ground truth for training.
As they put a strong emphasis on reconstruction accuracy, these methods also serve a different purpose than the primitive based abstractions we are interested in.

Another quite recent set of works~\cite{ganin2021computer,wu2021deepcad,uy2022point2cyl} aims at decomposing complex shapes into simple geometric primitives.
They purely operate on 3D inputs of single objects, though, with CAD based modelling as their target application.
Since these methods are trained in a fully-supervised fashion, they require detailed annotations of the ground truth primitives.

More closely related to ours are the methods of~\cite{Jiang2013:Linear,Lin2013:Holistic,Xiao2012:Localizing}.
The authors decompose indoor scenes into sets of cuboids, using either RGB images~\cite{Xiao2012:Localizing} or RGB and depth~\cite{Lin2013:Holistic, Jiang2013:Linear} as input.
However, unlike our approach, they also require ground truth cuboid annotations for training.
}
The works most related to ours are~\cite{tulsiani2017learning,paschalidou2019superquadrics,Paschalidou2020:Decomposition}.
In~\cite{tulsiani2017learning}, the authors propose a method for 3D shape abstraction using cuboids. 
Based on a neural network, it directly regresses cuboid parameters from either an RGB image or a 3D input, and is trained in an unsupervised fashion.
Similarly, the authors of~\cite{paschalidou2019superquadrics} decompose objects into sets of superquadric surfaces.
They extend this to hierarchical sets of superquadrics in~\cite{Paschalidou2020:Decomposition}.
These approaches are only evaluated on relatively simple shapes, such as those in ShapeNet~\cite{Chang2015:ShapeNet}. 
While they yield good results for such objects, they are unable to predict reasonable primitive decompositions for complex real-world scenes  (cf.~Sec.~\ref{sec:exp}).

\subsection{Robust Multi-Model Fitting}
{
Robust fitting of parametric models, given data contaminated with noise and outliers, is a key problem in Computer Vision. 
RANSAC~\cite{Fischler1981:RANSAC} -- the most commonly known approach --  generates model hypotheses from randomly sampled minimal sets of observations, and selects the hypothesis with the largest number of inliers, \ie observations whose distance to the hypothesis is below an inlier threshold.
This method has proven effective for finding single instances of parametric models.
If multiple models are apparent in our data, we need to apply robust \emph{multi-model} fitting techniques instead.
Sequential RANSAC~\cite{vincent2001seqransac} fits multiple models sequentially by removing inliers from the set of all observations after each iteration.
Other methods, such as PEARL~\cite{isack2012energy}, instead fit multiple models simultaneously~\cite{barath2016multi, pham2014interacting, amayo2018geometric, barath2018multi}. 
They optimise an energy-based functional which is initialised via stochastic sampling.
Progressive-X~\cite{barath2019progressive} is a hybrid approach, which guides hypothesis generation via intermediate estimates by interleaving sampling and optimisation steps. 
J-Linkage~\cite{toldo2008robust} and T-Linkage~\cite{magri2014t} cluster observations agglomeratively using preference analysis~\cite{zhang2006nonparametric,chin2009robust}.
MCT~\cite{magri2019fitting} is a multi-class generalisation of T-Linkage, while RPA~\cite{magri2015robust} uses spectral instead of agglomerative clustering. 
Progressive-X+~\cite{barath2021progressive} combines Progressive-X with preference analysis and allows observations to be assigned to multiple models. It is faster and more accurate than its predecessor.
CONSAC~\cite{KluBra2020} is the first approach to utilise deep learning for robust multi-model fitting.
Using a neural network to guide the hypothesis sampling, it discovers model instances sequentially and predicts new sampling weights after each step.
}
In this work, we extend~\cite{KluBra2020} for robustly parsing real-world scenes using 3D shape primitives.
Unlike~\cite{KluBra2020}, we not just learn the parameters of the sampling weight from data, but train it conjointly with a depth estimation network {and a neural network based cuboid solver} in an end-to-end manner. 

\begin{figure*}	
\centering
\includegraphics[width=0.99\linewidth]{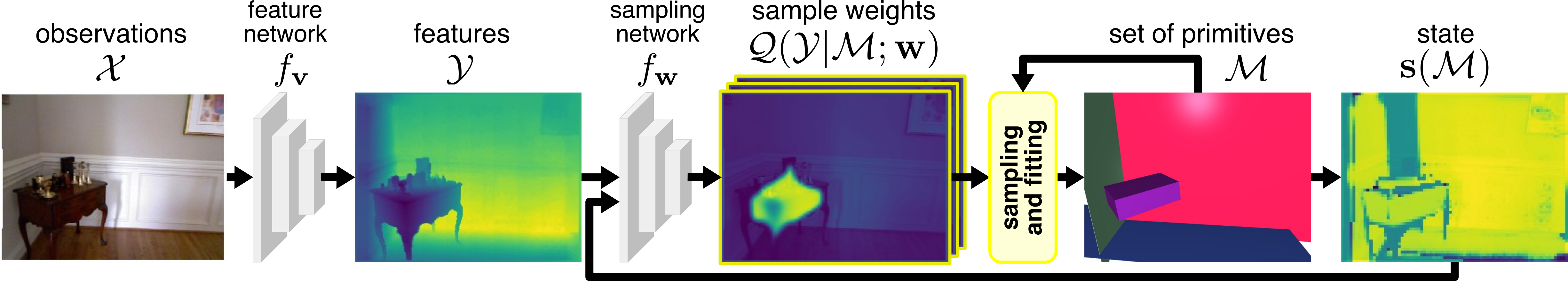}
\caption{ \textbf{Overview:} 
Given observations $\set{X}$ (RGB image), we predict 3D features $\set{Y}$ (depth map) using a neural network with parameters $\vec{v}$.
Conditioned on a state $\vec{s}$, a second neural network with parameters $\vec{w}$ predicts sampling weights $p(\vec{y}|\vec{s}; \vec{w}) \in \set{Q}$ for each feature $\vec{y} \in \set{Y}$.
Using these weights, a RANSAC-based estimator samples minimal sets of features, and generates primitive (cuboid) hypotheses $\set{H}$.
It selects the best hypothesis $\vec{\hat{h}} \in \set{H}$ and appends it to the set of previously recovered primitives $\set{M}$.
We update the state $\vec{s}$ based on $\set{M}$ and repeat the process in order to recover all primitives step-by-step.
}
\label{fig:system}
\end{figure*} 

\section{Method}
\label{sec:method}
We formulate the problem of abstracting a target scene $\set{Z}$ by fitting a set of shape primitives $\set{M}$ to features $\set{Y}$, which may either be provided directly, or extracted from observations $\set{X}$.
Here, $\set{Z}$ represents the ground truth depth or its corresponding point cloud, and $\set{X}$ is an RGB image of the scene aligned with $\set{Z}$.
{Each primitive $\vec{h} \in \set{M}$ is a cuboid with variable size and pose.}
Features $\set{Y}$ are either equal to $\set{Z}$ {if a depth map is available}, or estimated via a function $\set{Y}=f_\vec{v}(\set{X})$.
{
When only an RGB image $\set{X}$ is available, the function $f_{\vec{v}}$ is a depth estimator which gives us the features $\set{Y}$ in the form of a pixel-wise depth map.
The depth estimator is realised as a convolutional neural network with parameters $\vec{v}$.
We convert $\set{Y}$ into a point cloud via backprojection using known camera intrinsics $\vec{K}$.
}
{
Fig.~\ref{fig:system} gives an overview of the complete pipeline with examples for $\set{X}$, $\set{Y}$ and $\set{M}$.}

In order to fit primitives $\set{M}$ to features $\set{Y}$, we build upon the robust multi-model estimator of Kluger et al.~\cite{KluBra2020}. 
This estimator predicts sampling weights $\vec{p} = f_{\vec{w}}(\set{Y}, \vec{s})$ from $\set{Y}$ and a state $\vec{s}$ via a neural network with parameters $\vec{w}$, which are learnt from data.
\new{The state $\vec{s}$ encodes primitives which have been detected in previous iterations.}
The estimator then samples minimal sets of features from $\set{Y}$ according to $\vec{p}$, and fits primitive hypotheses $\set{H}$ via a minimal solver $f_h$.
From these hypotheses, it selects the best primitive $\vec{\hat{h}} \in \set{H}$ according to an inlier criterion, and adds it to the current set of primitives $\set{M}$. 
Based on $\set{M}$, it then updates the state $\vec{s}$ and predicts new sampling weights $\vec{p}$ in order to sample and select the next primitive.
This process, as visualised in Fig.~\ref{fig:system}, is repeated until \new{it reaches a stopping criterion based on the number of added inlier points}.

The scenes we are dealing with in this work have been captured with an RGB-D camera.
We therefore do not have full 3D shapes available as ground-truth for the scenes.
Instead, we only have 2.5D information, \ie 3D information for visible parts of the scene, and no information about the occluded parts.
If not taken into account, this fact can lead to spurious, oversized or ill-fitting primitives being selected, as visualised in Fig.~\ref{fig:occlusion}.
For this reason we present occlusion-aware distance and inlier metrics.

Unlike~\cite{KluBra2020}, we learn the parameters of the sampling weight predictor $f_{\vec{w}}$ jointly with the feature extractor $f_{\vec{v}}$ in an end-to-end manner. 
We show how to achieve this {even when} backpropagation through a minimal (cuboid) solver $f_h$ {directly is unreasonable}.
{We also present a new neural network based cuboid solver which is inherently fully differentiable and provides a significant speed-up.}
In addition, we generalise $f_{\vec{w}}$ so that it predicts multiple sets of sampling weights $\set{Q} = \{ \vec{p}_1, \dots, \vec{p}_Q \} $ at once, which enables it to distinguish between different primitive instances more effectively.

\subsection{Cuboid Parametrisation}
A cuboid is described by its {shape} $(a_x, a_y, a_z)$ and {pose} $(\vec{R}, \vec{t})$. 
The shape corresponds to width, height and length in a cuboid-centric coordinate system, while the pose translates the latter into a world coordinate system.
We represent rotation $\vec{R}$ in axis-angle notation $\vec{r} = \theta \vec{u}$.
Each cuboid thus has nine degrees of freedom.

\paragraph{Point-to-Cuboid Distance}
\label{subsubsec:point_to_cuboid_distance}
When computing the distance between point $\vec{y} = (x, y, z)\tran$ and cuboid $\vec{h} = (a_x, a_y, a_z, \vec{R}, \vec{t})$, we first translate $\vec{y}$ into the cuboid-centric coordinate frame:
\begin{equation}
  \vec{\hat{y}} = \vec{R}\tran(\vec{y} - \vec{t}) \, .
\end{equation}
We then compute its squared distance to the cuboid surface:
\begin{align}
d( \vec{h}, \vec{y})^2 =&  
 \max(\min_{c \in \{x, y, z\}}(a_c-|\hat{c}|), 0)^2 + \nonumber  \\
 & \sum_{c \in \{x, y, z\}} \max(|\hat{c}|-a_c, 0)^2 \, .
 \label{eq:point_cuboid_dist}
\end{align}
Similarly, we can compute the distances to any of the six individual sides of the cuboid.
Defining, for example, the plane orthogonal to the $x$-axis in its positive direction as the first side, we define the distance of a point $\vec{y}$ to it as:
\begin{align}
    d^{\text{P}}_{1}(\vec{h}, \vec{y})^2 = & (\hat{x}-a_x)^2 + \\ & \max(|\hat{y}|-a_y, 0)^2 +  \max(|\hat{z}|-a_z, 0)^2 \, . \nonumber
\end{align}
Distances for the other sides are calculated accordingly.

\begin{figure}[b]
\centering
\includegraphics[width=0.5\linewidth]{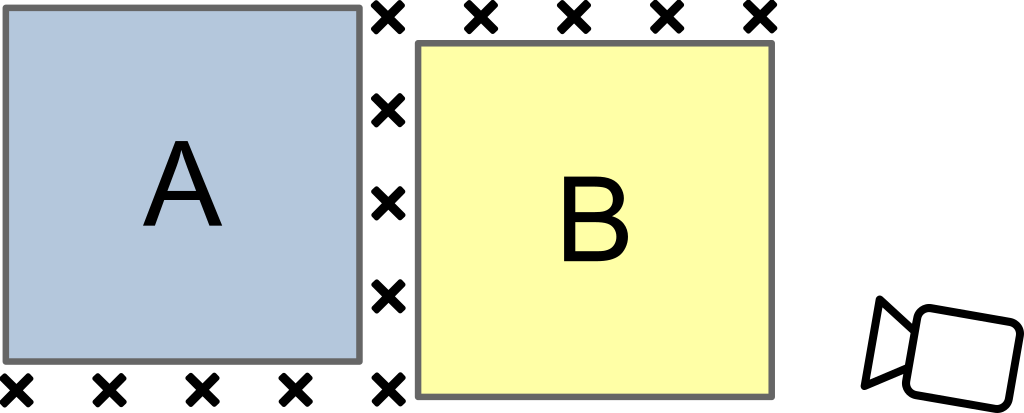}
\caption{ \textbf{Occlusion:} Given are a point cloud (\ding{53}), two cuboids (A and B) and a camera observing the scene. Cuboid A is a better fit since it does not occlude any points.}
\label{fig:occlusion}
\end{figure}

\paragraph{Occlusion Detection}
\label{subsubsec:oa_point_cuboid_distance}
When dealing with 2.5D data which only represents visible parts of the scene, simply using minimal point-to-cuboid distances as in Eq.~\ref{eq:point_cuboid_dist} is not adequate. 
Fig.~\ref{fig:occlusion} gives an intuitive example:
The mean distance of all points to their closest surface of either cuboid A or B is the same.
However, the {visible} surfaces of cuboid B occlude all points while not representing any structure present in the scene.
Cuboid A, on the other hand, does not occlude any points, and its visible surfaces fit very well to an existing structure.
Although other parts of its surface do not represent any structure either, they are self-occluded and thus of no interest.
So, cuboid A is a much better fit than cuboid B.

{
In order to detect whether a cuboid $\vec{{h}}$ occludes a point $\vec{y}$ from the perspective of a camera with centre $\vec{c}$, we must first translate $\vec{c}$ into the cuboid-centric coordinate frame.
Without loss of generality, we assume that the global coordinate frame aligns with the camera, \ie $\vec{c} = (0,0,0)\tran$.
The camera centre in the cuboid-centric coordinate system is thus $\vec{\hat{c}} = -\vec{R}\tran\vec{t}$.
We parametrise the line of sight for a point $\vec{y}$ as:
\begin{equation}
    \vec{x}(\lambda) = \vec{\hat{y}} + \lambda \vec{v} \, , \, \mathrm{with} \, \vec{v} = \vec{\hat{c}}-\vec{\hat{y}} \, .
\end{equation}
We determine its intersections with each of the six cuboid planes.
For the first plane orthogonal to the $x$-axis, this implies that:
\begin{equation}
    \lambda = \frac{a_x - \vec{\hat{y}}\tran\vec{e}_x}{\vec{v}\tran\vec{e}_x},
\end{equation}
where $\vec{e}_x$ denotes the $x$-axis unit vector. 
If $\lambda < 0$, $\vec{{y}}$ lies in front of the plane and is thus not occluded. 
Otherwise, we must check whether the intersection actually lies on the cuboid, \ie $d(\vec{h}, \vec{x}(\lambda)) = 0$.
If that is the case, then $\vec{y}$ is occluded by this part of the cuboid. 
We repeat this check accordingly for the other five cuboid planes.
Via this procedure we define an indicator function:
\begin{equation}
\label{eq:chi_o}
    \chi_{\text{o}}(\vec{y}, \vec{h}, i) =
\begin{cases}
1 ~&\text{ if }~ i\text{-th plane of } \vec{h}  \text{ occludes }  \vec{y}, \\
0 ~&\text{ else }, \, \text{with } i \in \{1, \dots, 6\} \, .
\end{cases}
\end{equation}
}

\paragraph{Occlusion-Aware Point-to-Cuboid Distance}
In order to correctly evaluate scenarios with occlusions, we propose an \emph{occlusion-aware} point-to-cuboid distance:
Given a point $\vec{y}$ and set of cuboids $\set{M} = \{\vec{h}_1, \dots, \vec{h}_{|\set{M}|}\}$, we compute its distance to the most distant occluding surface.
In other words, we compute the minimal distance $\vec{y}$ needs to travel in order to become visible:
\begin{equation}
d_{\text{o}}(\set{M}, \vec{y}) = 
\max_{\vec{h} \in \set{M}, \, i \in \{1, \dots, 6\}} 
\left( \,
\chi_{\text{o}}(\vec{h}, \vec{y}, i) \cdot d^{\text{P}}_{i}(\vec{h}, \vec{y}) 
\, \right) \, . \nonumber
\label{eq:occlusion.metric}
\end{equation}
If $\vec{y}$ is not occluded at all, this distance is zero.
We hence define the occlusion-aware distance of a point to a set of cuboids as:
\begin{equation}
    d_{\text{oa}}(\set{M}, \vec{y}) = 
    \max 
    \left( 
    \min_{\vec{h} \in \set{M}} d(\vec{h}, \vec{y}), \; d_{\text{o}}(\set{M}, \vec{y}) 
    \right) \, .
\end{equation}

\subsection{Robust Fitting}
\begin{figure}	
\centering
\includegraphics[width=0.65\linewidth]{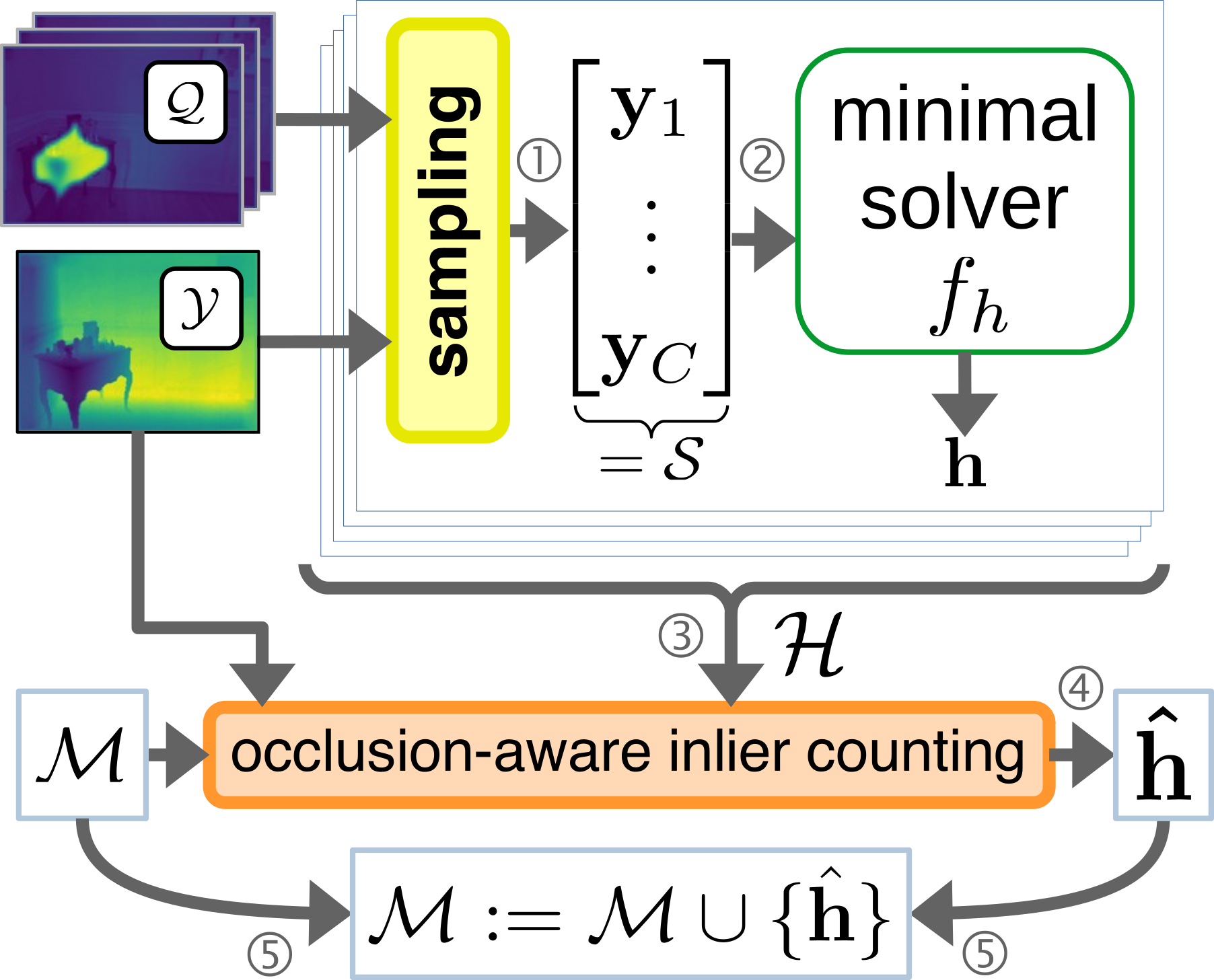}
\caption{ 
\textbf{Sampling and Fitting:}
(1) We sample minimal sets of features $\set{S} \subset \set{Y}$ using sampling weights $\set{Q}$ (Sec.~\ref{par:sampling}).
{(2) The solver $f_h$ (Sec.~\ref{subsubsec:fitting}-\ref{subsubsec:fitting_neural}, Fig.~\ref{fig:solvers}) computes cuboid parameters $\vec{h}$ from $\set{S}$. }
(3) We compute multiple cuboid hypotheses concurrently, resulting in a set of hypotheses $\set{H}$.
(4) Using occlusion-aware inlier counting (Sec.~\ref{subsec:occlusion_inlier}), we select the best hypothesis $\vec{\hat{h}}$ and (5) add it to the output set of recovered cuboids $\set{M}$.
}
\label{fig:fitting}
\end{figure} 
\label{subsec:robustfitting}
We seek to robustly fit a set of cuboids $\set{M} = \{\vec{h}_1, \dots, \vec{h}_{|\set{M}|}\}$ to the possibly noisy 3D features $\vec{y} \in \set{Y}$.
To this end, we build upon the robust multi-model fitting approach of Kluger et al.~\cite{KluBra2020}:
\begin{compactenum}
\item We predict sets of sampling weights $\vec{p}$ from data $\set{Y}$ using a neural network $f_{\vec{w}}$.
\item Using these sampling weights, a RANSAC-based estimator generates a cuboid instance $\vec{h}$, which we append to $\set{M}$.
\item Conditioned on $\set{M}$, we update the sampling weights $\vec{p}$ via $f_{\vec{w}}$ and generate the next cuboid instance.
\end{compactenum}
We repeat these steps multiple times, until all cuboids have been recovered one-by-one. 
Fig.~\ref{fig:system} gives an overview of the algorithm, while Fig.~\ref{fig:fitting} depicts the sampling and fitting stage in more detail.

\subsubsection{Sampling}
\label{par:sampling}
\begin{figure}
\centering
\includegraphics[width=0.99\linewidth]{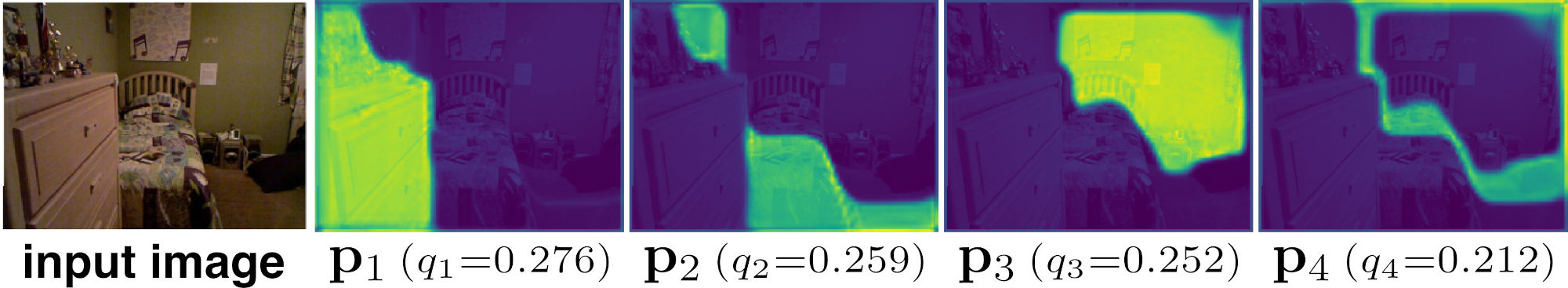}
\caption{ \textbf{Sampling Weights: } 
We predict multiple sets of sampling weights $\set{Q} = \{\vec{p}_1, \dots, \vec{p}_Q\}$ and corresponding selection probabilities $\vec{q} = [q_1, \dots, q_Q]$. In this example, the first three sampling weight sets roughly cover distinct parts of the scene.
The fourth set $\vec{p}_4$ does not, but also has the lowest selection probability. }
\label{fig:probability_maps}
\end{figure}

In~\cite{KluBra2020}, one set of sampling weights $\vec{p}(\set{Y} \, | \, \set{M})$ is predicted at each step.
Optimally, these weights should highlight a single coherent structure in $\set{Y}$ and suppress the rest, in order to maximise the likelihood of sampling an all-inlier set of features.
However, we often have multiple important structures present in a scene.
Which structure to emphasise and sample first may therefore be ambiguous, and the approach proposed in~\cite{KluBra2020} struggles with primitive fitting for that reason. 
In order to deal with this, we allow the network to 
instead predict several sets of sampling weights $\set{Q} = \{\vec{p}_1, \dots, \vec{p}_Q\}$ with selection probabilities $\vec{q}(\set{Y} \, | \, \set{M}) \in \mathbb{R}^Q$, as shown in Fig.~\ref{fig:probability_maps}.
This results in a two-step sampling procedure:
First, we randomly select one of the sampling weight sets $\vec{p} \in \set{Q}$ according to $\vec{q}$.
Then we sample a minimal set of features $\set{S} = \{\vec{y}_1, \dots, \vec{y}_C\} \subset \set{Y}$ according to the selected weights $\vec{p}$ in order to generate a cuboid hypothesis.
This allows the neural network $(\set{Q}, \vec{q}) = f_{\vec{w}}(\set{Y} | \set{M})$ to highlight multiple structures without them interfering with each other.
Ideally, $\vec{q}$ contains non-zero values for all sampling weight sets highlighting valid structures, and zero values otherwise.
At worst, it would degenerate to $\vec{q} = \vec{e}_i$, with $i \in \{1, \dots, Q\}$, effectively setting $Q=1$.
We avoid this via regularisation (cf. Sec.~\ref{sec:training}).
\subsubsection{Fitting: Iterative Numerical Optimisation}
\label{subsubsec:fitting}

Given a minimal set of sampled features $\set{S} \subset \set{Y}$, we want to find parameters of a cuboid $\vec{h} = f_h(\set{S})$ such that it fits these features optimally.
{Unfortunately, no closed form solution for $f_h$ exists.
We therefore approximate it using either of the following two approaches: iterative numerical optimisation, and regression via a neural network.
For numerical optimisation, }we define the cuboid solver $f_h$ via this objective function:
\begin{equation}
    F(\vec{y}, \vec{h}) = d(\vec{h}, \vec{y})^2 \cdot (a_x + a_y + a_z) \, .
    \label{eq:def.F}
\end{equation}
As the size of the cuboid is ambiguous when dealing with 2.5D data, we add the regularisation term $(a_x + a_y + a_z)$ in order to favour smaller cuboids.
{
We first estimate the initial position of the cuboid via the mean of the features, \ie $\vec{t}_0 = \frac{1}{|\set{S}|} \sum_{\vec{y} \in \set{S}} \vec{y}$. Then we estimate the rotation $\vec{R}$ via singular value decomposition:
\begin{equation}
\vec{R}_0 = \vec{V}\tran, \, \text{ with } \vec{USV}\tran = \left[\vec{y}_1 \dots \vec{y}_{|\set{S}|}  \right]\tran \, .
\end{equation}
Lastly, we initialise the cuboid size with the element-wise maximum of the absolute coordinates of the centred and rotated features, with $ i \in \{1, \dots, |\set{S}|\}$:
\begin{equation}
    \begin{pmatrix}
    a_{x0} \\ a_{y0} \\ a_{z0} 
    \end{pmatrix}
    =
    \begin{pmatrix}
    \max_{i} |\hat{x}_i| \\ \max_{i} |\hat{y}_i| \\ \max_{i} |\hat{z}_i| 
    \end{pmatrix}
    \, , \text{ with }
    \begin{pmatrix}
    \hat{x}_i \\ \hat{y}_i \\ \hat{z}_i
    \end{pmatrix}
    =
    \vec{R}_0 (\vec{y}_i - \vec{t}_0) \, .
    \nonumber
\end{equation}
This gives us the initial estimate $\vec{h}_0 = (a_{x0}, a_{y0}, a_{z0}, \vec{R}_0, \vec{t}_0)$.}
Starting from $\vec{h}_0$, we approximate {$\vec{h} \in \argmin_{\vec{h}} \| F(\set{S}, \vec{h}) \|_1$} via an iterative numerical optimisation method such as \mbox{L-BFGS}~\cite{liu89} or Adam~\cite{KingmaB14}. {Fig.~\ref{fig:numeric_solver} illustrates this procedure.}

{
\subsubsection{Fitting: Cuboid Regression Network}
\label{subsubsec:fitting_neural}

While numerical optimisation yields accurate results, it is computationally expensive and time consuming.
We thus propose an alternative approach, which uses a neural network
to predict cuboid parameters $\vec{h} = (a_x, a_y, a_z, \vec{R}, \vec{t})$ from a minimal set of features $\set{S}$ in a single forward pass.

\paragraph{Network Architecture}
The input to the network consists of a minimal set of features $\set{S}$ which we centre beforehand, \ie $\set{S}-\frac{1}{|\set{S}|} \sum_{\vec{y} \in \set{S}} \vec{y}$.
As depicted in Fig.~\ref{fig:neural_solver}, the network consists of a $1 \times 1$ convolutional layer, followed by four transformer encoder layers~\cite{vaswani2017attention}.
An average pooling layer reduces the feature set produced by the transformer layers to a constant size irrespective of size and ordering of $\set{S}$.
The order of the points in $\set{S}$ thus does not matter.
Lastly, five linear layers followed by tanh and sigmoid activation functions provide the actual cuboid parameters.\footnote{Please refer to the appendix for more details.}
\begin{figure}	
		\begin{subfigure}{0.4\linewidth}
			\includegraphics[width=\linewidth]{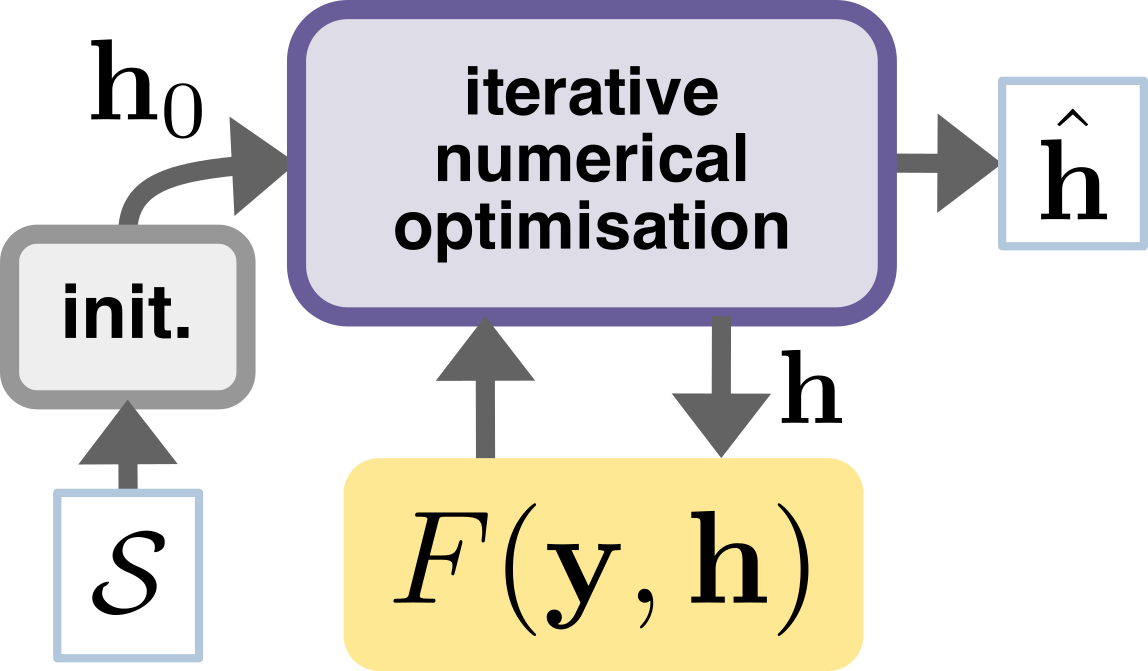}
			\caption{}
		\label{fig:numeric_solver}
		\end{subfigure}
		\hfill
		\begin{subfigure}{0.54\linewidth}
			\includegraphics[width=\linewidth]{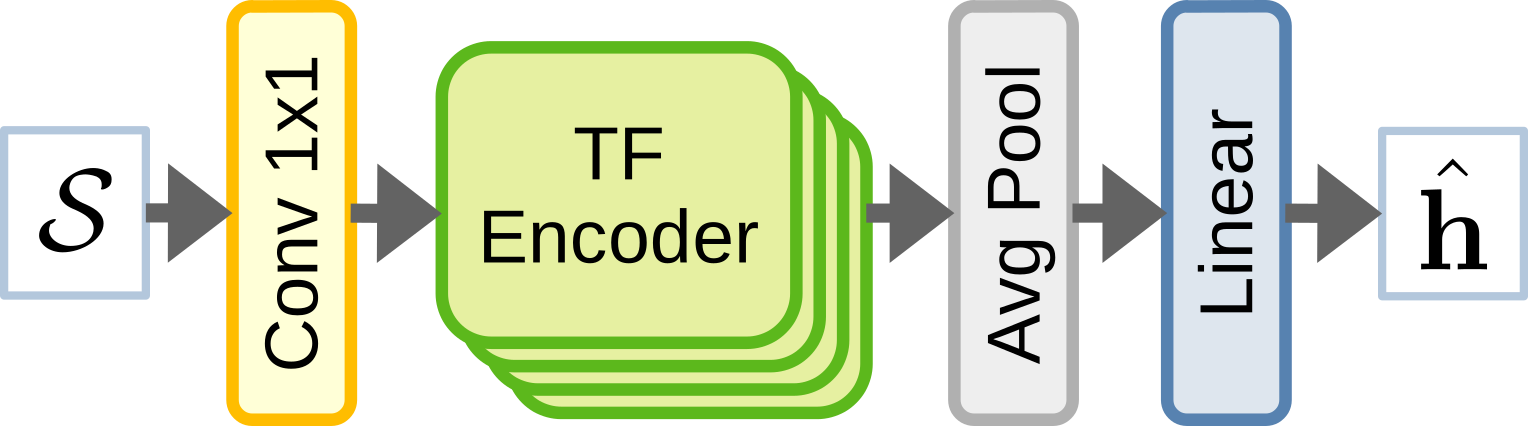}
			\caption{}
		\label{fig:neural_solver}
		\end{subfigure}

		\caption{{\textbf{Cuboid Solvers:} We propose two methods for estimating cuboid parameters $\hat{\vec{h}}$ from a minimal set of points $\set{S}$. 
		(a) Iterative numerical optimisation: starting from an initial estimate $\vec{h}_0$, we minimise an objective function $F$ so that the points in $\set{S}$ lie on the surface of cuboid $\vec{h}$ (Sec.~\ref{subsubsec:fitting}). 
		(b) Cuboid regression network: using a neural network, we directly predict cuboid parameters $\hat{\vec{h}}$ from $\set{S}$ in a single forward pass (Sec.~\ref{subsubsec:fitting_neural}). } }
		\label{fig:solvers}
\end{figure}   

\paragraph{Pre-Training}
Before plugging the cuboid regression network into our cuboid fitting pipeline (Figs.~\ref{fig:system} and~\ref{fig:fitting}), we train it using synthetic data.
First, we randomly sample cuboid parameters $\tilde{\vec{h}}$.
Then, we sample a minimal set of points $\set{S}$ from the cuboid surfaces which are visible from the point of view of a virtual camera located at origin.
The sampling probability for each of the visible cuboid surfaces is proportional to the effectively visible area in order to mimic a realistic distribution of the points compared to points sampled from an actual depth map.
Using the following loss function, we optimise the neural network parameters so that the sampled points $\set{S}$ lie on the surface of the predicted cuboid $\vec{h}$:
\begin{equation}
    \ell_{\mathrm{pre}} = \frac{1}{|\set{S}|} \sum_{\vec{y} \in \set{S}} d(\vec{h}, \vec{y})^2 \, .
\end{equation}

}

\subsubsection{Occlusion-Aware Inlier Counting}
\label{subsec:occlusion_inlier}
In order to select the cuboid hypothesis $\vec{h} \in \set{H}$  which fits best to features $\set{Y}$, given a set of existing cuboids $\set{M}$, we need to define an inlier function $f_{\text{I}}(\cdot)$.
We could naively take $f_{\text{I}}(\vec{y}, \vec{h}) \in [0, 1]$, with $f_{\text{I}}(\vec{y}, \vec{h}) = 1$ if feature $\vec{y}$ is well represented by cuboid $\vec{h}$, and $f_{\text{I}}(\vec{y}, \vec{h}) = 0$ otherwise.
However, as described in Sec.~\ref{subsubsec:oa_point_cuboid_distance}, we want to avoid cuboids which create occlusions.
Hence, we define an occlusion-aware inlier function $f_{\text{OAI}}(\vec{y}, \set{M})$ with the additional property $f_{\text{OAI}}(\vec{y}, \set{M}) < 0$ if $\vec{y}$ is occluded by cuboids in $\set{M}$, but \emph{only} if it is occluded by cuboid sides to which it is not also an inlier. 
{Using an outlier function $f_{\text{O}}$ and the indicator function $\chi_{\text{o}}$ (Eq.~\ref{eq:chi_o}), w}e define: {
  $$  f_{\text{IO}}(\vec{y}, \vec{h}, i) = f_{\text{I}}(d_i^{\text{P}}(\vec{y}, \vec{h})^2) - \chi_{\text{o}}(\vec{y}, \vec{h}, i) \cdot f_{\text{O}}(d_i^{\text{P}}(\vec{y}, \vec{h})^2) \, , $$ }
to determine whether $\vec{y}$ is an inlier to the $i$-th side of cuboid $\vec{h}$ ($f_{\text{IO}}>0$), occluded by it ($f_{\text{IO}}<0$), or a regular outlier ($f_{\text{IO}}=0$). If $\vec{y}$ is occluded by any cuboid side in $\set{M}$, it must be marked as occluded. Otherwise it should be marked as an inlier to its closest cuboid, or as an outlier:
\begin{equation}
     f_{\text{OAI}}(\vec{y}, \set{M}) = \\
\begin{dcases}
\min_{\vec{h}, i} f_{\text{IO}}(\vec{y}, \vec{h}, i)                                          
    ~&\text{if }~ (\cdot) < 0  \, , \\ 
\max_{\vec{h}, i} f_{\text{IO}}(\vec{y}, \vec{h}, i)
    ~&\text{else} \, ,
\end{dcases}
\end{equation}
with $\vec{h} \in \set{M}$ and $i \in \{1, \dots, 6\}$, \ie  all sides of all cuboids.
This implies that features which are occluded indeed reduce the inlier count $I_{\text{c}}$, which we use to determine which cuboid hypothesis $\vec{h}$ shall be added to our current set of cuboids $\set{M}$:
\begin{equation}
    I_{\text{c}}(\set{Y}, \set{M} \cup \{\vec{h}\}) = \sum_{\vec{y} \in \set{Y}} f_{\text{OAI}}(\vec{y}, \set{M}\cup \{\vec{h}\}) \, .
    \label{eq:inlier_count}
\end{equation}
{
For $f_{\text{I}}$, we use a soft inlier measure derived from~\cite{KluBra2020}:
\begin{equation}
    f_{\text{I}}(d) = 1 - \sigma( \frac{\beta}{\tau}  d - \beta) \, ,
\end{equation}
with softness parameter $\beta$, inlier threshold $\tau$, and $\sigma(\cdot)$ being the sigmoid function. 
In~\cite{kluger2021cuboids}, we proposed $f_{\text{O}}(d) = 1-f_{\text{I}}(d)$ as the occlusion function, as it requires a behaviour opposite to $f_{\text{I}}$.

\paragraph{Occlusion Penalty}
\label{par:occlusion_penalty}
However, we have noticed that this $f_{\text{O}}$ is not discriminative enough \wrt points which have a large distance to an occluding surface, since it eventually saturates, \ie $\lim \limits_{d \to \infty} f_{\text{O}}(d) = 1$.
We hence propose a new \emph{leaky} occlusion function which is linear above a threshold $\tau_c$: 
\begin{equation}
     f_{\text{O}}(d) = \\
\begin{dcases}
1-f_{\text{I}}(d)                                          
    ~&\text{if }~ d < \tau_c  \, , \\ 
m \cdot d + b
    ~&\text{else} \, ,
\end{dcases}
\end{equation}
with $m = -f_{\text{I}}'(\tau_c)$ and $b = 1- f_{\text{I}}(\tau_c) - m \cdot \tau_c$ in order to preserve the first order derivative at $d=\tau_c$.
By setting $\tau_c$ appropriately, we can decide to penalise occlusion more or less strongly than before.
We set it to a fixed value $\tau_c = 2 \cdot \tau$ during inference.
During training, we start with a large value $\tau_c >> \tau$ (\ie small occlusion penalty) and linearly decrease it over time:
\begin{equation}
    \tau_c = \frac{\mathrm{epoch}}{N_e} (\tau-1) + 1 \, ,
\end{equation}
with $N_e$ being the maximum number of training epochs.
}
{
\subsection{Minimal Set Size}
\label{subsec:minimal_set_size}

In the context of robust model fitting, the minimal set size $C$ is the smallest number of data points which must be used to unambiguously determine the parameters of a model.
$C$ is defined by the degrees of freedom $d$ of the model and the number of constraints $c$ imposed on the model by each data point: 
$C = \lceil \frac{d}{c} \rceil$.
For example, when fitting a planar homography to 2D point correspondences ($d=8$, $c=2$), the minimal set size is $C=4$.
In our case, each cuboid has $d=9$ degrees of freedom and each point imposes a one dimensional constraint according to Eq.~\ref{eq:def.F}, \ie $c=1$ and hence $C=9$.
However, as we are dealing with 2.5D depth data, we cannot determine the parameters of a cuboid unambiguously without additional constraints, since at least three sides of the cuboid will always appear self-occluded.
One such constraint is already present in Eq.~\ref{eq:def.F} for each size parameter $(a_x, a_y, a_z)$, which aims at finding the smallest cuboid that fits the selected data points.
It hence appears sufficient to sample minimal sets of size $C=6$.
In order to find out which approach works best in practice, we consider the following settings:

$C=9$: This is the setting used in~\cite{kluger2021cuboids} and, as described above, defines the minimum number of points needed to fit a cuboid to three-dimensional data unambiguously. 
    As illustrated by Fig.~\ref{fig:mss9}, we do not actually require that many points when dealing with 2.5D data and the additional constraint that the cuboid shall be as small as possible. 
    While sampling larger than minimal sets can result in higher quality model hypotheses~\cite{tennakoon2015robust}, this comes at the cost of increasing the probability of sampling outliers.

$C=6$: As Fig.~\ref{fig:mss6} illustrates, at most six points are sufficient to define the 2D silhouette of a cuboid, and hence for determining the parameters of the cuboid with the smallest size from 2.5D data.
    
$C<6$ will generally not lead to desirable results, as it is possible to find a smaller cuboid than with $C \geq 6$.
This cuboid may fit the sampled points well, but will be different from the true cuboid which optimally describes the geometric structure from which the points are sampled, as illustrated by Fig.~\ref{fig:mss5}.

\begin{figure}	
\centering
		\begin{subfigure}{0.18\linewidth}
			\includegraphics[width=\linewidth]{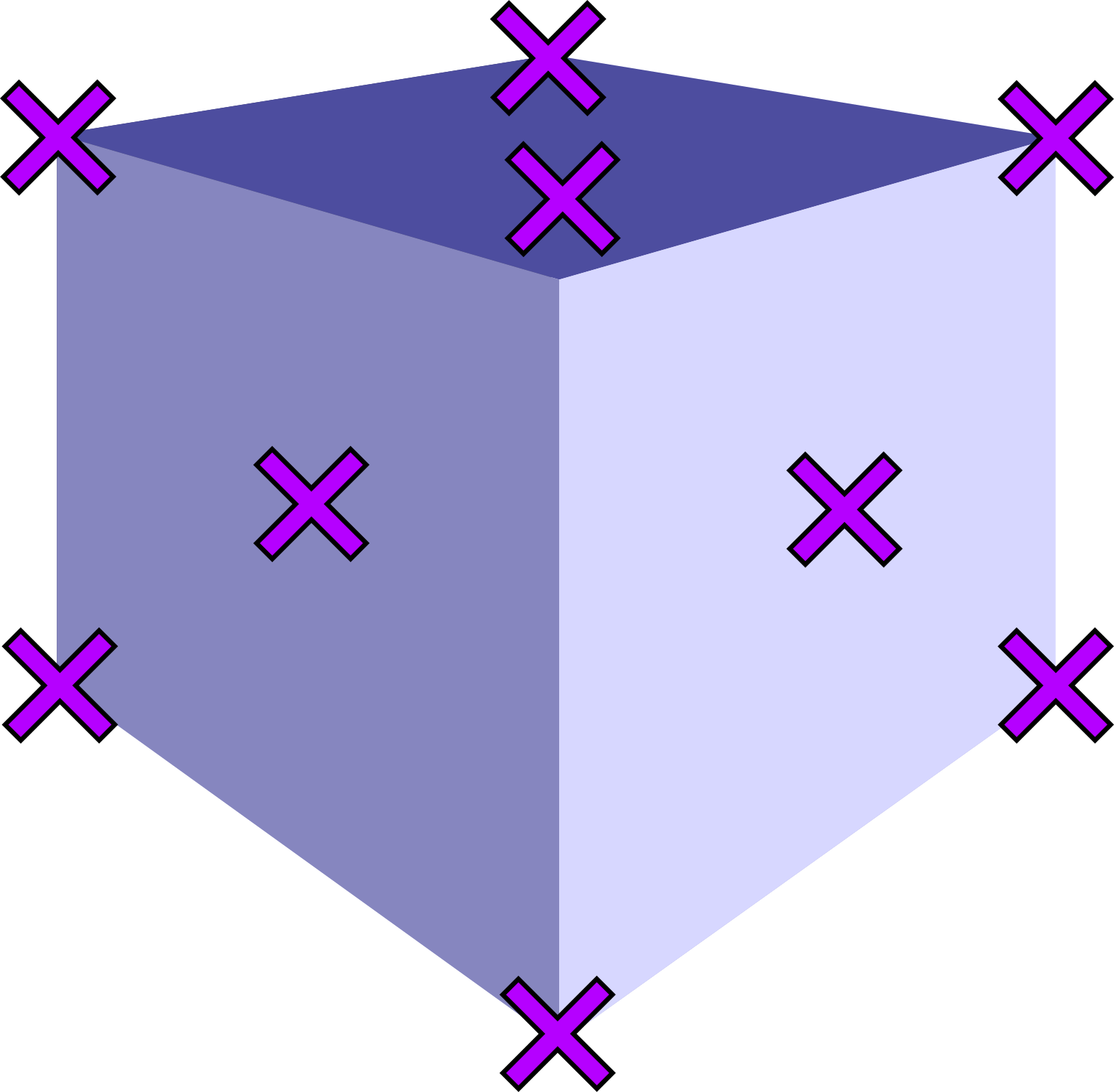}
			\caption{}
		\label{fig:mss9}
		\end{subfigure}
            \hspace{2em}
		\begin{subfigure}{0.18\linewidth}
			\includegraphics[width=\linewidth]{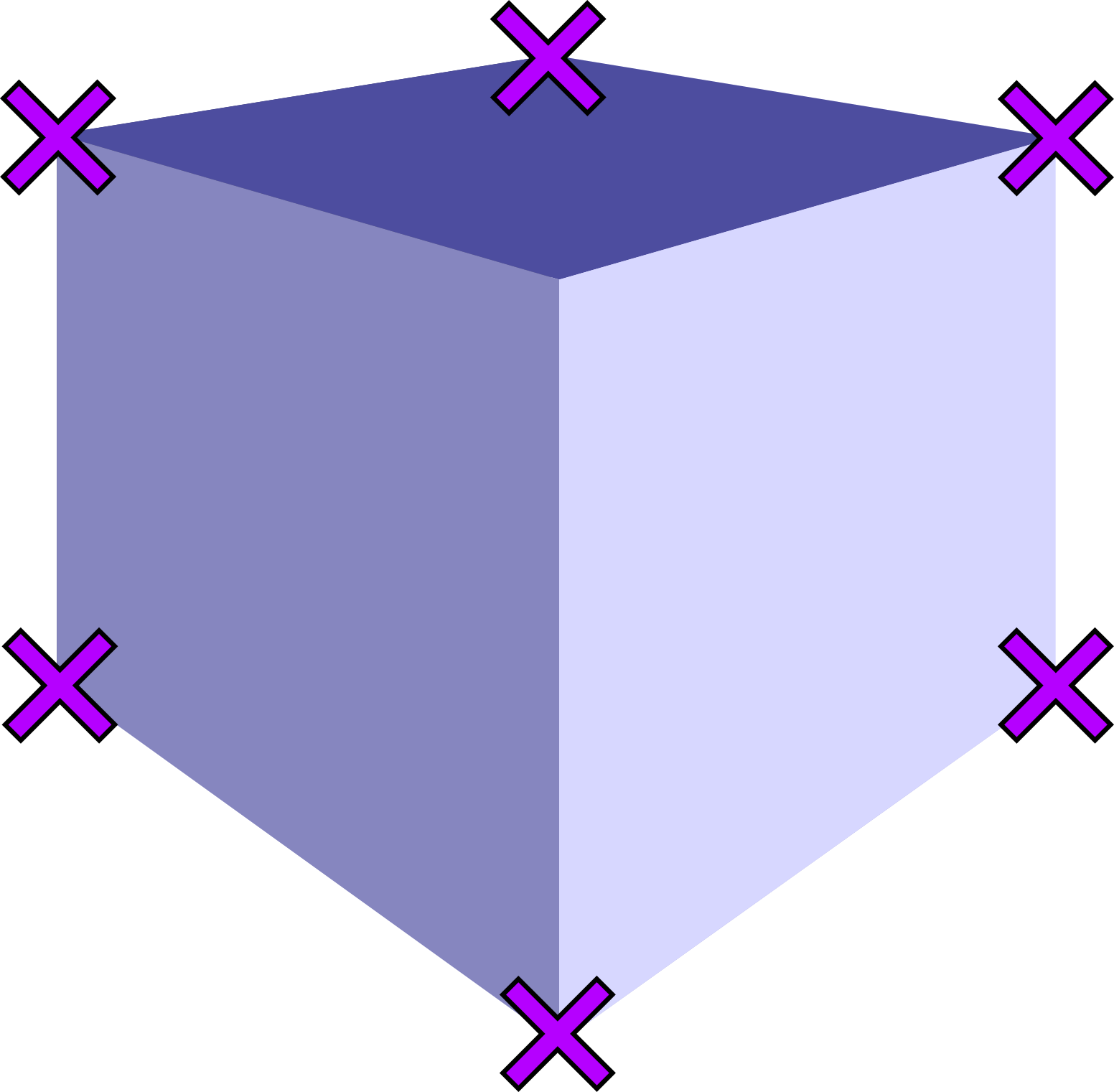}
			\caption{}
		\label{fig:mss6}
		\end{subfigure}
            \hspace{2em}
		\begin{subfigure}{0.18\linewidth}
			\includegraphics[width=\linewidth]{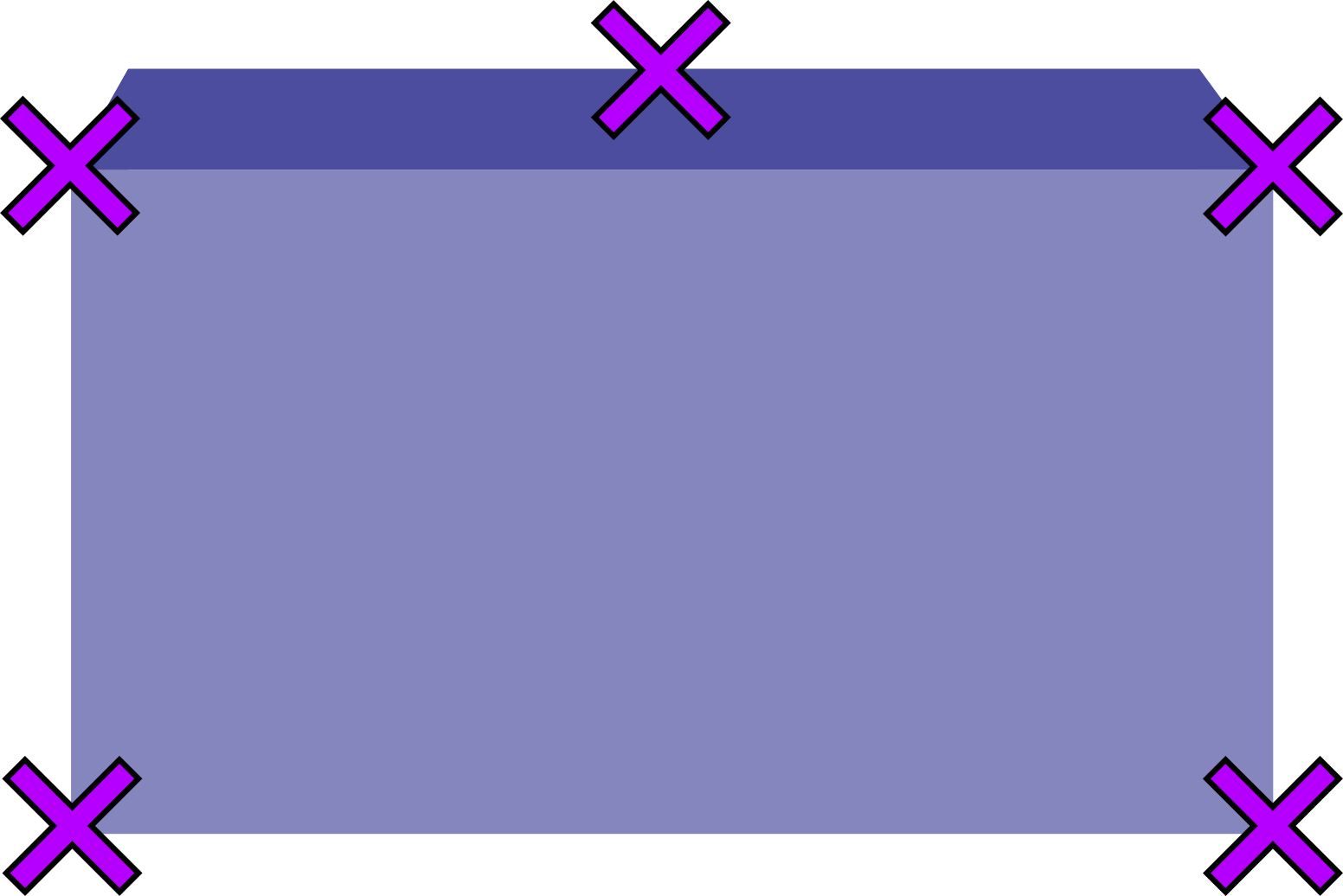}
			\caption{}
		\label{fig:mss5}
		\end{subfigure}

		\caption{{
		\textbf{Minimal Set Size:} We illustrate different choices for the minimal set size $C$ in the context of 2.5D data.
		(a) $C=9$: Nine points are more than what is needed to define a cuboid if using a minimal volume constraint. 
		(b) $C=6$: If sampled appropriately, six points suffice to define the cuboid unambiguously.
		(c) $C<6$: With fewer points, the resulting cuboid is likely to be different due to the minimal volume constraint.
		See Sec.~\ref{subsec:minimal_set_size} for details.
		}}
		\label{fig:mss}
\end{figure}
}
{
\subsection{Stopping Criterion}
\label{sec:stopping_criterion}
Our proposed method predicts cuboids sequentially, with each cuboid potentially increasing the reconstruction accuracy.
However, as our goal is 3D scene \emph{abstraction} instead of reconstruction, we need to define a stopping criterion in order to avoid over-segmentation of the scene.
In~\cite{kluger2021cuboids}, we defined such a criterion via the number of inliers added by each new cuboid:
\begin{equation}
    I_{\text{c}}(\set{Y}, \set{M} \cup \{\vec{h}\}) - I_{\text{c}}(\set{Y}, \set{M}) > \Theta \, .
    \label{eq:stopping}
\end{equation}
\ie we add a newly selected cuboid $\vec{h}$ only to $\set{M}$ if it would increase the inlier count by at least $\Theta$, which is a user definable parameter.
Otherwise the method terminates and returns $\set{M}$ as the recovered primitive configuration.
Alternatively, we could employ existing model selection criteria, such as the Akaike information criterion (AIC)~\cite{akaike1974new} and the Bayesian information criterion (BIC)~\cite{schwarz1978estimating}.
The goal of such criteria is to strike a balance between low model fitting error (overfitting) and low model complexity (underfitting).
If we define a set of cuboids $\set{M}$ as a single compound geometric model, then adding an additional cuboid to $\set{M}$ is equivalent to an increase in model complexity.
For the task of fundamental matrix and homography fitting, the Geometric Robust Information Criterion (GRIC)~\cite{torr2002bayesian} has been shown to provide superior results compared to both AIC and BIC:
\begin{equation}
    \mathrm{GRIC} = \sum_i \rho \left( \frac{e^2_i}{\sigma^2} \right) + \lambda  n  d + k \cdot \log(n) \, ,
    \label{eq:gric_orig}
\end{equation}
with $n$ being the number of data points, $d$ being the dimensionality of the model constraint, and $k$ being the number of parameters in the model.
Assuming a Gaussian error distribution, the first term of Eq.~\ref{eq:gric_orig} sums the errors $e_i$ of all data points, normalised by standard deviation $\sigma$ and truncated to reduce the influence of outliers:
\begin{equation}
    \rho(e) = \min(e,T) \, .
\end{equation}
This term becomes smaller when a (more complex) model reduces per-point errors $e^2_i$ below a threshold $\sigma^2 \cdot T$, and behaves reciprocal to a soft truncated inlier count.
We thus replace it with the negative inlier count $-I_c$ (Eq.~\ref{eq:inlier_count}) using our occlusion-aware inlier metric.
As the dimensionality $d$ of the model constraint is constant in our case, the second term as a whole is constant and can thus be ignored.
Lastly, the number of model parameters is $k = 9 \cdot |\set{M}|$.
With this, we define the following selection criterion:
\begin{equation}
    \mathrm{C}(\set{Y}, \set{M}) = -I_{\text{c}}(\set{Y}, \set{M}) + 9 \cdot |\set{M}| \cdot \log(n) \, ,
\end{equation}
We then add a new cuboid $\vec{h}$ only to $\set{M}$ if it decreases $\mathrm{C}(\cdot)$, \ie: 
\begin{equation}
\mathrm{C}(\set{Y} , \set{M} \cup \{\vec{h}\}) < \mathrm{C}(\set{Y}, \set{M}) \, .
\end{equation}
This is equivalent to Eq.~\ref{eq:stopping}, with $\Theta = 9 \cdot \log(n)$.
}

\subsection{Training}
\label{sec:training}
As in~\cite{KluBra2020}, we want to optimise parameters $\vec{w}$ of the sampling weight network in order to increase the likelihood of sampling all-inlier minimal sets of features.
To achieve this, we minimise the expectation of a task loss $\ell(\vec{h}, \set{M})$ which measures how well a new cuboid $\vec{h}$ and previously determined cuboids $\set{M}$ fit to a scene:
\begin{equation}
    \mathcal{L}(\vec{w}) = \mathbb{E}_{\set{H} \sim p(\set{H} | \set{M}; \vec{w})} \left[ \ell (\vec{\hat{h}}, {\set{{M}}}) \right] \, ,
    \label{eq:exp_task_loss_w}
\end{equation}
with $\vec{\hat{h}}$ being the cuboid hypothesis selected according to Eq.~\ref{eq:inlier_count}:
\begin{equation}
    \vec{\hat{h}} \in \argmax_{\vec{h} \in \set{H}} I_{\text{c}}(\set{Y}, \set{M} \cup \{\vec{h}\}) \, .
\end{equation}
As described in~\cite{Brachmann2017:DSAC, Brachmann2019:NGRANSAC}, this discrete hypothesis selection prohibits learning parameters $\vec{v}$ of the feature extraction network.
We thus turn hypothesis selection into a probabilistic action:
{
\begin{equation}
    \vec{\hat{h}} \sim p(\vec{\hat{h}} | \set{H}, \set{M}) = \frac{\exp ( \alpha \cdot I_{\text{i}}(\set{Y}, \set{M}, \vec{\hat{h}})) } { \sum_{\vec{{h}} \in \set{H} } \exp ( \alpha \cdot I_{\text{i}}(\set{Y}, \set{M}, \vec{{h}})) } \, ,
    \label{eq:probabilistic_selection}
\end{equation}
with parameter $\alpha$ controlling the selectivity of the softmax operator, and $I_{\text{i}}$ denoting the difference in inlier count due to $\vec{h}$:
\begin{equation}
    I_{\text{i}}(\set{Y}, \set{M}, \vec{\hat{h}}) = I_{\text{c}}(\set{Y}, \set{M} \cup \{\vec{\hat{h}}\}) - I_{\text{c}}(\set{Y}, \set{M}) \, .
\end{equation}
As we select and add cuboids to $\set{M}$ sequentially, the absolute inlier count $I_{\text{c}}$ generally increases, thus shifting the selectivity of Eq.~\ref{eq:probabilistic_selection}.
Using the inlier count difference instead decouples Eq.~\ref{eq:probabilistic_selection} from the inliers already captured by the cuboids in $\set{M}$.
}
This allows us to compute the expected loss over cuboid hypotheses $\set{H}$, which is differentiable:
\begin{equation}
    \mathcal{L}(\vec{v}) = \mathbb{E}_{\vec{h} \sim p(\vec{h} | \set{H}, \set{M}) } \left[ \ell (\vec{h}, \set{M}) \right] \, .
    \label{eq:exp_task_loss_v}
\end{equation}
Combining Eq.~\ref{eq:exp_task_loss_w} and~\ref{eq:exp_task_loss_v}, we train both networks end-to-end:
\begin{equation}
    \mathcal{L}(\vec{v}, \vec{w}) = \mathbb{E}_{\set{H} \sim p(\set{H} | \set{M}; \vec{w})} \mathbb{E}_{\vec{h} \sim p(\vec{h} | \set{H}, \set{M}) } \left[ \ell (\vec{h}, \set{M}) \right] \, .
    \label{eq:exp_task_loss_vw}
\end{equation}
Computing the exact expectation in Eq.~\ref{eq:exp_task_loss_w} is intractable, so we approximate its gradient  by drawing $K$ samples of $\set{H}$: %
\begin{align}
    \frac{\partial\mathcal{L}(\vec{v},\vec{w})}{\partial (\vec{v},\vec{w})}   \approx 
     \frac{1}{K} \sum_{k=1}^K \left[ 
    \mathbb{E}_{\vec{h}} \left[ \ell \right] \frac{\partial \log p(\set{H}_k; \vec{w})}{\partial (\vec{v},\vec{w})}  +
    \frac{\partial \mathbb{E}_{\vec{h}}[\ell]]}{\partial (\vec{v},\vec{w})} 
    \right] \, . 
    \label{eq:exp_task_loss_grad_approx}
\end{align}

\paragraph{Differentiable Solver}
The above requires that the gradient for a particular 3D feature can be computed \wrt a resulting cuboid. 
However, {when we compute} cuboid parameters via iterative numerical optimisation (cf. Sec.~\ref{subsubsec:fitting}),  
tracking the operations through this step results in inaccurate gradients and prohibitively high computational costs. 
In the following, we describe a solution which does not necessitate tracking the computations.
Instead, it relies on the implicit function theorem to directly compute the desired gradients. 
Given a set of features $\set{S} = \{\vec{y}_1, \dots, \vec{y}_C\}$ and 
an optimally fit cuboid $\vec{h}$, we seek to compute the partial derivatives ${\partial\vec{h}}/{\partial\set{S}}$.
From Eq.~\ref{eq:def.F}, we have $F(\set{S}, \vec{h}) = \vec{0}$. Via the implicit function theorem, we can therefore obtain: 
\begin{equation}
   \frac{\partial\vec{h}}{\partial\set{S}} = - {\left(\frac{\partial F}{\partial\vec{h}}(\set{S}, \vec{h})\right)}\inverse \cdot \frac{\partial F}{\partial\set{S}}(\set{S}, \vec{h}) \, .
 \end{equation}
In practice, the partial derivatives by the size parameters $\partial F / \partial a_{\{x,y,z\}}$ are mostly zero or close to zero.
Inversion of the Jacobian $\frac{\partial F}{\partial\vec{h}}(\set{S}, \vec{h})$ is not possible or numerically unstable in this case. 
We thus approximate it by masking out the possibly zero derivatives and using the pseudo-inverse:
\begin{equation}
    \frac{\partial\vec{h}}{\partial\set{S}} \approx - \left( \frac{\partial F}{\partial(\vec{R}, \vec{t})}(\set{S}, \vec{h})\right)\pinv \cdot \frac{\partial F}{\partial\set{S}}(\set{S}, \vec{h}) \, .
\end{equation}
This allows us to perform end-to-end backpropagation even without a differentiable cuboid solver $\vec{h} = f_h(\set{S})$.
{Note that this derivation is not required if we use the neural solver (cf. Sec.~\ref{subsubsec:fitting_neural}) as it is inherently differentiable.}

\paragraph{Task Loss}
During training, we aim to maximise the expected inlier counts of  sampled cuboids, \ie we minimise: 
\begin{equation}
    \ell(\vec{{h}}, {\set{{M}}}) = -I_{\text{c}}(\set{Y}, \set{M} \cup \{\vec{h}\}) \, .
\end{equation}
In order enable gradient computation according to Eq.~\ref{eq:exp_task_loss_grad_approx}, $\ell$ must also be differentiable.
Instead of a hard threshold, we thus use a soft inlier measure derived from~\cite{KluBra2020}:
\begin{equation}
    f_{\text{I}}(\vec{y}, \vec{h}, i) = 1 - \sigma(\beta (\frac{1}{\tau}  d_i^{\text{P}}(\vec{y}, \vec{h})^2 - 1)) \, ,
\end{equation}
with softness parameter $\beta$, inlier threshold $\tau$, and $\sigma(\cdot)$ being the sigmoid function. 
This loss is based on geometrical consistency only, and does not need any additional labels.

\paragraph{Regularisation}
\label{sec:training_regularisation}
In order to prevent mode collapse of the sampling weights $\set{Q}$ (cf. Sec.~\ref{par:sampling}), we apply a regularisation term during training.
We minimise the correlation coefficients between individual sets of sampling weights $\vec{p} \in \set{Q}$:
\begin{equation}
    \ell_{\text{corr}}(\set{Q}) = \sum_{\substack{\vec{p}_i, \vec{p}_j \in \set{Q},\\  i \neq j}} \frac{\operatorname{cov}(\vec{p}_i, \vec{p}_j)}{\sigma_{\vec{p}_i} \sigma_{\vec{p}_j}} \, ,
\end{equation}
with covariances $\operatorname{cov}(\cdot,\cdot)$ and standard deviations $\sigma$. 
For the same reason, we also maximise the entropy of selection probabilities $\vec{q}$:
\begin{equation}
    \ell_{\text{entropy}} = -\operatorname{H}(\vec{q}) \, .
\end{equation}
{
We furthermore inherit the inlier masking regularisation (IMR) loss $\ell_{\text{im}}$ from~\cite{KluBra2020}.
The final loss $\ell_{\text{final}}$ is thus a weighted sum of all these losses:
\begin{equation}
    \ell_{\text{final}} = \ell + \kappa_{\text{corr}} \cdot \ell_{\text{corr}} + \kappa_{\text{entropy}} \cdot \ell_{\text{entropy}} + \kappa_{\text{im}} \cdot \ell_{\text{im}} \, ,
\end{equation}
where weights $\kappa_{\text{corr}}, \kappa_{\text{entropy}}, \kappa_{\text{im}}$ balance the individual losses.
}

\section{Experiments}
\label{sec:exp}

{
In this section, we evaluate different variants of our method:
\begin{itemize}
    \item Input: RGB image or ground truth depth
    \item Solver: numerical (Sec.~\ref{subsubsec:fitting}) or neural (Sec.~\ref{subsubsec:fitting_neural})
\end{itemize}
We compare these variants against each other, and against multiple other baselines.
In addition, we provide several ablation studies which examine the impact of important design choices and hyperparameters.
Unless stated otherwise, we use the ground truth depth as input, the numerical solver for cuboid fitting, and the parameters given in \new{the appendix} for our experiments.

The remainder of this section provides information about the datasets we used in our experiments, implementation details for our method, and an overview of the baseline algorithms we use for comparison.
In Sec.~\ref{subsec:metrics}, we explain our evaluation metrics.
Sec.~\ref{subsec:eval_minimal_solver} compares the performance of our numerical and neural cuboid solvers in detail, while Sec.~\ref{subsec:baseline_comparison} provides a comparison of our method with other baselines.
In addition, we show ablation studies \wrt the impact of the minimal set size (Sec.~\ref{subsec:eval_mss}), occlusion-aware inlier counting (Sec.~\ref{subsec:eval_oai}), as well as the number of sampling weight maps, regularisation losses and end-to-end network fine-tuning (Sec.~\ref{subsec:more_ablation}).
We provide the main quantitative results in Tab.~\ref{tab:main_results}, comparing variations of our method -- numerical and neural cuboid solver, depth and RGB input -- against other baselines.
In Figs.~\ref{fig:examples} \new{and~\ref{fig:smh_examples}}, we additionally show selected qualitative examples\footnote{\new{We provide more examples in the appendix.}}.

}

\paragraph{Datasets}
We provide the majority of qualitative and quantitative results on the NYU Depth v2 dataset~\cite{silberman2012indoor}. 
It contains 1449 images of indoor scenes with corresponding ground truth depth recorded with a Kinect camera.
We use the same dataset split as~\cite{lee2019big}: 654 images for testing and 795 images for training, of which we reserved 195 for validation.
{We perform the main experiments (Sec.~\ref{subsec:baseline_comparison}) on the test set, while we use the validation set for all ablation studies (Sec.~\ref{subsec:eval_minimal_solver} ff.).} 
\new{We additionally provide results for our method on the Synthetic Metropolis Homographies (SMH) dataset~\cite{kluger2024parsac}.
It consists of 48002 synthetically generated image pairs of outdoor scenes showing a 3D model of a city, and contains ground truth homographies and depth maps.
We selected a subset of scenes which have at least three homographies, \ie at least three different planar surfaces.
Using one of the two provided depth maps of each image pair for our experiments, this yields 29000 scenes for training, 1037 for validation and 2851 for testing.}

\paragraph{Implementation Details}
\label{subsec:implementation_details}
For experiments with RGB image input, we use the BTS~\cite{lee2019big} depth estimator pre-trained on NYU as our feature extraction network. 
BTS achieves state-of-the-art results on NYU with code publicly available\footnote{https://github.com/cogaplex-bts/bts}.
We implemented the sample weight estimator as a variant of the fully convolutional neural network used in~\cite{Brachmann2019:NGRANSAC}. 
We pre-train the sample weight estimation network with ground truth depth as input {and using the numerical solver} {and select the checkpoint which achieved the highest AUC$_{@\SI{20}{\centi\meter}}$ on the validation set.}. 
This is also the network we use for experiments with depth input.
{We pre-train the neural solver on synthetic data as described in Sec.~\ref{subsubsec:fitting_neural}.} 
{We then continue training all networks end-to-end for three different configurations separately: depth input and neural solver, RGB input and numerical solver, and RGB input and neural solver.
For this fine-tuning step, we disabled all regularisation losses (cf. Sec.~\ref{sec:training_regularisation}) as we observed a degradation of performance otherwise.
}
We implement the {numerical} solver (cf. Sec.~\ref{subsubsec:fitting}) by applying the Adam~\cite{KingmaB14} optimiser to perform gradient descent \wrt cuboid parameters for 50 iterations.
{Computation time was measured on a workstation with an i7-7820X CPU and RTX 3090 GPU running Ubuntu Linux 20.04.4 LTS.}
We provide a listing of hyperparameters in \new{the appendix}.

\paragraph{Baselines}
While multiple works in the field of primitive based 3D shape parsing have been published in recent years, not all of them can be used for comparison. 
We cannot use methods which require ground truth shape annotations~\cite{niu2018im2struct, zou20173d} or watertight meshes~\cite{genova2019learning, deng2020cvxnet,Paschalidou2020:Decomposition} for training, as the data we are concerned with does not provide these.
For the cuboid based approach of~\cite{tulsiani2017learning}, source code is available\footnote{https://github.com/shubhtuls/volumetricPrimitives}. %
However, following the instructions they provided, we were unable to obtain sensible results on NYU.
The neural network of~\cite{tulsiani2017learning} appears to degenerate when trained on NYU, predicting similar unreasonable cuboid configuration for all scenes (cf.~qualitative results in the appendix). 
We therefore did not include it in our evaluation.

{
For depth input, we compare against the superquadric-based approach of~\cite{paschalidou2019superquadrics} (SQ-Parsing), trained on NYU using their provided source code\footnote{https://github.com/paschalidoud/superquadric\_parsing}.}
Since SQ-Parsing is designed for meshes as input, we preprocess the ground truth depth point clouds of NYU by applying Poisson surface reconstruction~\cite{kazhdan2006poisson}.
We further compare our method against a variant of Sequential RANSAC~\cite{vincent2001seqransac} from depth input{, with and without occlusion-aware inlier counting (OAI), as well as a variant of CONSAC~\cite{KluBra2020} adapted for cuboid fitting, using our numerical cuboid solver but without OAI.}

{
For RGB input, we} compare against a variant of~\cite{paschalidou2019superquadrics} which directly operates on RGB images (SQ-Parsing RGB), and also against SQ-Parsing when applied to the prediction of the same depth estimation network~\cite{lee2019big} which we employ for our method. 
We used the same hyperparameters as~\cite{paschalidou2019superquadrics} for these experiments.

\subsection{Metrics}

\label{subsec:metrics}
Previous works~\cite{paschalidou2019superquadrics, Paschalidou2020:Decomposition} evaluated their results using Chamfer distance and volumetric IoU.
These works, however, deal with full 3D shapes of watertight objects, while we fit 3D primitives to scenes where only a 2.5D ground truth is available.
As we explain in Sec.~\ref{subsubsec:oa_point_cuboid_distance}, we therefore evaluate using the occlusion-aware distance metric, OA-$L_2$ for short, instead.
We calculate the occlusion-aware distances of all ground-truth points to the recovered primitives per scene.

{
For the task of 3D reconstruction, the main objective is to minimise the geometric distance between the predicted 3D model and the 3D ground truth.
While this is also one of the objectives for 3D \emph{abstraction}, we additionally want our scene representations to be parsimonious, \ie to use as few primitives as possible to describe the scene adequately.
Defining a single evaluation metric which accommodates both objectives is not trivial.
It would require knowledge about the desired balance between reconstruction accuracy and parsimony, which depends on the target application.
We thus provide multiple metrics, each focusing on different qualities of the results:
}
{

\paragraph{Mean $|\set{M}|$}
The mean number of primitives $|\set{M}|$ is a direct indicator of the parsimony of the abstraction.
Given a certain reconstruction accuracy, we prefer a lower $|\set{M}|$.

\paragraph{Coverage}
We compute the percentage of pixels covered by at least one primitive in the image in 2D.
This indicates how much of a scene is at all represented by the predicted primitives, without taking reconstruction accuracy into account.

\paragraph{Mean OA-$L_2$}
The mean of the occlusion-aware distance of all points gives a rough indication of the reconstruction accuracy.
It is susceptible to outliers, \ie points with a large distance to any primitive.
It is hence by itself not well suited to differentiate between, for example, results with few but well-fitting primitives, and results with many but ill-fitting primitives.

\paragraph{AUC}
Based on the OA-$L_2$ distances we compute relative areas under the recall curve (AUC) for two upper bounds: ${\SI{20}{\centi\meter}}$ and ${\SI{5}{\centi\meter}}$. 
Using the larger upper bound, the AUC$_{@\SI{20}{\centi\meter}}$ metric indicates how many points are at least \emph{roughly} represented by the primitives, while the AUC$_{@\SI{5}{\centi\meter}}$ is more restrictive, only taking points into account with an OA distance of at most ${\SI{5}{\centi\meter}}$.
AUC values are less influenced by outliers than mean distances, and gauge how many points are covered by primitives within the upper bound.
Still, AUC by itself is not an adequate quality measure for abstraction.
Adding any primitive to a part of the scene not yet covered will likely increase the AUC values, as it reduces the OA distances of the surrounding points, even if said primitive is too small or ill-fitting to provide a sensible abstraction.

\paragraph{Covered OA-$L_2$}
For this reason, we additionally look at the \emph{covered} OA distances.
Using the coverage described above, we compute the mean of the OA-$L_2$ distances only for those points which are covered by a primitive.
This excludes outliers and hence indicates how well the primitives actually fit to their respective parts of the scene.

For all learning based methods, we perform five training runs using different random seeds.
Additionally, we perform five evaluation runs -- per training run, if applicable -- for every non-deterministic method, including ours.
We report mean and standard deviation for all metrics over all runs.
}

{
\begin{figure*}
		\begin{subfigure}[t]{0.99\linewidth}
			\includegraphics[width=\linewidth]{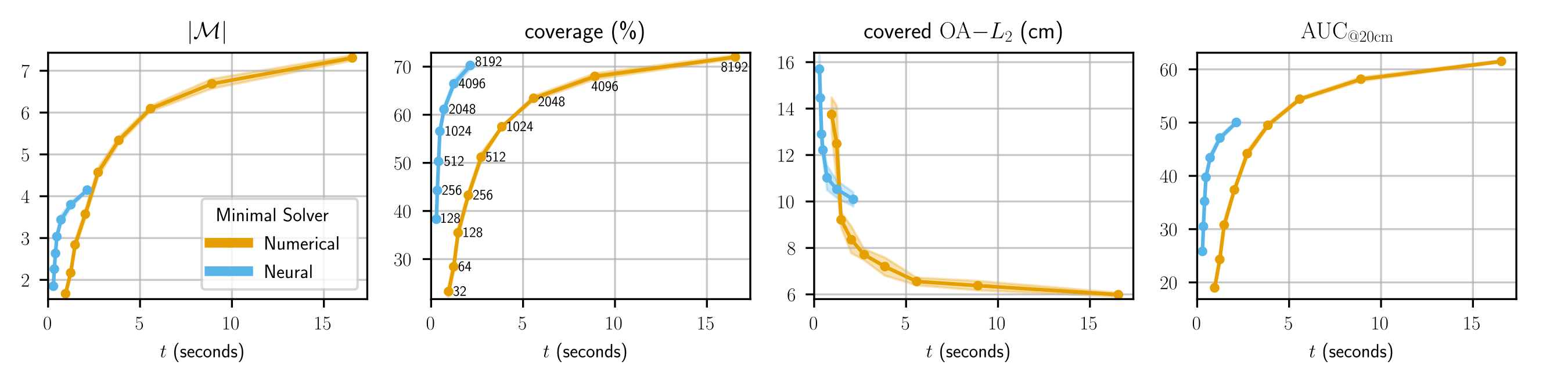}
			\caption{{
			We plot number of cuboids $|\set{M}|$, image coverage, occlusion-aware (OA) distance of covered points and AUC$_{@\SI{20}{\centi\meter}}$ as functions of computation time $t$.
			We vary the computation time via the number of cuboid hypotheses $|\set{H}|$, which we annotate in the second plot.
			}}
			\label{fig:graphs_nn_adam:time}
		\end{subfigure}
		
		\begin{subfigure}[t]{0.99\linewidth}
			\includegraphics[width=\linewidth]{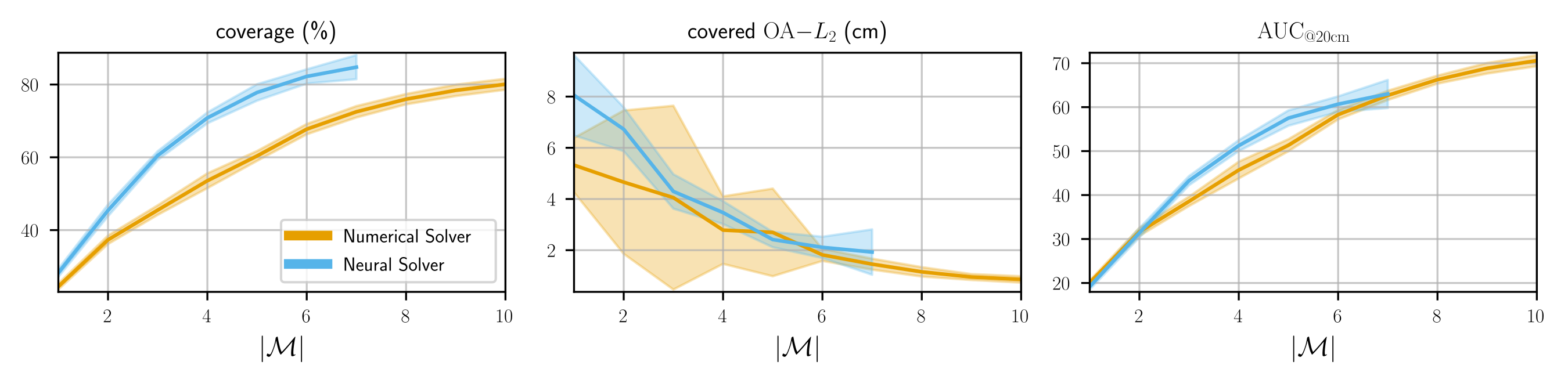}
			\caption{{
			We plot image coverage, occlusion-aware (OA) distance of covered points and AUC$_{@\SI{20}{\centi\meter}}$ as a function of the number of extracted cuboids $|\set{M}|$.
			A smaller number of cuboids indicates a more parsimonious abstraction.
			If more than $|\set{M}|$ cuboids were extracted for a scene, we only consider the first $|\set{M}|$ cuboids for evaluation.
			}}
			\label{fig:graphs_nn_adam:per_cuboid}
		\end{subfigure}
		\caption{
		{
		\textbf{Cuboid Solver:} We compare the numerical solver (Sec.~\ref{subsubsec:fitting}) against the neural network based solver (Sec.~\ref{subsubsec:fitting_neural}) for fitting cuboids to minimal sets of points on the NYU validation set.
		We look at the computational efficiency in Fig.~(a), while Fig.~(b) shows the performance of the algorithms as a function of parsimony. Shaded areas indicate the $\pm \sigma$ interval. See Sec.~\ref{subsec:eval_minimal_solver} for details. 
		}
  \vspace{-.5em}
		}
		\label{fig:graphs_nn_adam}
\end{figure*}

\subsection{Cuboid Solver}
\label{subsec:eval_minimal_solver} 

In this section, we compare the performance of our two proposed minimal solvers for cuboid fitting: numerical optimisation (Sec.~\ref{subsubsec:fitting}) and neural network based regression (Sec.~\ref{subsubsec:fitting_neural}). 
We refer to these two approaches as \emph{numerical} and \emph{neural} solver, respectively.
While we focus primarily on the results generated from depth input, our conclusions hold for RGB input as well.

\paragraph{NYU Depth v2}
As Tab.~\ref{tab:main_results} shows, the neural solver has a clear advantage \wrt computation time, as it is roughly $8-9$ times faster than the numerical solver in practice.
In Fig.~\ref{fig:graphs_nn_adam:time}, we further investigate this aspect by showing each evaluation metric as a function of computation time, which can be controlled by varying the number of sampled cuboid hypotheses $|\set{H}|$.
Given a constrained time budget of roughly one second per image or less, the neural solver outperforms the numerical solver, achieving higher coverage and AUC, and lower OA distances.
When the number of cuboid hypotheses is fixed however, the neural solver generates fewer cuboids but with a similar coverage to the numerical solver, indicating a generally more parsimonious abstraction.
Unsurprisingly, this comes at the cost of reconstruction accuracy, as the numerical solver achieves smaller errors \wrt the occlusion-aware distance.
The numerical solver reduced the mean OA distance for the covered points by $38.5\%$, and by $27.7\%$ for all points, while increasing the AUC values by roughly ten percentage points.
Judging by these numbers alone, it is not clear which method provides the better trade-off between accuracy and parsimony, and which method generates the better fitting cuboids.

In Fig.~\ref{fig:graphs_nn_adam:per_cuboid}, we show the evaluation metrics as a function of the number of cuboids $|\set{M}|$ per image.
For every step on the x-axis, we only consider images for which at least $|\set{M}|$ cuboids were found, and evaluate using only the first $|\set{M}|$ cuboids extracted from these images.
As these graphs show, the neural solver achieves a higher coverage and lower OA distance for all points, as well as similar AUC values to the numerical solver, given the same number of cuboids.
The OA distances of covered points is higher on average for the neural solver if only a small number of cuboids ($|\set{M}| \leq 2$) is considered, but on par with the numerical solver beyond that.
Overall, this indicates that while the neural solver has a lower reconstruction accuracy as it generally finds fewer cuboids, it provides more parsimonious abstractions which may be more meaningful.

Examining some qualitative examples in Fig.~\ref{fig:examples}, we observe that the neural solver is able to extract cuboids for large structures usually just as well as the numerical solver, if not better.
For instance, in \new{examples one and five} the neural solver (third column) generates one large cuboid for the beds and table, while the numerical solver splits these objects into multiple, more cluttered cuboids.
In addition, the neural solver abstracts walls and floors with few large cuboids that are well-aligned with the scene, while the numerical solver often produces multiple, partially overlapping and askew cuboids instead.
Conversely, the numerical solver performs better at representing less prominent structures and objects.
For example, the smaller chests of drawers in rows one and five are represented by cuboids with the numerical solver, while the neural solver (first column) missed them.

\begin{table*}
\setlength\tabcolsep{.37em}
\renewcommand{\arraystretch}{1.2} 
	\begin{center}
	\begin{tabular}{|lll|cc|ccc|ccc|cc|cc|cc|cc|}
	\cline{4-19}
	\multicolumn{3}{c|}{} &
	\multicolumn{8}{c|}{} &
	\multicolumn{8}{c|}{occlusion-aware $L_2$-distance} \\
	\multicolumn{3}{c|}{} & 
	mean  & \multirow{ 2}{*}{$\downarrow$} &
	\multicolumn{2}{c}{{\# primitives}} &\multirow{ 2}{*}{$\downarrow$}  &
	\multicolumn{2}{c}{image area} & \multirow{ 2}{*}{$\uparrow$} &
	\multicolumn{4}{c|}{mean (cm) $\downarrow$} &
	\multicolumn{4}{c|}{AUC (\%) $\uparrow$} \\
	
	\multicolumn{3}{c|}{} & 
	time (s) & &
	\multicolumn{2}{c}{$|\set{M}|$} & &
	\multicolumn{2}{c}{covered (\%)} & &
	\multicolumn{2}{c|}{covered points}      & 
	\multicolumn{2}{c|}{all points} &
	\multicolumn{2}{c|}{$@\SI{20}{\centi\meter}$} & 
	\multicolumn{2}{c|}{${@\SI{5}{\centi\meter}}$}  \\
	
	\hline
	\multicolumn{19}{|c|}{\small{RGB input}} \\
 
	\multicolumn{3}{|l|}{SQ-Parsing RGB~\cite{paschalidou2019superquadrics}}                    
	& \multicolumn{2}{c|}{\best{0.009}}
	& \third{7.6} & \multicolumn{2}{c|}{$\scriptstyle \pm 1.14 $ }
	& \second{ 85.3} & \multicolumn{2}{c|}{$\scriptstyle \pm 10.52 $  }
	& $ 133.9 $ & $\scriptstyle \pm 36.17 $  
	& $ 133.2 $ & $\scriptstyle \pm 36.51 $  
	& $ 2.5 $ & $\scriptstyle \pm 2.29 $  
	& $ 0.2 $ & $\scriptstyle \pm 0.16 $  \\
	\multicolumn{3}{|l|}{SQ-Parsing~\cite{paschalidou2019superquadrics} + BTS~\cite{lee2019big}}  
	& \multicolumn{2}{c|}{\second{0.89}}
	& $ 10.4 $ & \multicolumn{2}{c|}{$\scriptstyle \pm 1.51 $}
	& \best{ 91.7 } & \multicolumn{2}{c|}{$\scriptstyle \pm 1.09 $}
	& \third{68.7} & $\scriptstyle \pm 2.16 $
	& \third{68.9} & $\scriptstyle \pm 2.47 $
	& \third{11.0} & $\scriptstyle \pm 0.73 $
	& \third{1.0} & $\scriptstyle \pm 0.14 $ \\
	\parbox[t]{2mm}{\multirow{3}{*}{\rotatebox[origin=c]{90}{\textbf{Ours}}}} &
	\textbf{solver:} & & & & & & & & & & & & & & & & &\\
	 & numerical & 
	 & \multicolumn{2}{c|}{$8.16$}
 	& \second{7.1} & \multicolumn{2}{c|}{$\scriptstyle \pm 0.06 $}
	& \third{65.0} & \multicolumn{2}{c|}{$\scriptstyle \pm 0.31 $}
	& \best{ 23.6 } & $\scriptstyle \pm 0.21 $
	& \best{ 42.1 } & $\scriptstyle \pm 0.65 $
	& \best{ 28.0 } & $\scriptstyle \pm 0.14 $
	& \best{ 7.6 } & $\scriptstyle \pm 0.06 $ \\
	 & neural & 
	 & \multicolumn{2}{c|}{	\third{0.99}}
	 & \best{ 4.2 } & \multicolumn{2}{c|}{$\scriptstyle \pm 0.04 $}
	& {$ 63.7 $} & \multicolumn{2}{c|}{$\scriptstyle \pm 0.45 $}
	& \second{26.1} & $\scriptstyle \pm 0.27 $
	& \second{51.3} & $\scriptstyle \pm 0.83 $
	& \second{24.1} & $\scriptstyle \pm 0.16 $
	& \second{6.6} & $\scriptstyle \pm 0.07 $ \\

	\hline
	\multicolumn{19}{|c|}{\small{depth input}} \\
	\multicolumn{3}{|l|}{SQ-Parsing~\cite{paschalidou2019superquadrics}}                          
	& \multicolumn{2}{c|}{\best{0.003}}
	& $ 10.4 $ & \multicolumn{2}{c|}{$\scriptstyle \pm 1.51 $}
	& \second{93.9} & \multicolumn{2}{c|}{$\scriptstyle \pm 1.29 $}
	& $ 37.9 $ & $\scriptstyle \pm 3.28 $
	& \second{37.7} & $\scriptstyle \pm 3.42 $
	& $ 20.3 $ & $\scriptstyle \pm 1.84 $
	& $ 2.2 $ & $\scriptstyle \pm 0.33 $ \\
	\multicolumn{3}{|l|}{Sequential RANSAC~\cite{vincent2001seqransac}}       
	& \multicolumn{2}{c|}{$9.30$ }
	& $ 7.7 $ & \multicolumn{2}{c|}{$\scriptstyle \pm 0.02 $}
	& \best{99.8} & \multicolumn{2}{c|}{$\scriptstyle \pm {0.01} $}
	& $ 139.3 $ & $\scriptstyle \pm 0.54 $
	& $ 139.1 $ & $\scriptstyle \pm {0.53} $
	& $ 6.6 $ & $\scriptstyle \pm {0.09} $
	& $ 2.4 $ & $\scriptstyle \pm {0.04} $ \\
	\multicolumn{3}{|l|}{Sequential RANSAC~\cite{vincent2001seqransac} + OAI}                           
	& \multicolumn{2}{c|}{\third{2.05}}
	& \best{1.5} & \multicolumn{2}{c|}{$\scriptstyle \pm {0.01} $}
	& $ 44.0 $ & \multicolumn{2}{c|}{$\scriptstyle \pm 0.45 $}
	& \third{16.5} & $\scriptstyle \pm {0.25} $
	& $ 67.3 $ & $\scriptstyle \pm 1.01 $
	& \third{28.3} & $\scriptstyle \pm 0.19 $
	& \third{12.0} & $\scriptstyle \pm 0.14 $ \\
	\multicolumn{3}{|l|}{CONSAC~\cite{KluBra2020}} 
	& \multicolumn{2}{c|}{$10.23$}
	& $ 8.4 $ & \multicolumn{2}{c|}{$\scriptstyle \pm 0.16 $}
	& \best{99.8} & \multicolumn{2}{c|}{$\scriptstyle \pm 0.02 $}
	& $ 121.6 $ & $\scriptstyle \pm 4.04 $
	& $ 121.5 $ & $\scriptstyle \pm 4.04 $
	& $ 7.7 $ & $\scriptstyle \pm 0.34 $
	& $ 2.7 $ & $\scriptstyle \pm 0.13 $ \\
	\parbox[t]{2mm}{\multirow{3}{*}{\rotatebox[origin=c]{90}{\textbf{Ours}}}} &
	\textbf{solver:} & & & & & & & & & & & & & & & & & \\
	 & numerical & 
	 & \multicolumn{2}{c|}{$7.63$}
	& \third{6.8} & \multicolumn{2}{c|}{$\scriptstyle \pm 0.30 $}
	& \third{69.7} & \multicolumn{2}{c|}{$\scriptstyle \pm 0.86 $}
	& \best{6.4} & $\scriptstyle \pm {0.25} $
	& \best{27.4} & $\scriptstyle \pm 1.20 $
	& \best{60.1} & $\scriptstyle \pm 0.92 $
	& \best{31.8} & $\scriptstyle \pm 0.71 $ \\
	 & neural & 
	 & \multicolumn{2}{c|}{\second{0.87}}
	& \second{3.9} & \multicolumn{2}{c|}{$\scriptstyle \pm 0.04 $}
	& $ 69.3 $ & \multicolumn{2}{c|}{$\scriptstyle \pm 0.39 $}
	& \second{10.4} & $\scriptstyle \pm 0.27 $
	& \third{37.9} & $\scriptstyle \pm 0.60 $
	& \second{49.4} & $\scriptstyle \pm 0.23 $
	& \second{22.1} & $\scriptstyle \pm 0.15 $ \\
	\hline
	\end{tabular}
    \end{center}
	\caption{
	\textbf{Quantitative Results on NYU:} 
	We evaluate on NYU Depth v2~\cite{silberman2012indoor}, for both RGB and depth inputs. We compare our method against variants of SQ-Parsing~\cite{paschalidou2019superquadrics}, Sequential RANSAC~\cite{vincent2001seqransac} 
	{ and CONSAC~\cite{KluBra2020}. 
	We present the computation time per scene, the mean number of primitives $|\set{M}|$, scene coverage in terms of image area, mean occlusion-aware (OA) $L_2$ distance of covered points and of all points, as well AUC values for two upper bounds of the OA $L_2$ distance (cf. Sec.~\ref{subsec:metrics}). 
        For each metric, arrows indicate \emph{higher is better} ($\uparrow$) or \emph{lower is better} ($\downarrow$).
	We highlight the best results in \best{\text{\textbf{green}}}, the second best in \second{\text{light green}} and the third best in \third{\text{\textit{blue}}}. 
	See Sec.~\ref{subsec:baseline_comparison} for a discussion of the results.} }
	\label{tab:main_results}
\end{table*}

\new{
\paragraph{Synthetic Metropolis Homographies}
\label{sec:smh_results}
We provide quantitative results on the SMH dataset in Tab.~\ref{tab:smh_results} and observe results similar to those on NYU:
The neural solver provides more parsimonious abstractions at the cost of coverage and reconstruction accuracy.
Overall, the occlusion-aware distances on SMH are higher than on NYU, because SMH is an outdoor dataset and thus covers a larger range of distances.
As qualitative examples in Fig.~\ref{fig:smh_examples} show, the neural solver extracts cuboids for these synthetic outdoor scenes often just as well (first and second example) or even better than (third example) the numerical solver, but tends to ignore structures in some cases (fourth example).
}

\begin{table*}
\renewcommand{\arraystretch}{1.1} 
	\begin{center}
	\begin{tabular}{|l|ccc|ccc|cc|cc|cc|cc|}
	\cline{2-15}
	  \multicolumn{1}{c|}{} & 
	\multicolumn{2}{c}{{\# primitives}} &\multirow{ 2}{*}{$\downarrow$}  &
	\multicolumn{2}{c}{image area} & \multirow{ 2}{*}{$\uparrow$} &
	\multicolumn{4}{c|}{OA-$L_2$ mean (m) $\downarrow$} &
	\multicolumn{4}{c|}{OA-$L_2$ AUC (\%) $\uparrow$} \\
        \multicolumn{1}{c|}{} &
	\multicolumn{2}{c}{$|\set{M}|$} & &
	\multicolumn{2}{c}{covered (\%)} & &
	\multicolumn{2}{c|}{covered points}      & 
	\multicolumn{2}{c|}{all points} &
	\multicolumn{2}{c|}{$@\SI{50}{\centi\meter}$} & 
	\multicolumn{2}{c|}{${@\SI{10}{\centi\meter}}$}  \\
	\hline
 
	\textbf{Ours} with numerical solver
	& $ 3.5 $ & \multicolumn{2}{c|}{$\scriptstyle \pm 0.20 $}
	& $ \mathbf{58.8} $ & \multicolumn{2}{c|}{$\scriptstyle \pm 1.17 $}
	& $ \mathbf{0.29} $ & $\scriptstyle \pm 0.04 $
	& $ \mathbf{6.58} $ & $\scriptstyle \pm 0.38 $
	& $ \mathbf{65.2} $ & $\scriptstyle \pm 1.62 $
	& $ \mathbf{34.9} $ & $\scriptstyle \pm 1.29 $ \\

 	\textbf{Ours} with neural solver
	& $ \mathbf{2.9} $ & \multicolumn{2}{c|}{$\scriptstyle \pm 0.02 $}
	& $ 49.4 $ & \multicolumn{2}{c|}{$\scriptstyle \pm 0.24 $}
	& $ 2.47 $ & $\scriptstyle \pm 0.25 $
	& $ 13.82 $ & $\scriptstyle \pm 0.49 $
	& $ 50.6 $ & $\scriptstyle \pm 0.23 $
	& $ 21.6 $ & $\scriptstyle \pm 0.15 $ \\
	    \hline

	\end{tabular}
    \end{center}
	\caption{\new{
	\textbf{Quantitative Results on SMH:} 
	We evaluate on the Synthetic Metropolis Homography dataset~\cite{kluger2024parsac} using depth inputs
 and compare our numerical and neural cuboid solvers.
	See Sec.~\ref{subsec:eval_minimal_solver} for a discussion of the results.} 	
 \vspace{-1em}}
 \label{tab:smh_results}
\end{table*}	

\subsection{Baseline Comparisons}
\label{subsec:baseline_comparison}
We compare our approach -- using either the neural or the numerical solver -- against SQ-Parsing~\cite{paschalidou2019superquadrics} for RGB and depth input, as well as Sequential RANSAC~\cite{vincent2001seqransac} and CONSAC~\cite{KluBra2020} for depth input only.
Tab.~\ref{tab:main_results} provides quantitative results on the test set of NYU Depth v2~\cite{silberman2012indoor}.
We additionally provide qualitative examples in Fig.~\ref{fig:examples}.
We compare the results of the neural solver and the numerical solver for our approach in Sec.~\ref{subsec:eval_minimal_solver}.

\paragraph{RGB Input}
As Tab.~\ref{tab:main_results} shows, SQ-Parsing RGB is by far the fastest method, being roughly 100 times faster than our approach with the neural solver. 
However, it also achieves very low reconstruction accuracy, with mean OA distances of more than ${\SI{1.3}{\meter}}$ and single-digit AUC values.
{Given that the average depth in the NYU dataset is just ${\SI{2.7}{\meter}}$},  this indicates that the predicted superquadrics do not provide any sensible scene abstraction.
SQ-Parsing + BTS reduces the OA distances by roughly $48\%$ and achieves the highest image coverage ($91.7\%$), but requires the largest number of primitives ($10.4$) on average as well.
In comparison, our method with the neural solver reduces the OA distances of covered points by another $62\%$, and those of all points by $25\%$.
Meanwhile, it uses almost $60\%$ fewer primitives and thus produces more parsimonious representations.

\paragraph{Depth Input}
As Tab.~\ref{tab:main_results} shows, our two proposed methods consistently achieve results within the top three compared to four baselines for almost all metrics.
While the other methods may excel \wrt certain metrics, they fall short on others.
SQ-Parsing is the fastest method by a large margin, achieves a high image coverage and consequently a relatively low OA distance \wrt all points.
However, its abstractions are the least parsimonious with $10.4$ superquadrics on average, and the high OA distance for covered points -- which is $5.9$ times the best result -- indicates that they do not fit the scene very well.
Sequential RANSAC~\cite{vincent2001seqransac} without occlusion-aware inlier counting (OAI) achieves a very high image coverage and better parsimony than SQ-Parsing, but performs significantly worse in all other respects.
With OAI, the accuracy of Sequential RANSAC for covered points improves drastically and the number extracted cuboids shrinks to the lowest value of all ($1.5$), but this also results in a very low coverage of just $44\%$.
CONSAC~\cite{KluBra2020} performs very similarly to Sequential RANSAC without OAI, with only slight improvements \wrt the OA distances.
In contrast to these baselines, our methods achieve the lowest OA distances for covered points, and using the numerical solver we also achieve the lowest OA distances for all points, with improvements of up to $61\%$ compared to the next best baseline.
Our methods produce very parsimonious abstractions, using just $3.9$ (neural solver) or $6.8$ (numerical solver) cuboids on average, yet achieve a high image coverage of more than $69\%$.

\begin{figure*}	

\setlength\tabcolsep{0.1em}
\renewcommand{\arraystretch}{1.} 
	\begin{center}
        \begin{tabular}{ccoooobb}
             &
             & \multicolumn{4}{c}{\cellcolor{ours}{\textbf{Ours}}} 
             & \multicolumn{2}{c}{\cellcolor{baseline}{Baselines}} \\
             &
             & \multicolumn{2}{c}{\cellcolor{ours}{Depth Input}} 
             & \multicolumn{2}{c}{\cellcolor{ours}{RGB Input}} 
             & \multicolumn{2}{c}{\cellcolor{baseline}{Depth Input}} \\
             &
             Ground Truth & 
             Neural Solver & 
             Numerical Solver & 
             Neural Solver & 
             Numerical Solver & 
             CONSAC~\cite{KluBra2020} & 
             SQ-Parsing~\cite{paschalidou2019superquadrics} \\
                \parbox[t]{3mm}{{\rotatebox[origin=l]{90}{RGB Image}}} &
			\includegraphics[width=0.136\linewidth]{fig/examples_new/rgb/final_0388_image.png} &
			\includegraphics[width=0.136\linewidth]{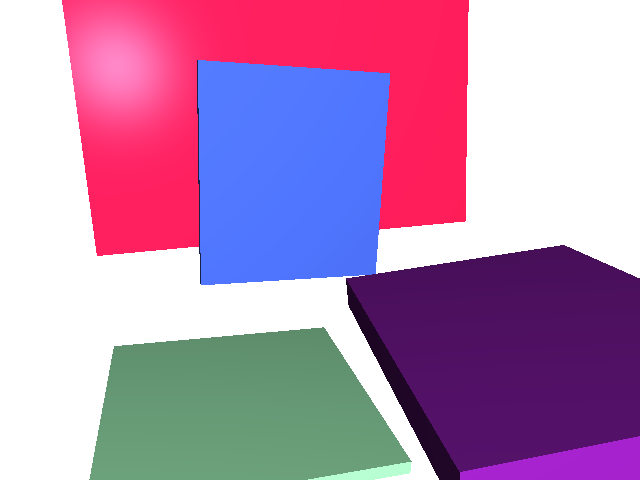} &
			\includegraphics[width=0.136\linewidth]{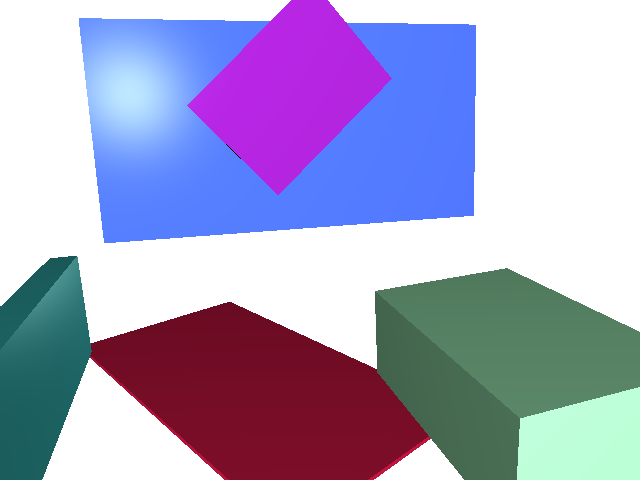} &
			\includegraphics[width=0.136\linewidth]{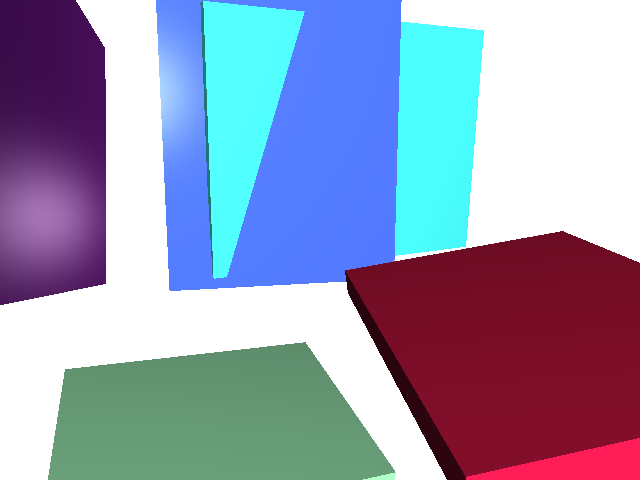} &			
			\includegraphics[width=0.136\linewidth]{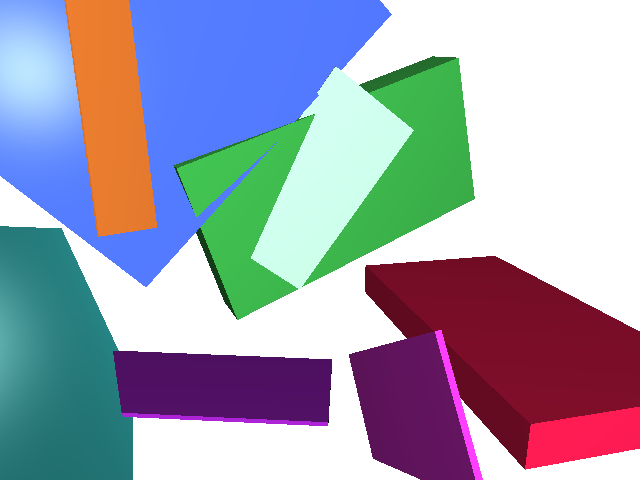} &
                \includegraphics[width=0.136\linewidth]{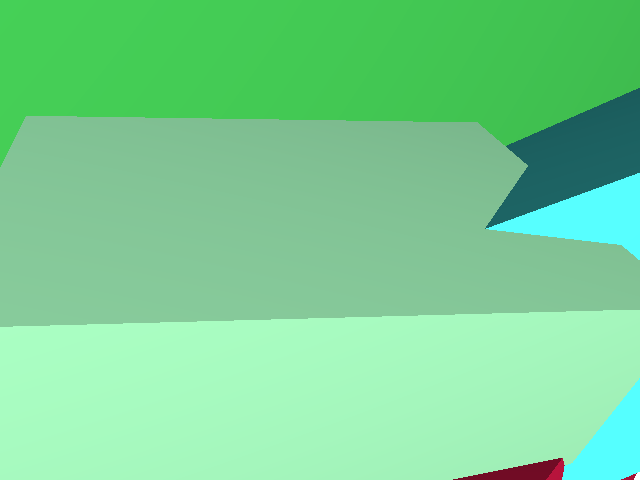} &
			\includegraphics[width=0.136\linewidth]{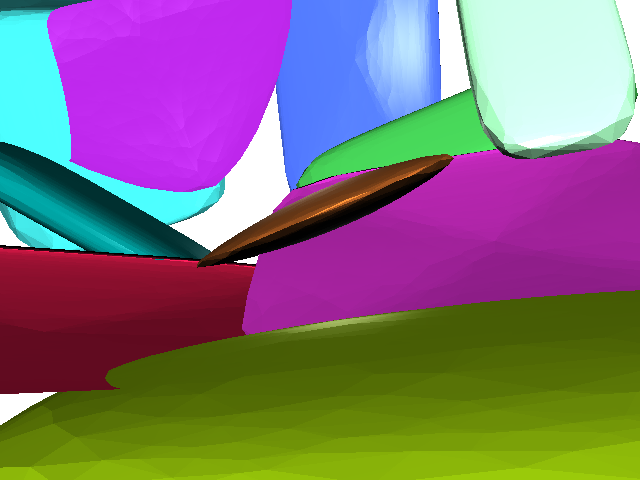} \\
   
                \parbox[t]{3mm}{{\rotatebox[origin=l]{90}{\hspace{1em} Depth}}} &
			\includegraphics[width=0.136\linewidth]{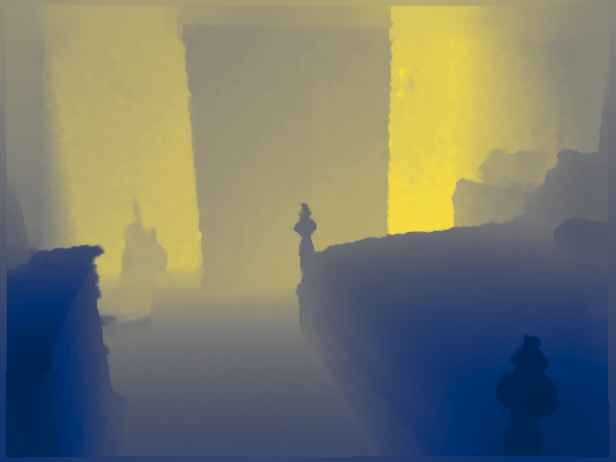} &
			\includegraphics[width=0.136\linewidth]{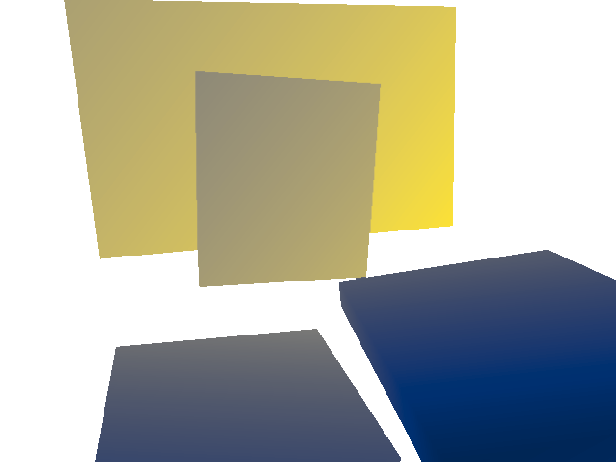} &
			\includegraphics[width=0.136\linewidth]{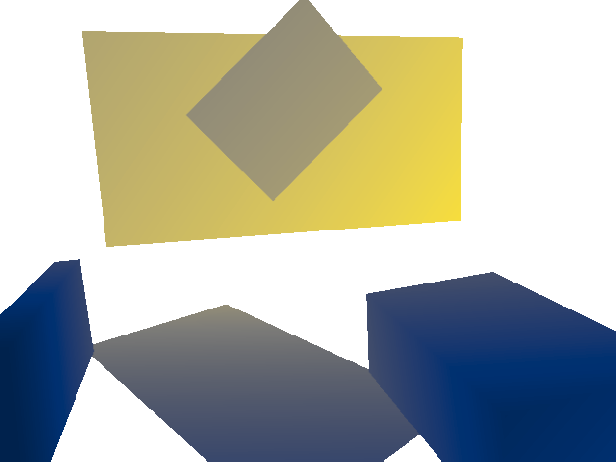} &
			\includegraphics[width=0.136\linewidth]{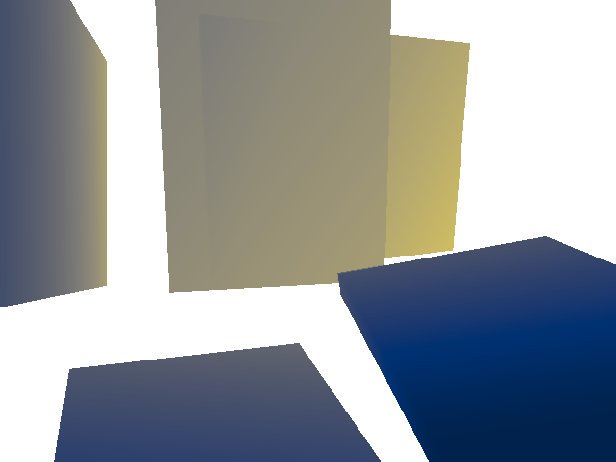} &			
			\includegraphics[width=0.136\linewidth]{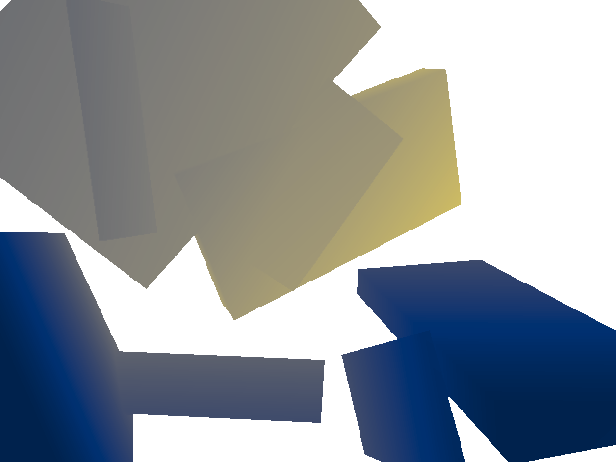} &
                \includegraphics[width=0.136\linewidth]{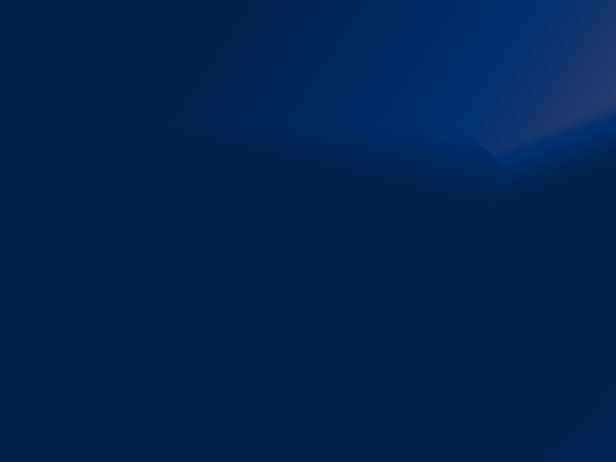} &
			\includegraphics[width=0.136\linewidth]{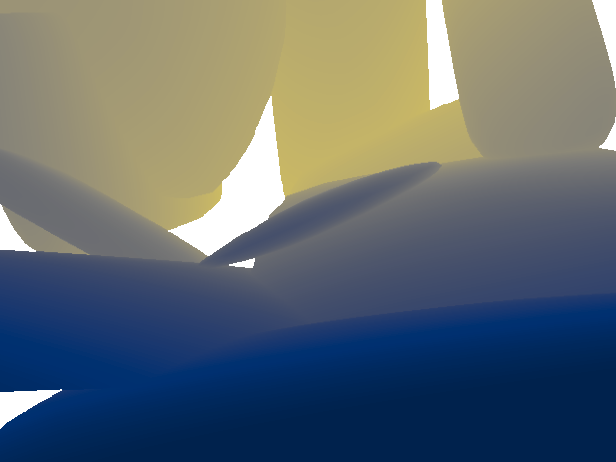} \\

                &
			\includegraphics[width=0.136\linewidth]{fig/examples_new/rgb/final_0160_image.png} &
			\includegraphics[width=0.136\linewidth]{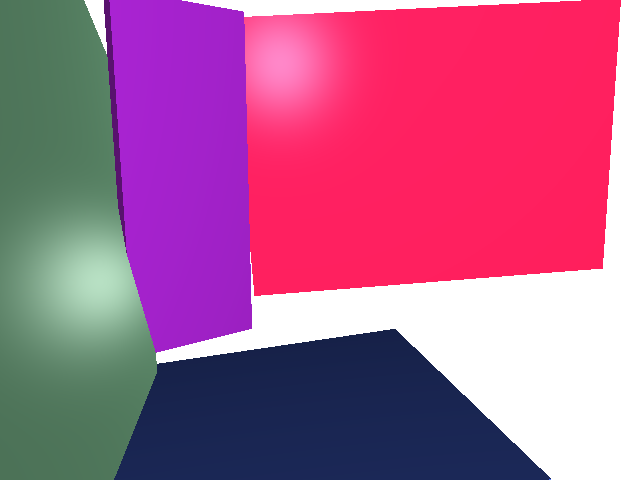} &
			\includegraphics[width=0.136\linewidth]{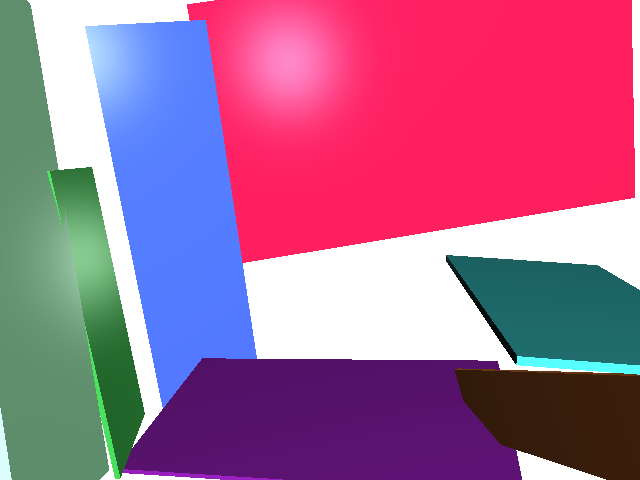} &
			\includegraphics[width=0.136\linewidth]{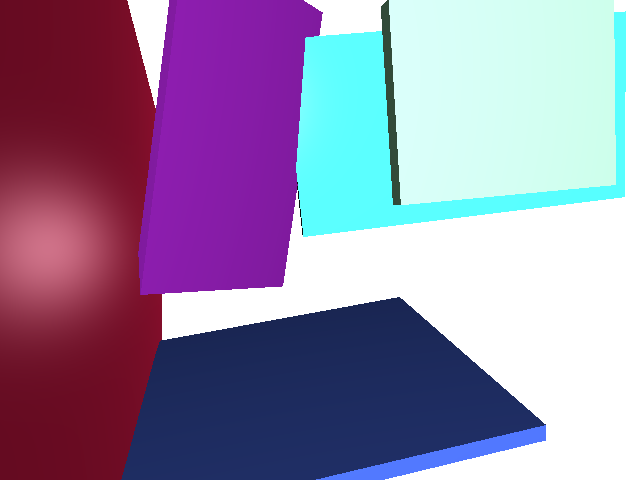} &
			\includegraphics[width=0.136\linewidth]{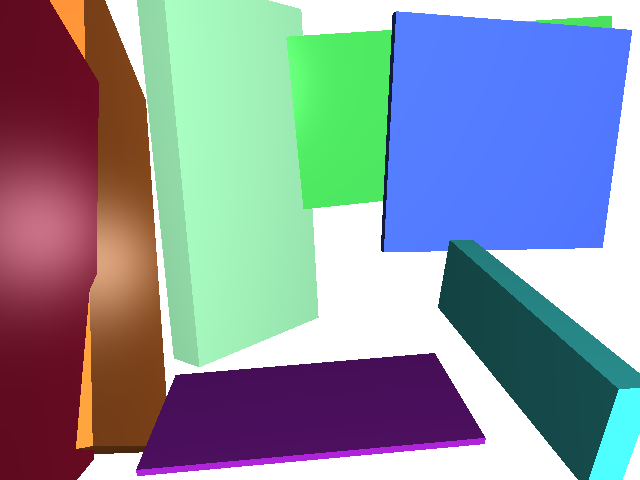} &
			\includegraphics[width=0.136\linewidth]{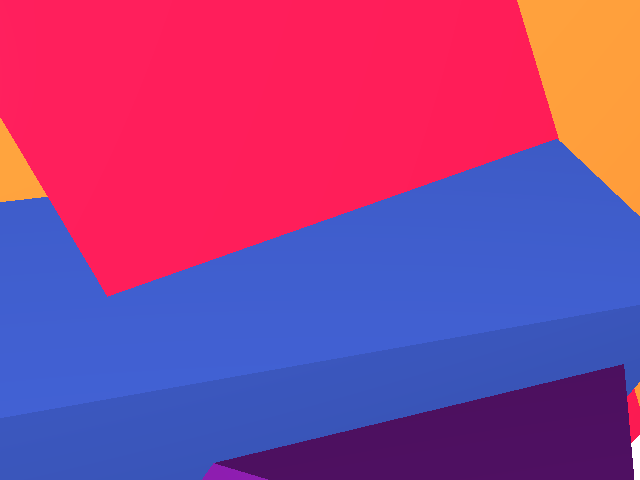} &
			\includegraphics[width=0.136\linewidth]{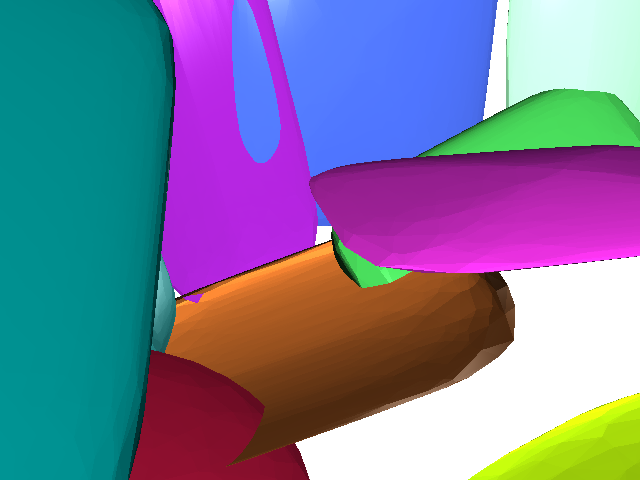} \\
   
                &
			\includegraphics[width=0.136\linewidth]{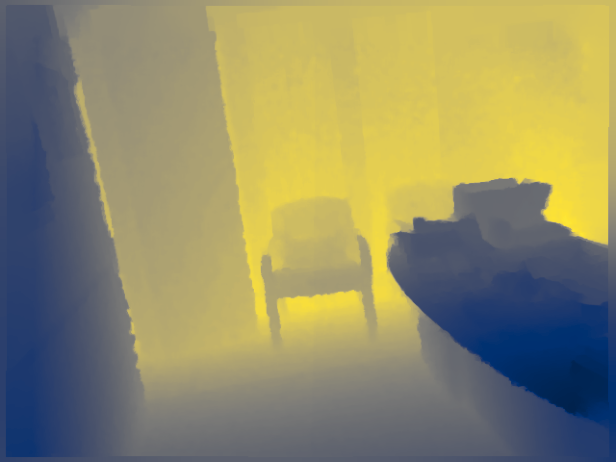} &
			\includegraphics[width=0.136\linewidth]{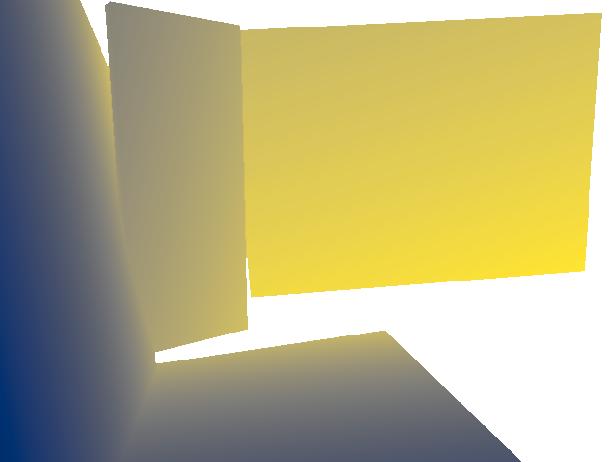} &
			\includegraphics[width=0.136\linewidth]{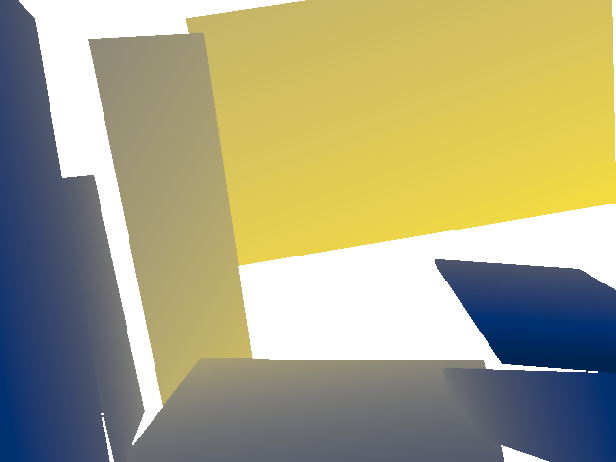} &
			\includegraphics[width=0.136\linewidth]{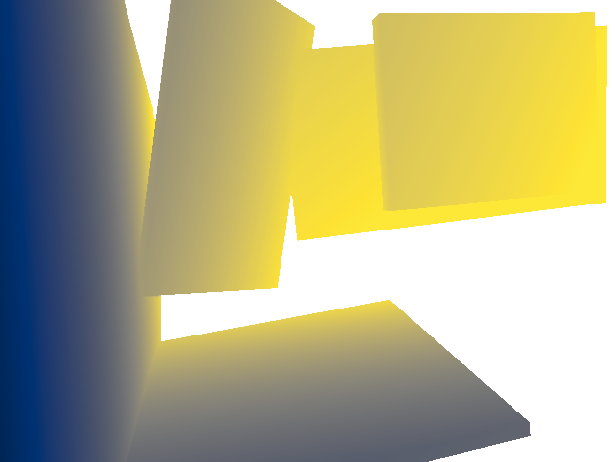} &
			\includegraphics[width=0.136\linewidth]{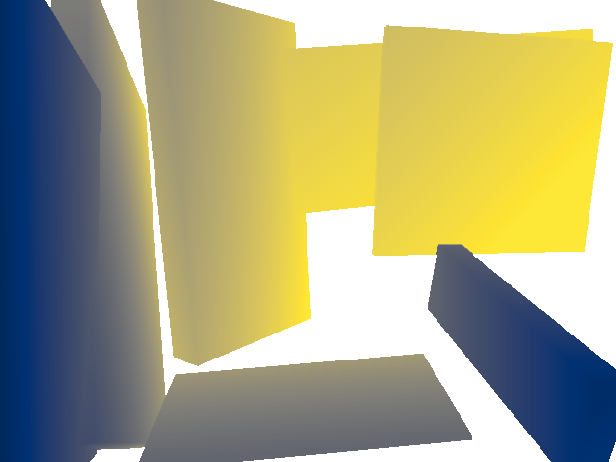} &
			\includegraphics[width=0.136\linewidth]{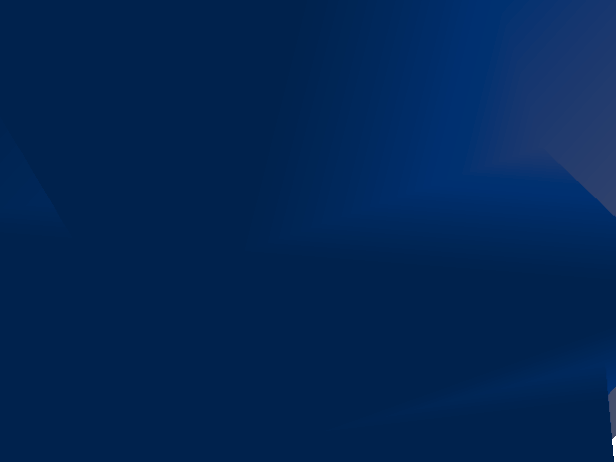} &
			\includegraphics[width=0.136\linewidth]{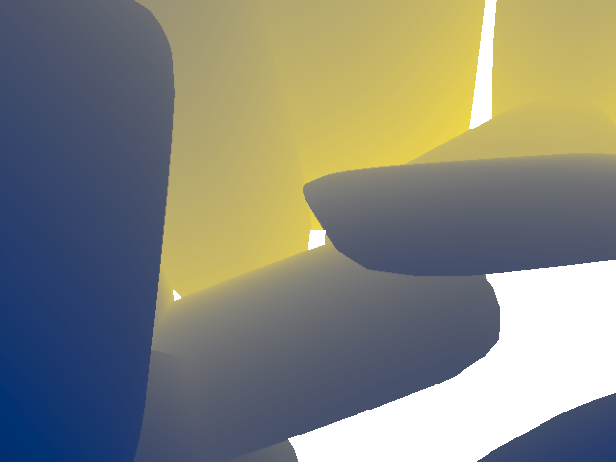} \\

                &
			\includegraphics[width=0.136\linewidth]{fig/examples_new/rgb/final_0215_image.png} &
			\includegraphics[width=0.136\linewidth]{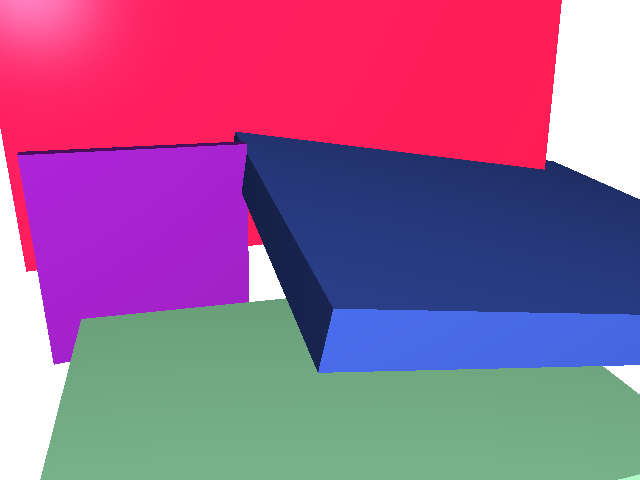} &
			\includegraphics[width=0.136\linewidth]{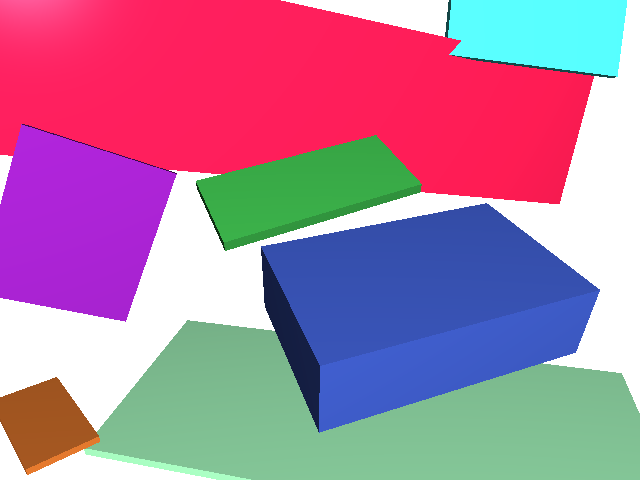} &
			\includegraphics[width=0.136\linewidth]{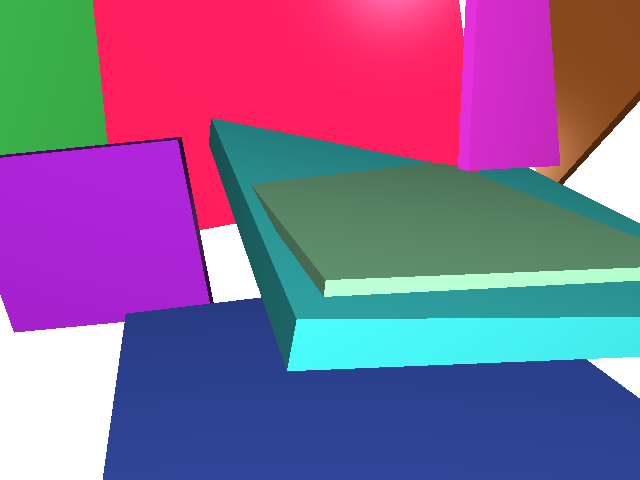} &
			\includegraphics[width=0.136\linewidth]{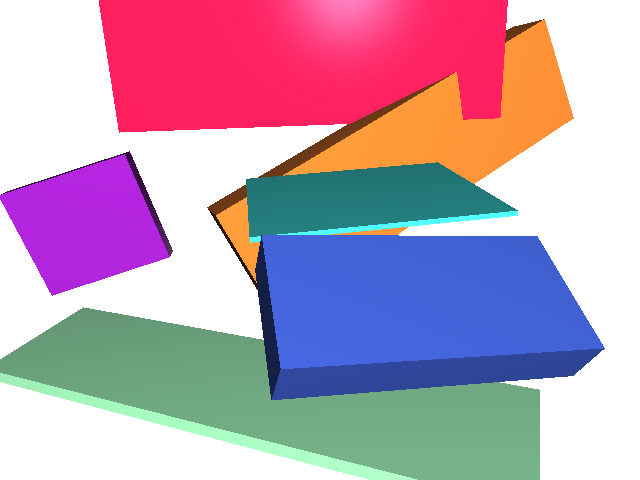} &
			\includegraphics[width=0.136\linewidth]{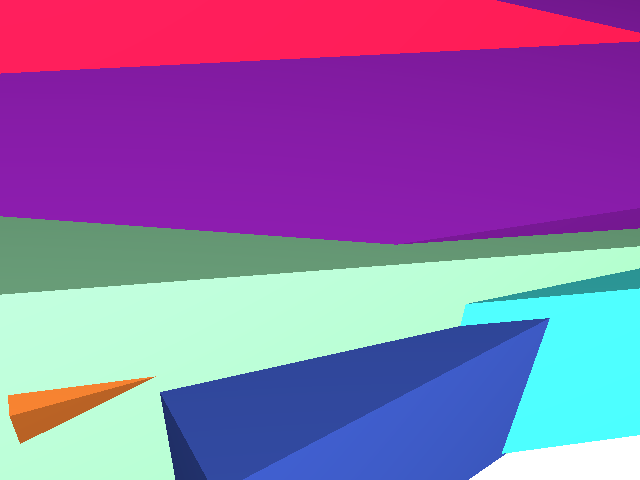} &
			\includegraphics[width=0.136\linewidth]{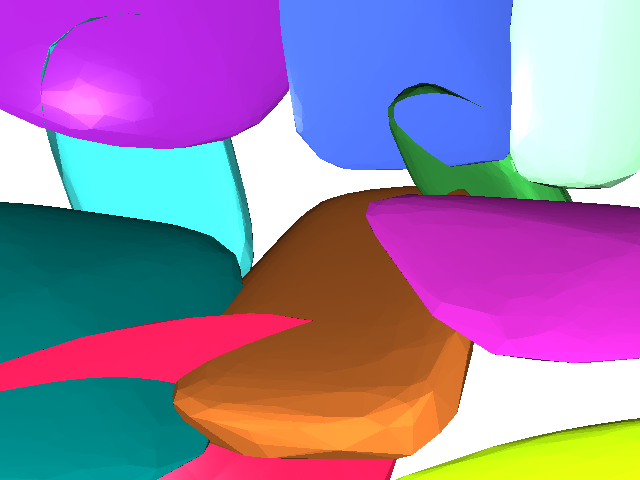} \\
   
                &
			\includegraphics[width=0.136\linewidth]{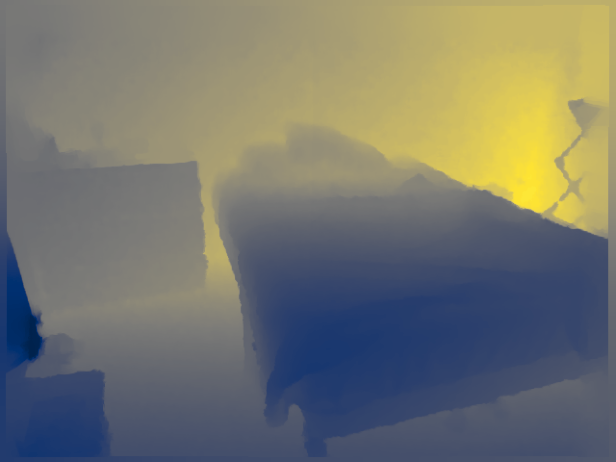} &
			\includegraphics[width=0.136\linewidth]{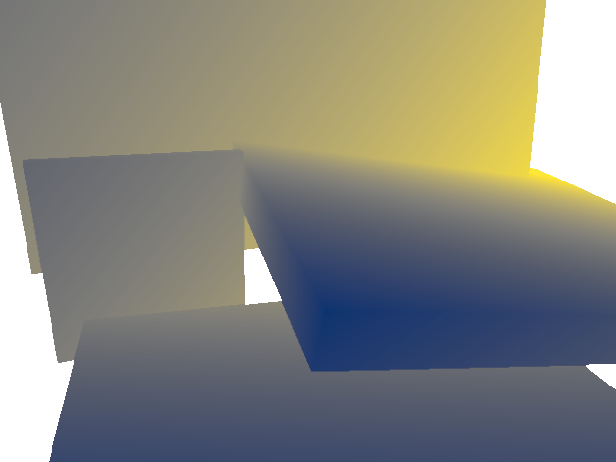} &
			\includegraphics[width=0.136\linewidth]{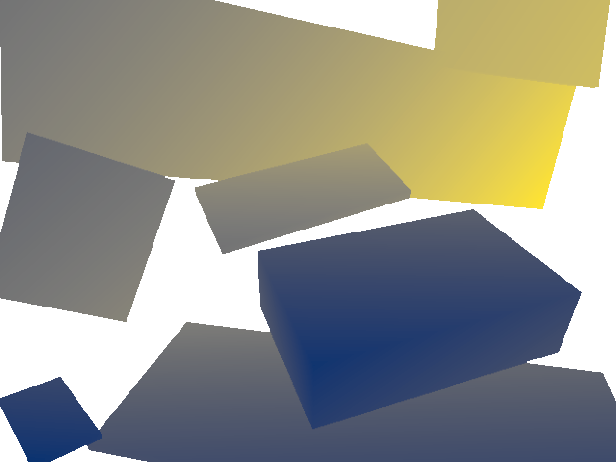} &
			\includegraphics[width=0.136\linewidth]{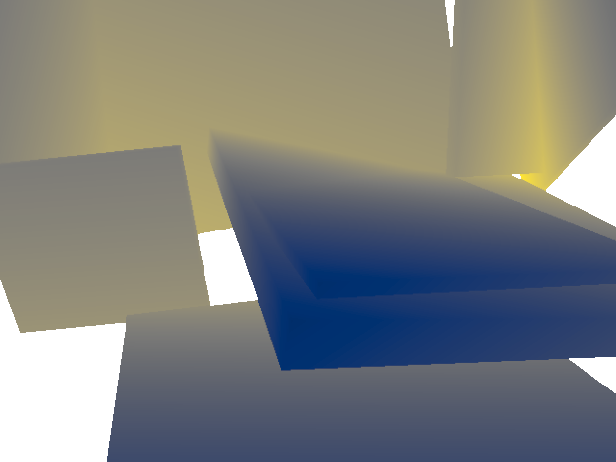} &
			\includegraphics[width=0.136\linewidth]{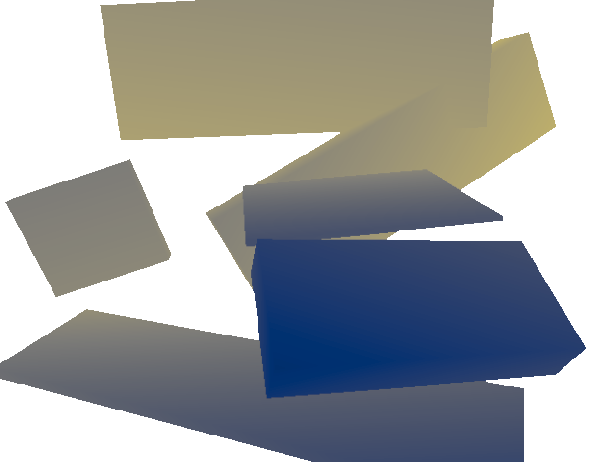} &
			\includegraphics[width=0.136\linewidth]{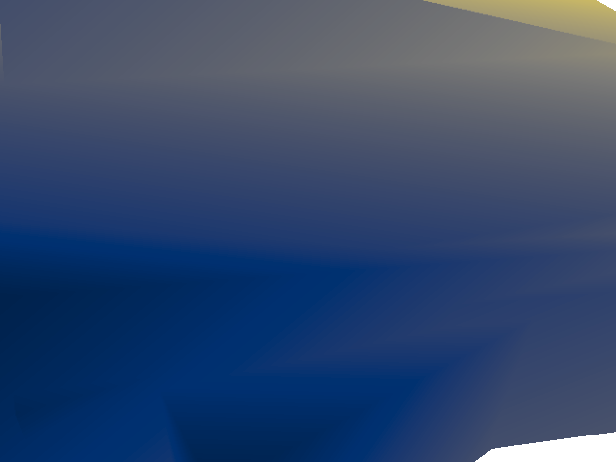} &
			\includegraphics[width=0.136\linewidth]{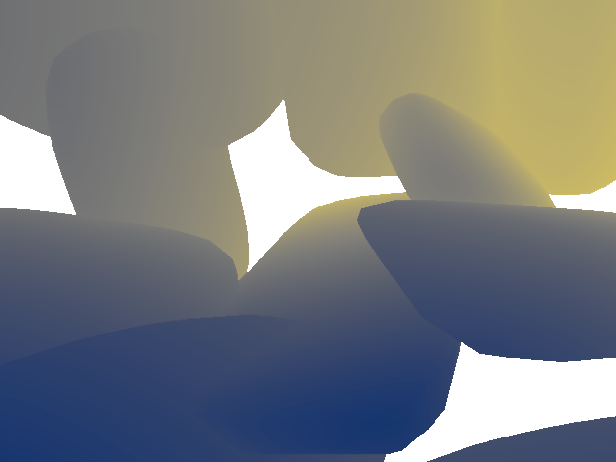} \\

                &
			\includegraphics[width=0.136\linewidth]{fig/examples_new/rgb/final_0290_image.png} &
			\includegraphics[width=0.136\linewidth]{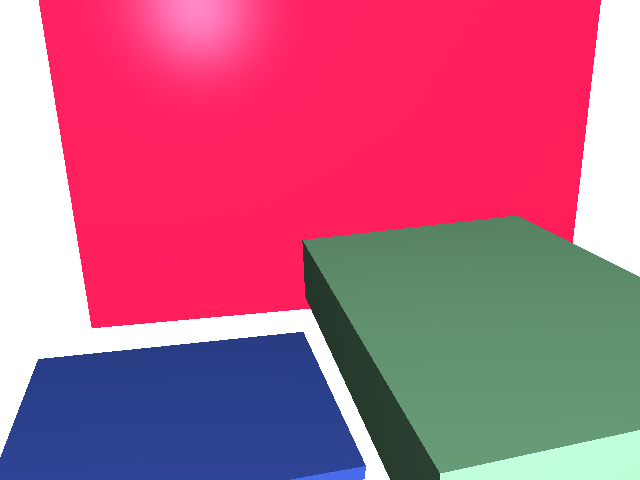} &
			\includegraphics[width=0.136\linewidth]{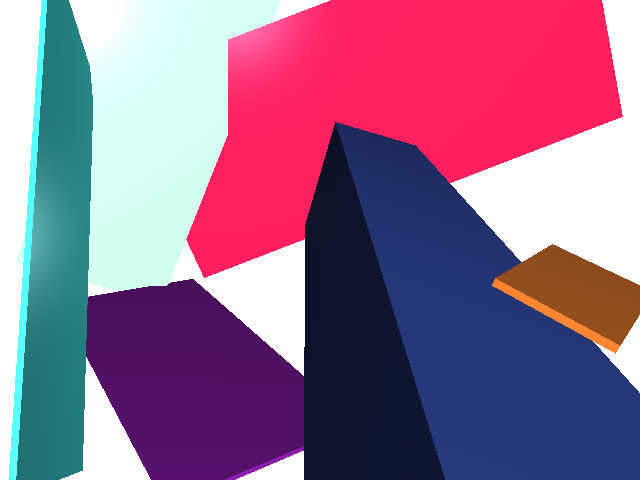} &
			\includegraphics[width=0.136\linewidth]{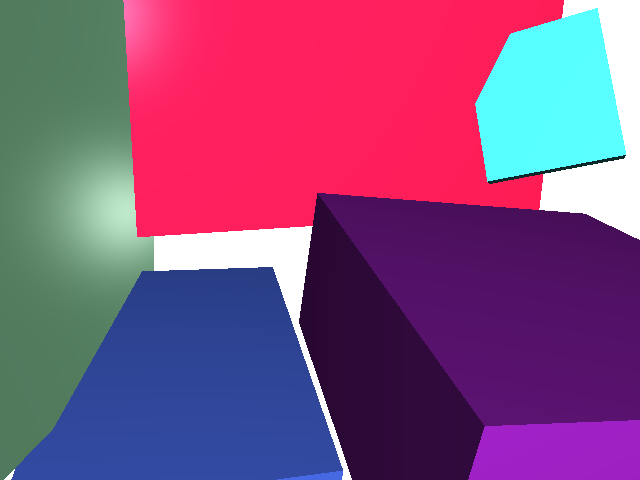} &
			\includegraphics[width=0.136\linewidth]{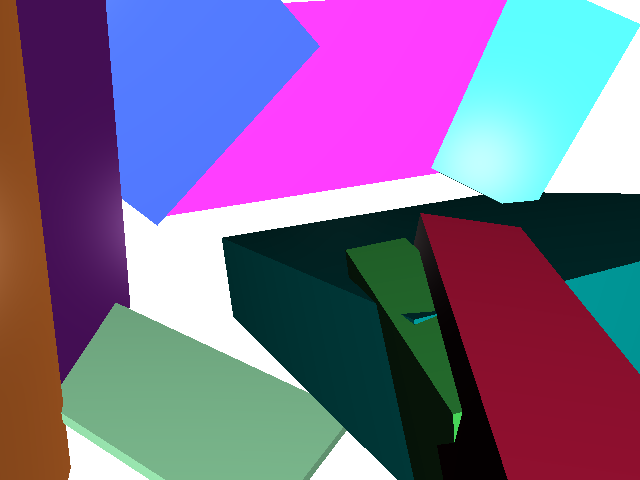} &
			\includegraphics[width=0.136\linewidth]{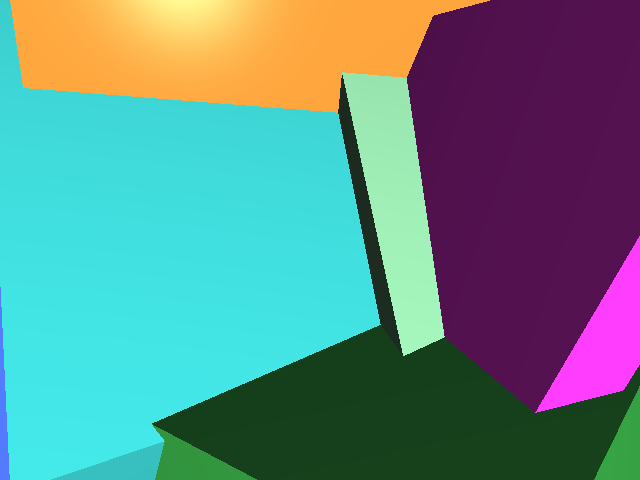} &
			\includegraphics[width=0.136\linewidth]{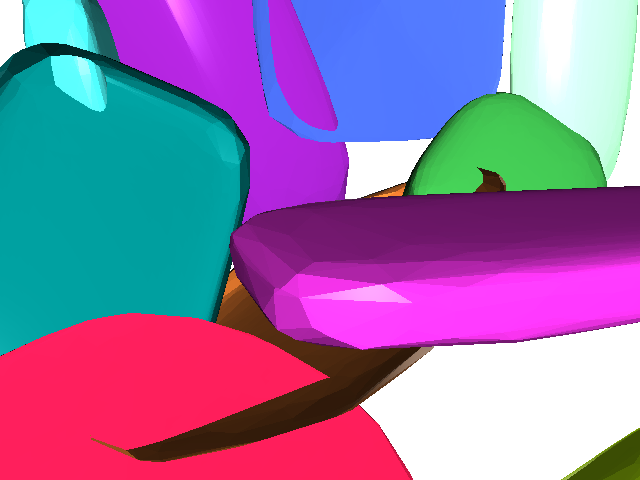} \\
   
                &
			\includegraphics[width=0.136\linewidth]{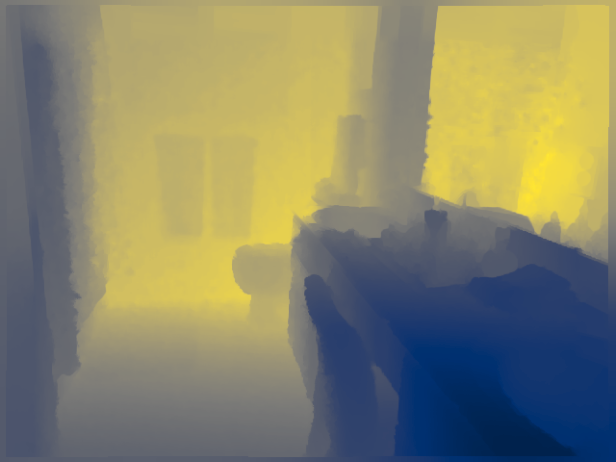} &
			\includegraphics[width=0.136\linewidth]{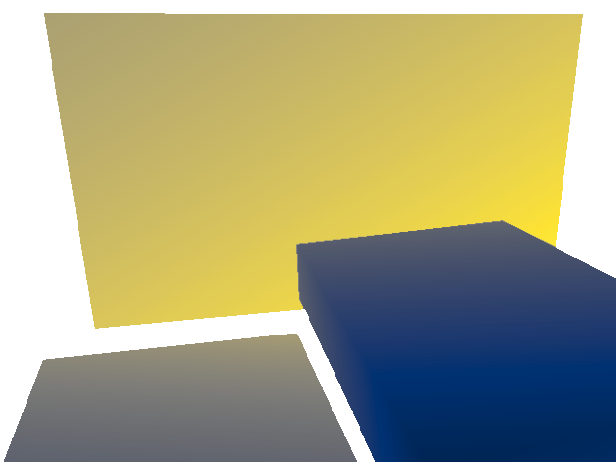} &
			\includegraphics[width=0.136\linewidth]{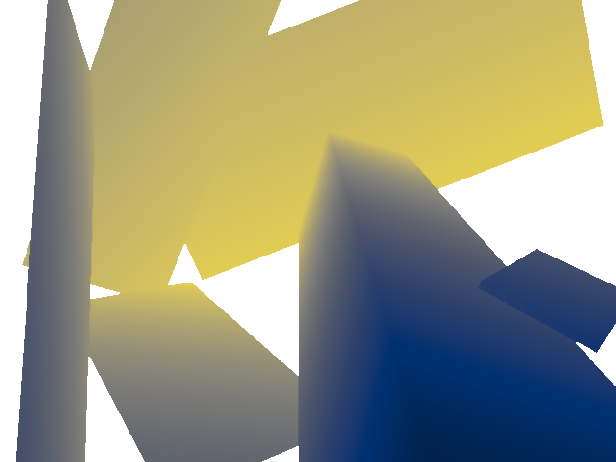} &
			\includegraphics[width=0.136\linewidth]{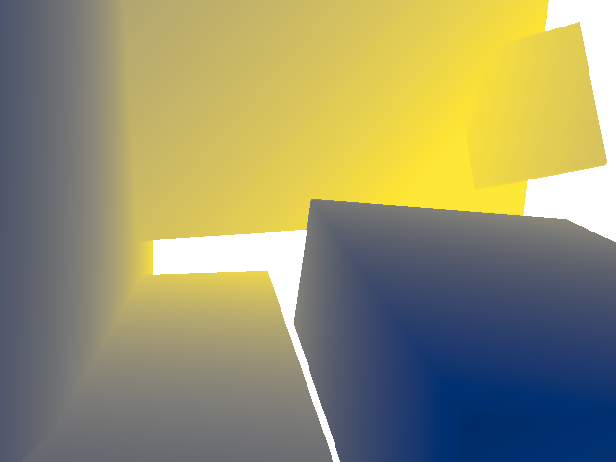} &
			\includegraphics[width=0.136\linewidth]{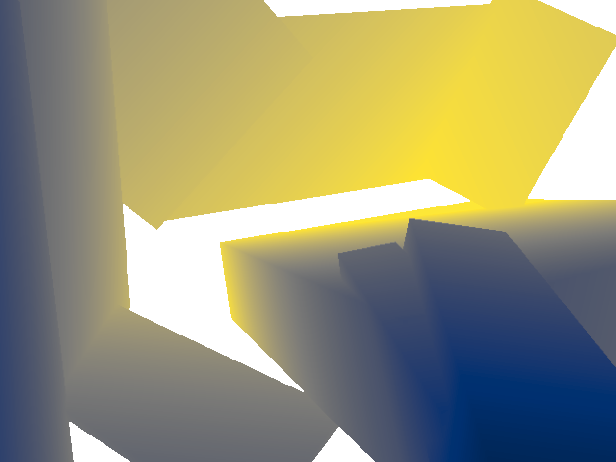} &
			\includegraphics[width=0.136\linewidth]{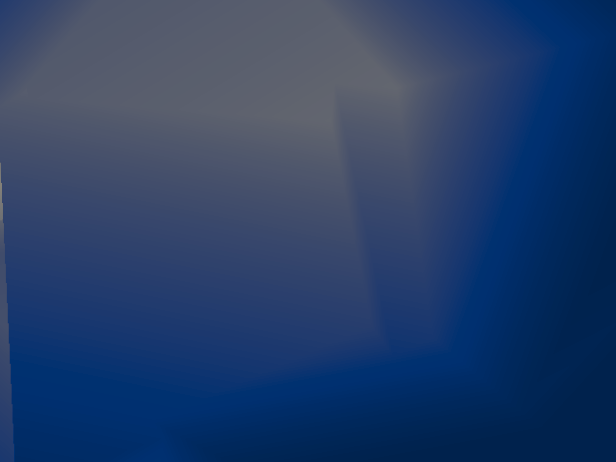} &
			\includegraphics[width=0.136\linewidth]{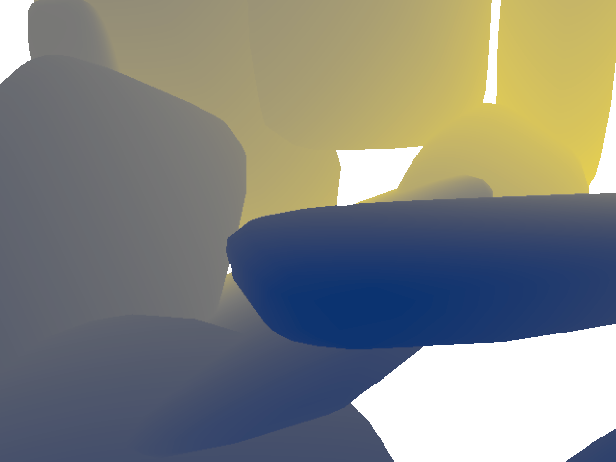} \\

                &
			\includegraphics[width=0.136\linewidth]{fig/examples_new/rgb/final_0489_image.png}  &
			\includegraphics[width=0.136\linewidth]{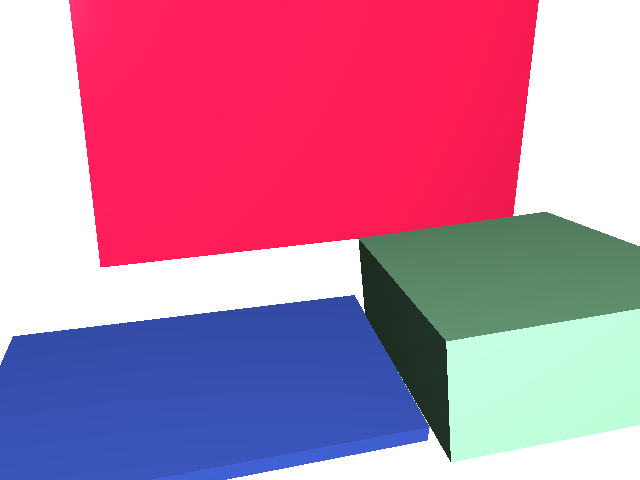} &
			\includegraphics[width=0.136\linewidth]{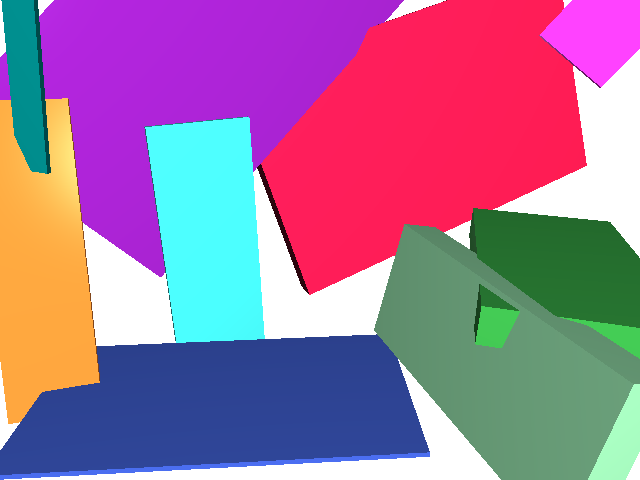} &
			\includegraphics[width=0.136\linewidth]{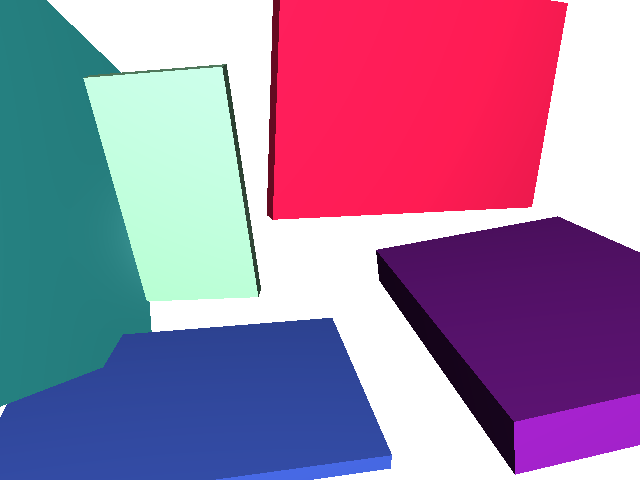} &
			\includegraphics[width=0.136\linewidth]{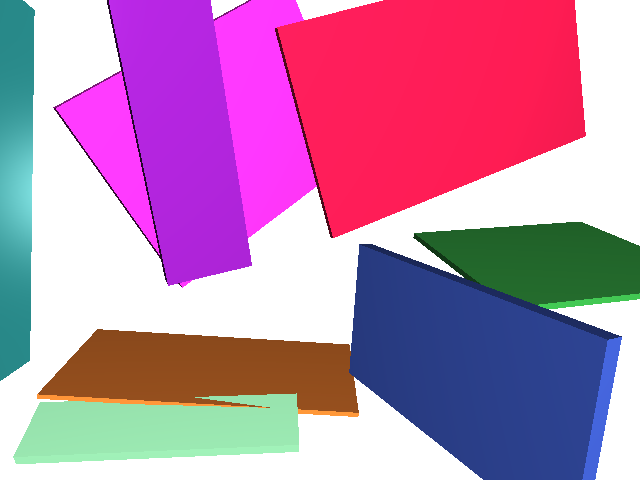} &
			\includegraphics[width=0.136\linewidth]{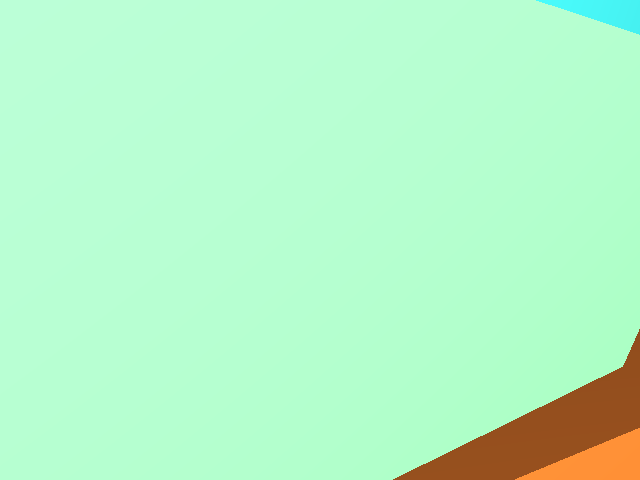} &
			\includegraphics[width=0.136\linewidth]{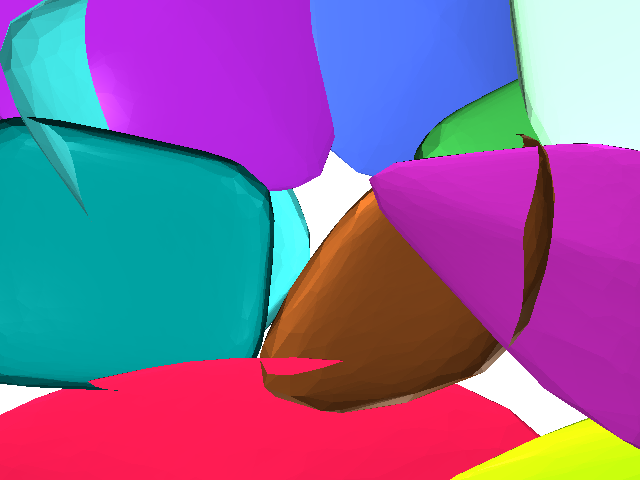} \\
   
                &
			\includegraphics[width=0.136\linewidth]{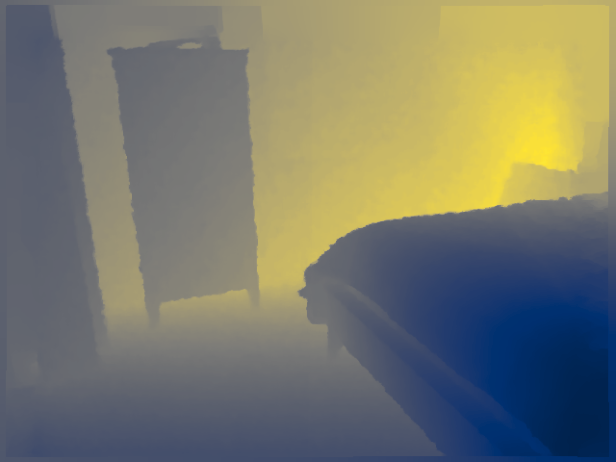}  &
			\includegraphics[width=0.136\linewidth]{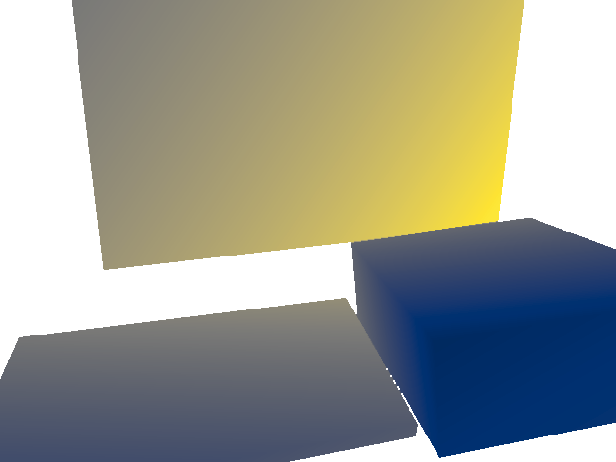} &
			\includegraphics[width=0.136\linewidth]{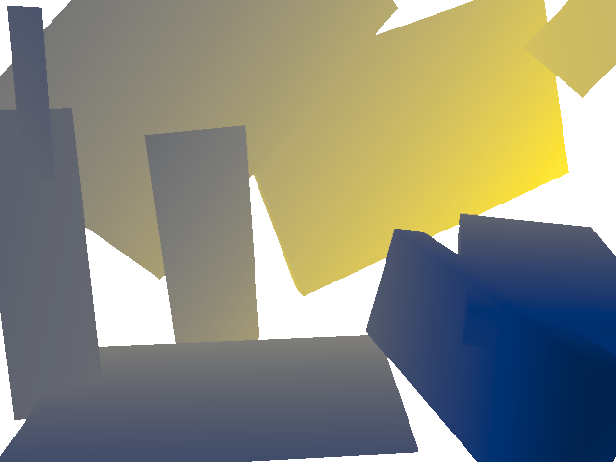} &
			\includegraphics[width=0.136\linewidth]{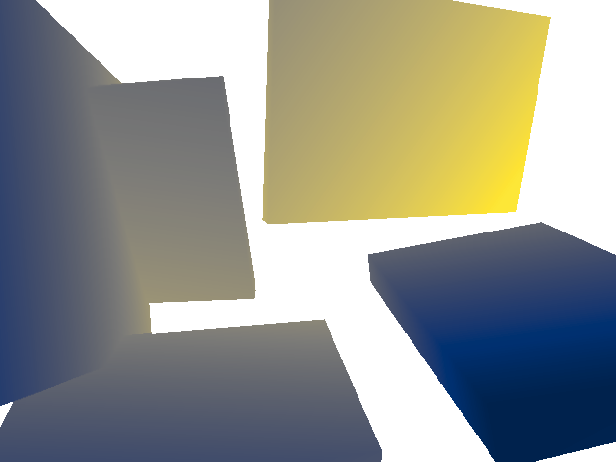} &
			\includegraphics[width=0.136\linewidth]{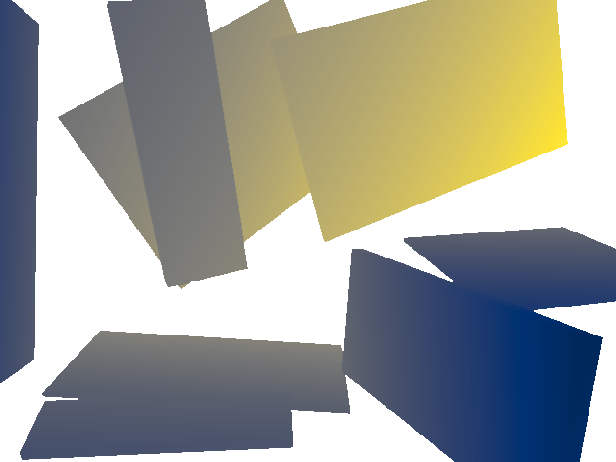} &
			\includegraphics[width=0.136\linewidth]{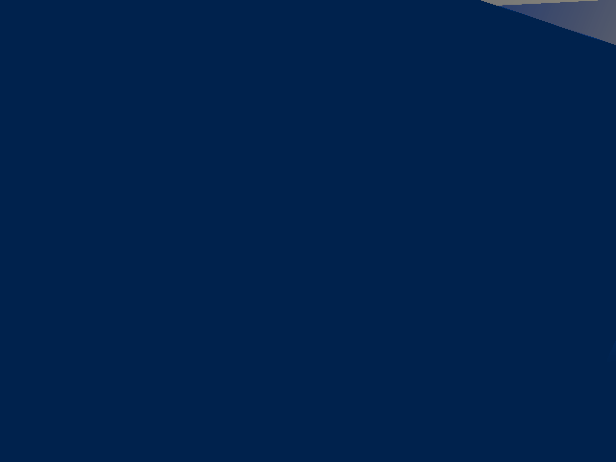} &
			\includegraphics[width=0.136\linewidth]{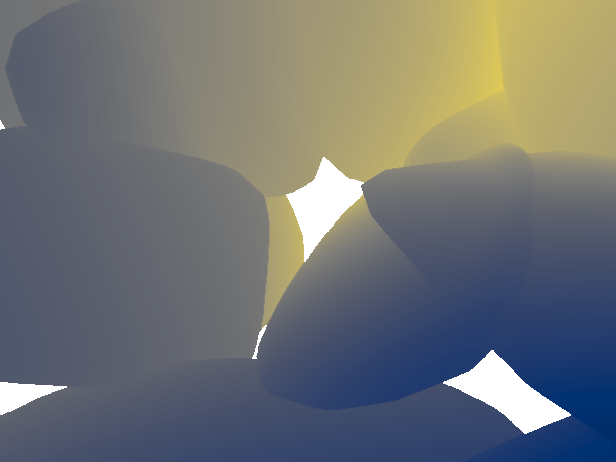} \\

        \end{tabular}
     
		\caption{\textbf{NYU: Qualitative Results.}
		{
		First column: Input RGB images. 
		Columns 2-5: Cuboids obtained with our proposed methods, using either ground truth depth or RGB images as input. 
		Column 6: Cuboids obtained using CONSAC~\cite{KluBra2020} adapted for cuboid fitting.
		Column 7: Superquadrics obtained with SQ-Parsing~\cite{paschalidou2019superquadrics}, trained on NYU~\cite{silberman2012indoor}.
		For our methods and for CONSAC, the colours convey the order in which the cuboids have been selected: \textcolor{c_red}{red}, \textcolor{c_blue}{blue}, \textcolor{c_green}{green}, \textcolor{c_purple}{purple}, \textcolor{c_cyan}{cyan}, \textcolor{c_orange}{orange}.
		}}
		\label{fig:examples}
  \end{center}
\end{figure*}   

\newlength{\imgwidth}
\setlength{\imgwidth}{0.305\linewidth}
    
\begin{figure}	

\setlength\tabcolsep{0.1em}
\renewcommand{\arraystretch}{1.} 
	\begin{center}
        \begin{tabular}{cccc}
             &
             Ground Truth & 
             Neural Solver & 
             Numerical Solver  \\
             
\parbox[t]{3mm}{{\rotatebox[origin=l]{90}{\hspace{1.1em} RGB Image}}}
& \includegraphics[width=\imgwidth]{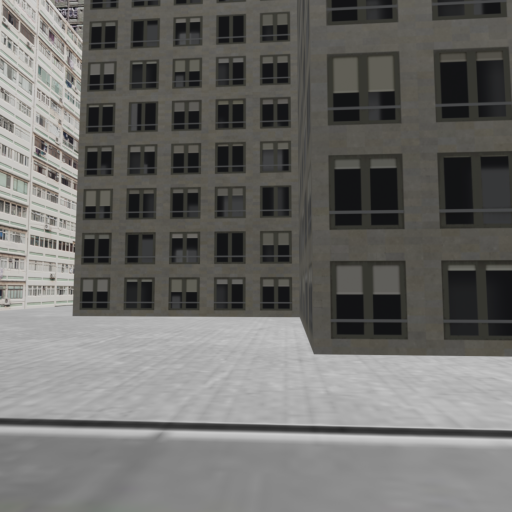} 
& \includegraphics[width=\imgwidth]{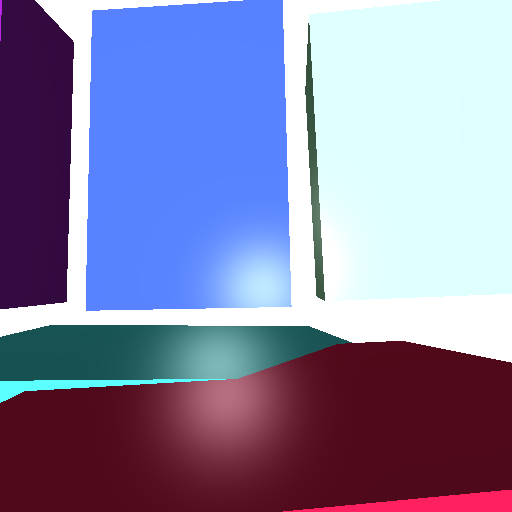} 
& \includegraphics[width=\imgwidth]{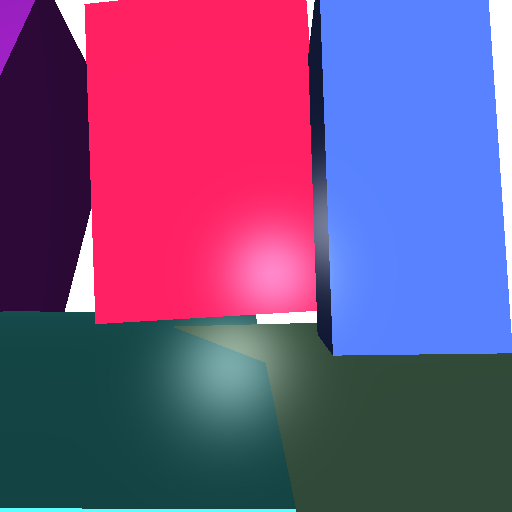} \\
\parbox[t]{3mm}{{\rotatebox[origin=l]{90}{\hspace{2em} Depth}}}
& \includegraphics[width=\imgwidth]{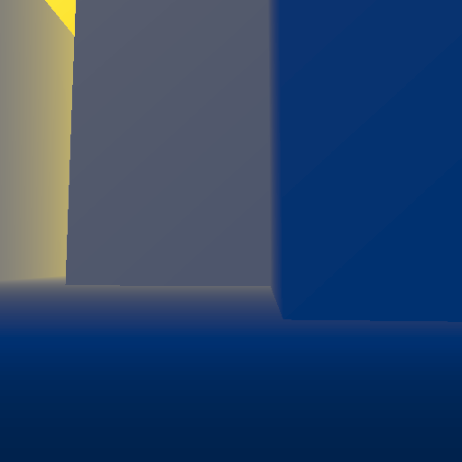} 
& \includegraphics[width=\imgwidth]{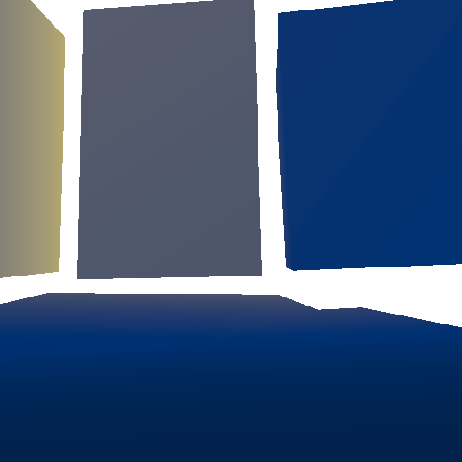} 
& \includegraphics[width=\imgwidth]{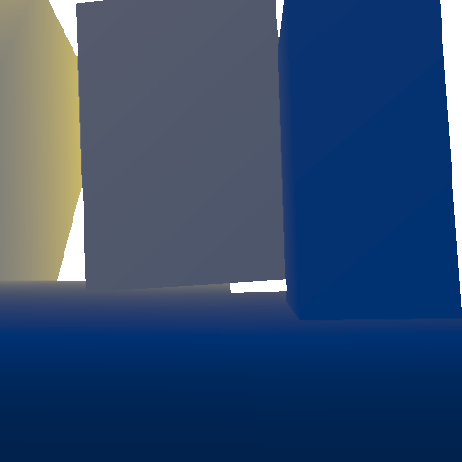} \\

& \includegraphics[width=\imgwidth]{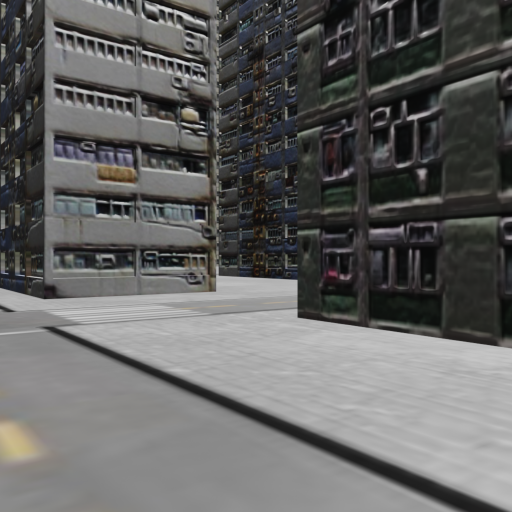} 
& \includegraphics[width=\imgwidth]{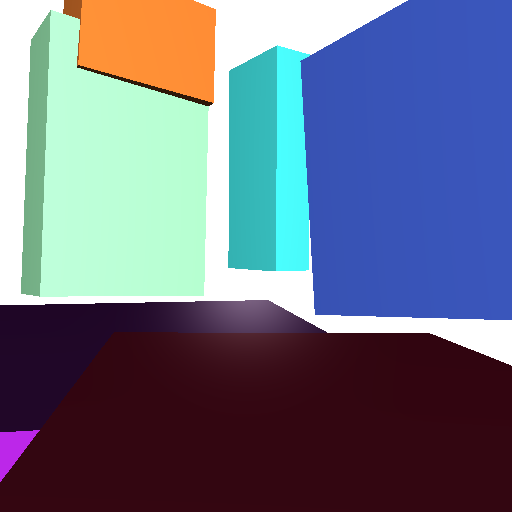} 
& \includegraphics[width=\imgwidth]{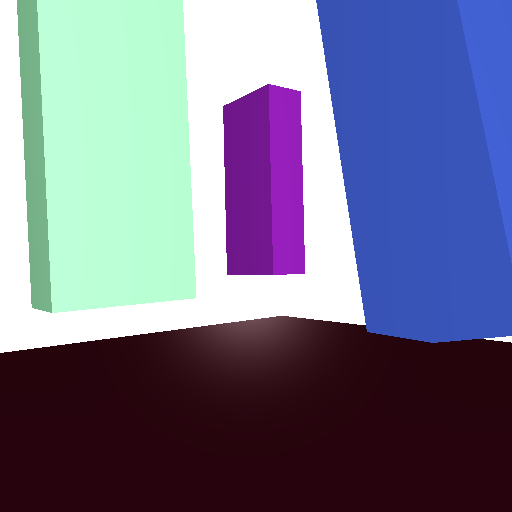} \\
& \includegraphics[width=\imgwidth]{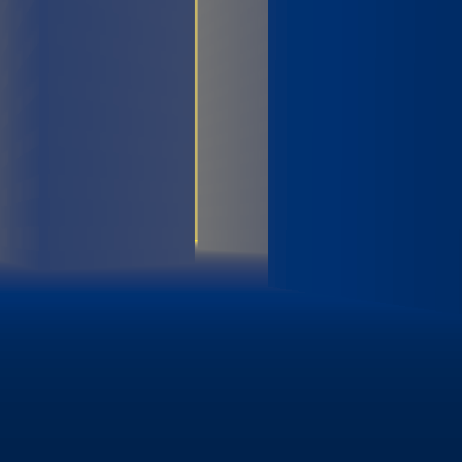} 
& \includegraphics[width=\imgwidth]{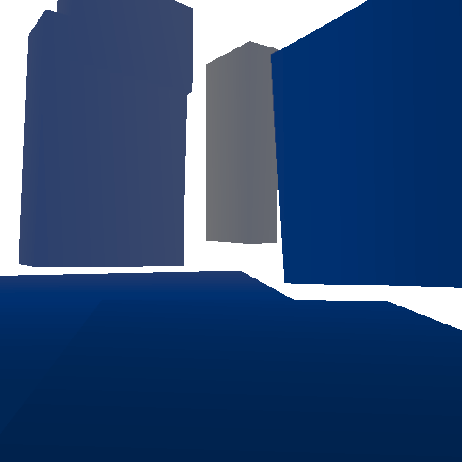} 
& \includegraphics[width=\imgwidth]{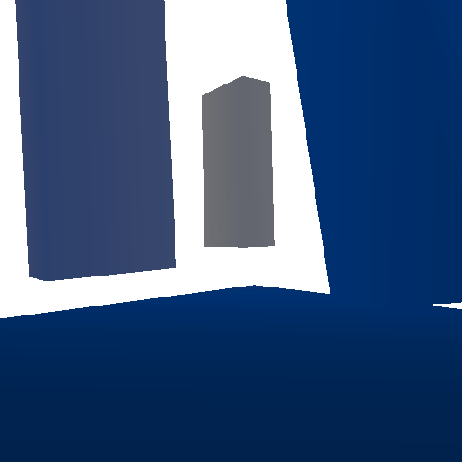} \\

& \includegraphics[width=\imgwidth]{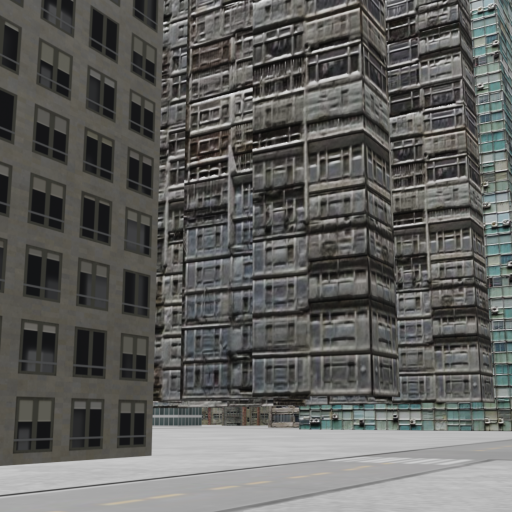} 
& \includegraphics[width=\imgwidth]{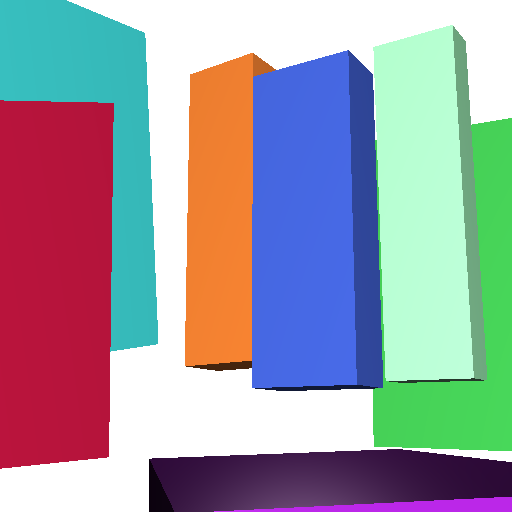} 
& \includegraphics[width=\imgwidth]{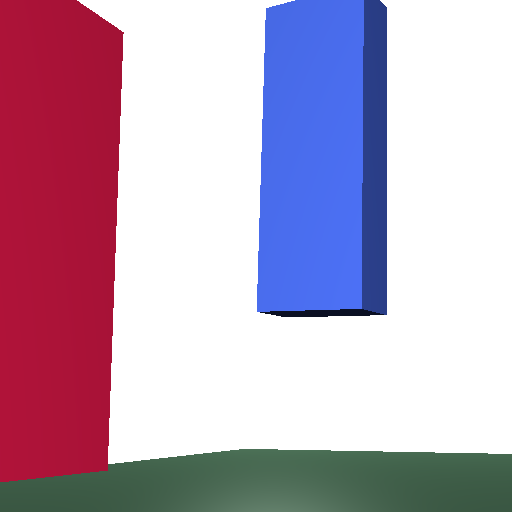} \\
& \includegraphics[width=\imgwidth]{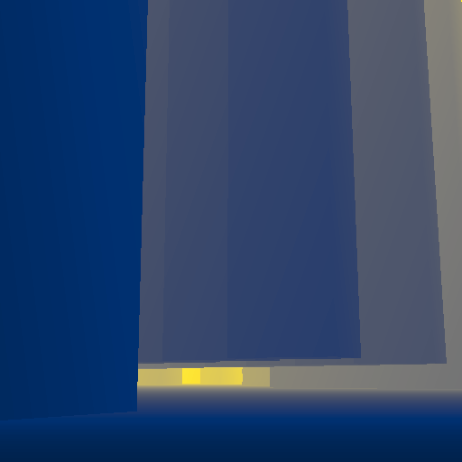} 
& \includegraphics[width=\imgwidth]{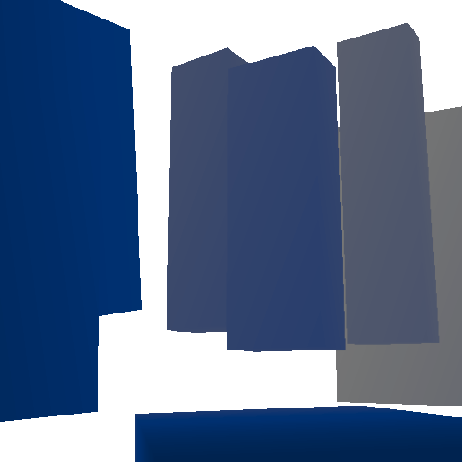} 
& \includegraphics[width=\imgwidth]{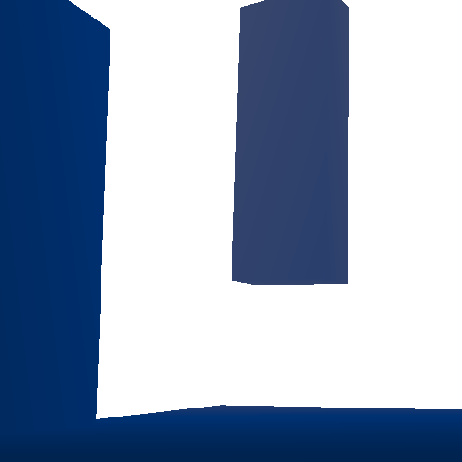} \\

& \includegraphics[width=\imgwidth]{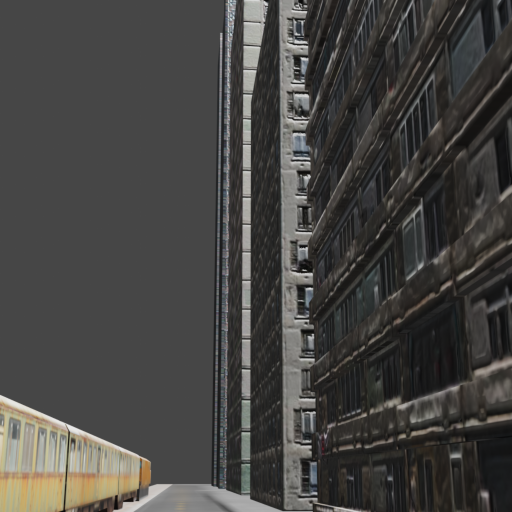} 
& \includegraphics[width=\imgwidth]{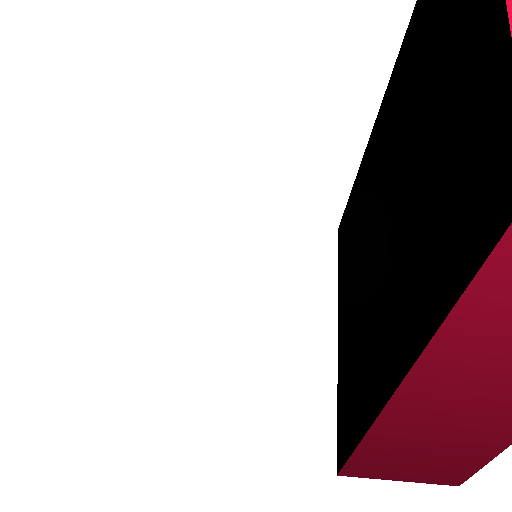} 
& \includegraphics[width=\imgwidth]{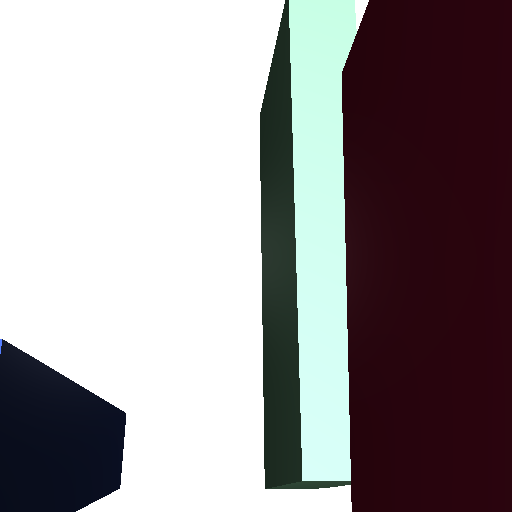} \\
& \includegraphics[width=\imgwidth]{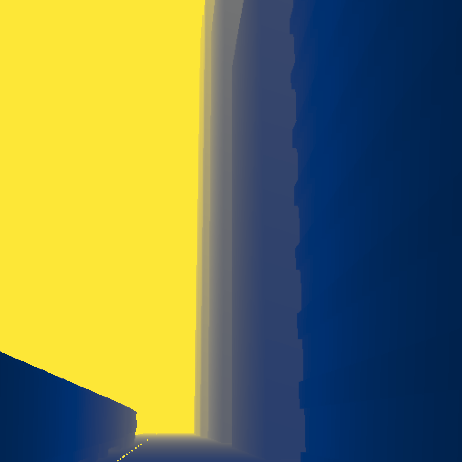} 
& \includegraphics[width=\imgwidth]{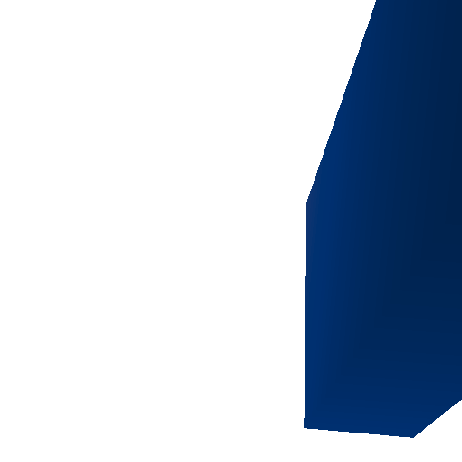} 
& \includegraphics[width=\imgwidth]{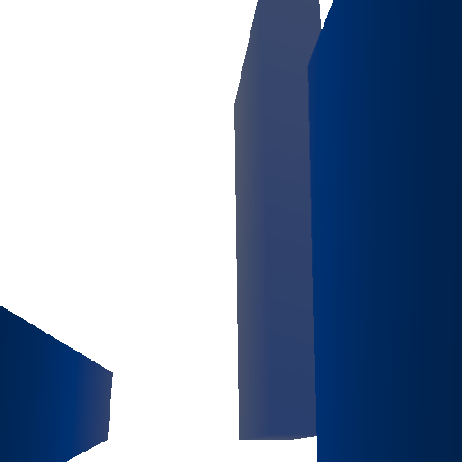} \\

        \end{tabular}
     
		\caption{\new{\textbf{SMH: Qualitative Results.}
		First column: Ground truth images and depth. 
		Columns 2-3: Cuboids obtained with our proposed methods, using ground truth depth as input. 
		}}
		\label{fig:smh_examples}
  \end{center}
\end{figure}

\begin{figure}	
		\centering
		\begin{subfigure}[t]{0.92\linewidth}
			\centering
			\includegraphics[width=0.32\linewidth]{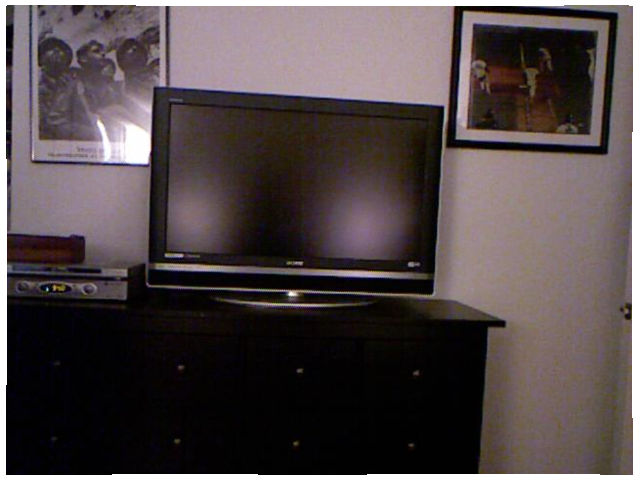}
			\includegraphics[width=0.32\linewidth]{fig/oai/008_0322_0764_original.jpg}
			\includegraphics[width=0.32\linewidth]{fig/oai/015_0459_1095_original.jpg}
   \vspace{0.15em}
		\end{subfigure}

		\begin{subfigure}[t]{0.92\linewidth}
			\centering
			\includegraphics[width=0.32\linewidth]{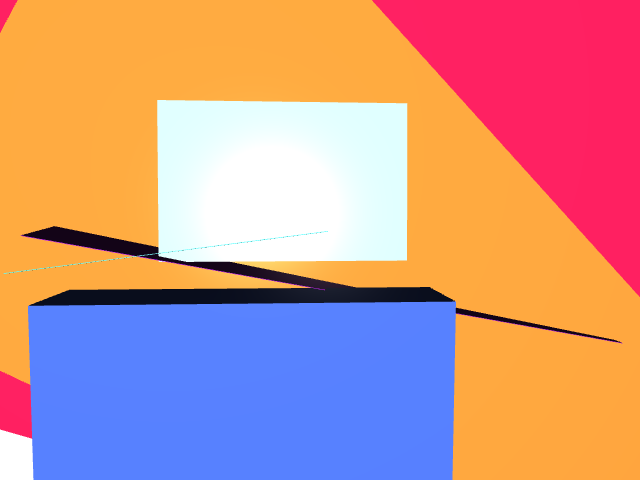}
			\includegraphics[width=0.32\linewidth]{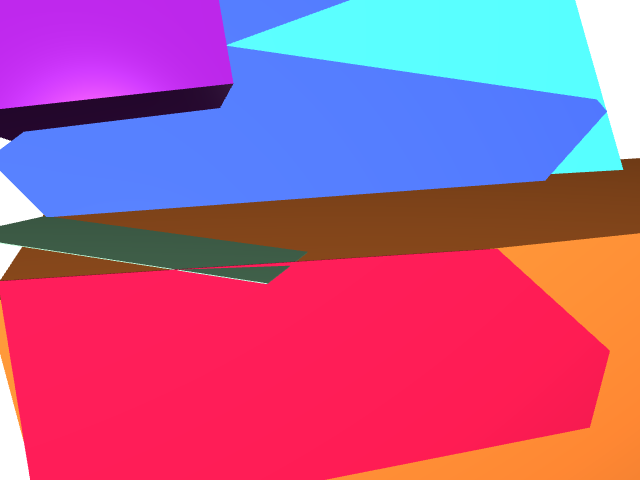}
			\includegraphics[width=0.32\linewidth]{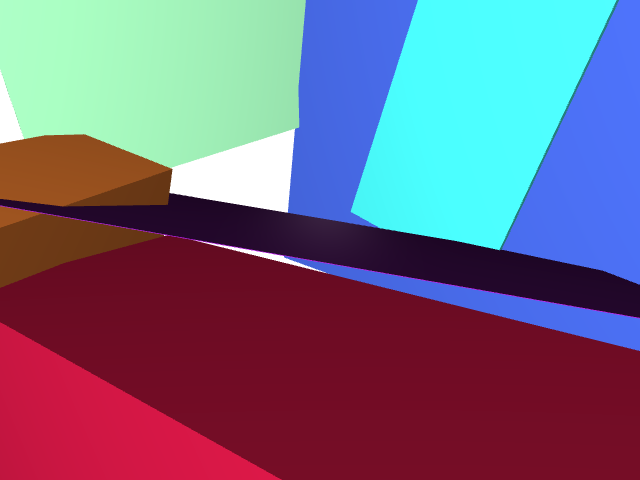}
   \vspace{0.15em}
		\end{subfigure}
	\begin{subfigure}[t]{0.92\linewidth}
			\centering
			\includegraphics[width=0.32\linewidth]{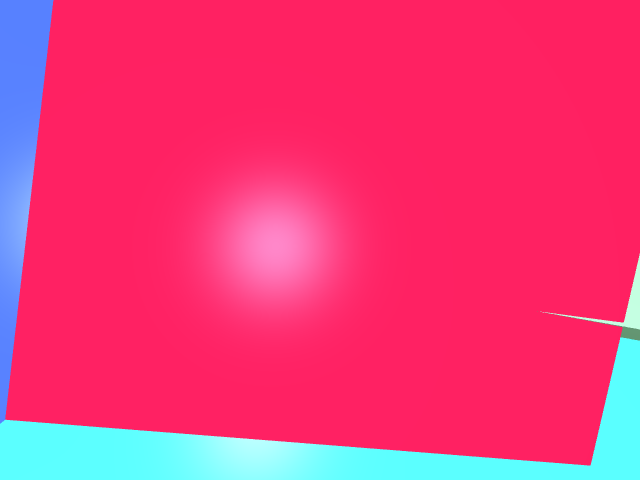}
			\includegraphics[width=0.32\linewidth]{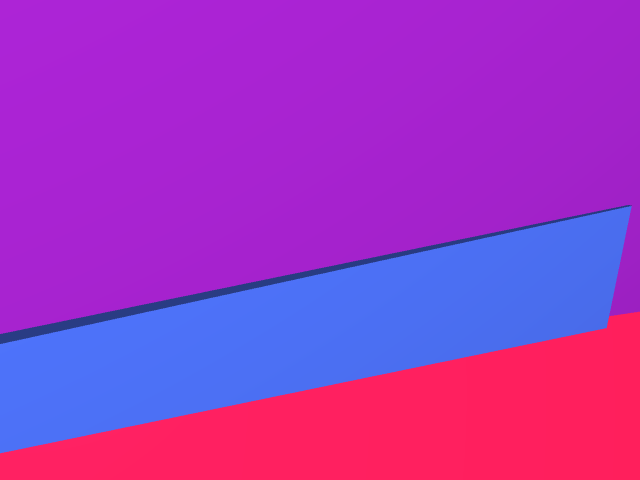}
			\includegraphics[width=0.32\linewidth]{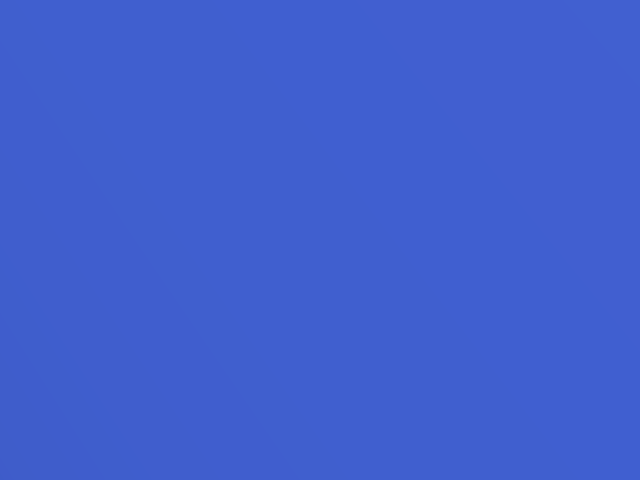}
		\end{subfigure}

		\caption{{\textbf{Occlusion-Aware Inlier Counting:} We compare cuboid based scene abstractions with (second row) and without (third row) our occlusion-aware inlier counting. Original images are on the first row.
		See Sec.~\ref{par:ablation_oai} for details.}
  \vspace{-1em}
  }
		\label{fig:oai_examples}
\end{figure}

\begin{figure*}	
		\begin{subfigure}[t]{0.99\linewidth}
			\includegraphics[width=\linewidth]{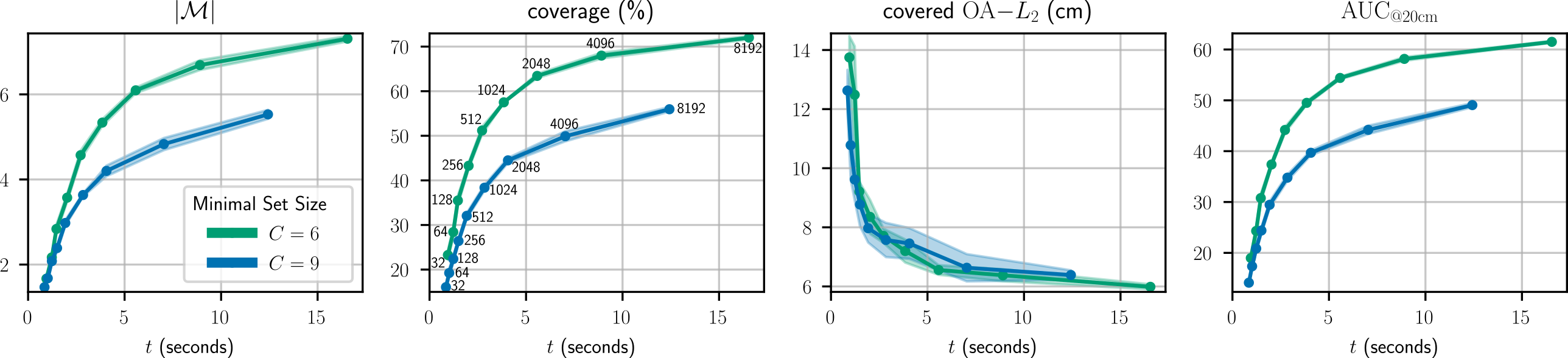}
			\caption{{
			We plot the number of cuboids $|\set{M}|$, image coverage, occlusion-aware (OA) distance of covered points and AUC$_{@\SI{20}{\centi\meter}}$ as a function of computation time $t$.
			We vary the computation time via the number of sampled cuboid hypotheses $|\set{H}|$, which we denote via small annotations in the second plot.
			}}
		\label{fig:graphs_mss:time}
		\end{subfigure}
		
		\begin{subfigure}[t]{0.99\linewidth}
			\includegraphics[width=\linewidth]{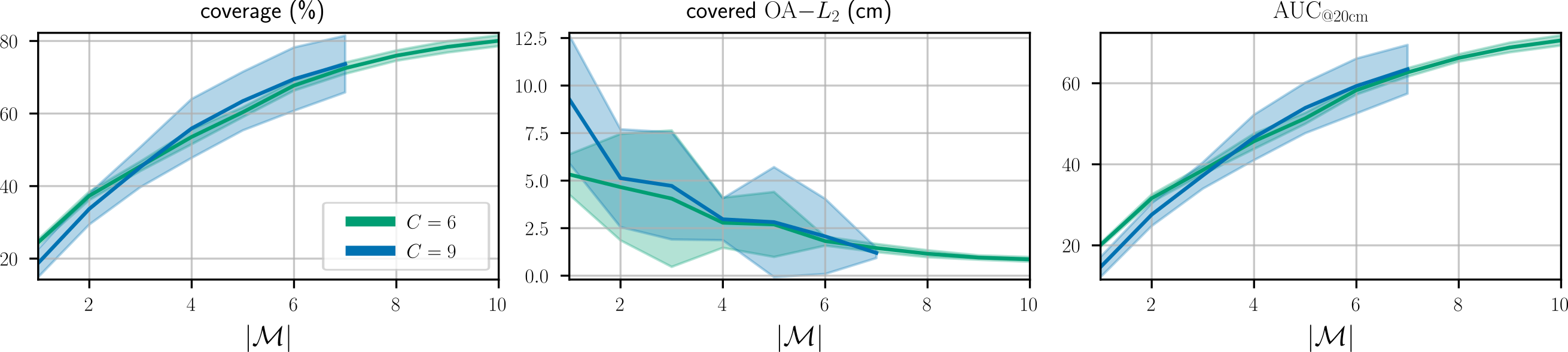}
			\caption{{
			We plot image coverage, occlusion-aware (OA) distance of covered points and AUC$_{@\SI{20}{\centi\meter}}$ as a function of the number of extracted cuboids $|\set{M}|$.
			A smaller number of cuboids indicates a more parsimonious abstraction.
			If more than $|\set{M}|$ cuboids were extracted for a scene, we only consider the first $|\set{M}|$ cuboids for evaluation.
			}}
		\label{fig:graphs_mss:per_cuboid}
		\end{subfigure}
		\caption{
		{
		\textbf{Minimal Set Size:} We analyse the influence of minimal set size $C$ on performance, comparing $C=6$ with $C=9$ on the NYU validation set.
		We look at computational efficiency in Fig.~(a), while Fig.~(b) shows the performance of both variants as a function of parsimony. Shaded areas indicate the $\pm \sigma$ interval.  See Sec.~\ref{subsec:minimal_set_size} for a theoretical analysis and Sec.~\ref{subsec:eval_mss} for a discussion of the results. 
		}
  \vspace{-.5em}
		}
		\label{fig:graphs_mss}
\end{figure*}   
From the qualitative examples in Fig.~\ref{fig:examples}, we observe that the cuboids generated by CONSAC are overly large and cluttered.
This explains the high image coverage it achieves, but results in cuboid configurations which do not abstract the scene in any meaningful way.
The superquadrics generated by SQ-Parsing appear to be generally smaller, but still very cluttered.
They do not resemble the structures and objects present in the images, and hence do not provide sensible abstractions either.
All variants of our method, in contrast, are usually able to generate cuboids which represent large objects, walls and the floor, and we can sometimes identify smaller objects as well.
While size and pose of the cuboids may not always be optimal, the abstractions are significantly more meaningful than those provided by the baseline methods.

\subsection{Minimal Set Size}
\label{subsec:eval_mss}

As we discuss in Sec.~\ref{subsec:minimal_set_size}, it is not immediately clear from a theoretical point of view which minimal set size $C$, \ie the number of points to be sampled in order to determine the parameters of a single cuboid, is optimal.
We thus perform an ablation study investigating both $C=6$ and $C=9$.
Similar to the comparison in Sec.~\ref{subsec:eval_minimal_solver}, we analyse the performance of both variants as a function of computation time, shown in Fig.~\ref{fig:graphs_mss:time}.
Given the same number of cuboid hypotheses $|\set{H}|$, our method eventually selects more cuboids $|\set{M}|$ when using a minimal set size of $C=6$.
Consequently, it achieves a higher image coverage and higher AUC values.
This indicates that $C=6$ is more sample efficient than $C=9$, as it is less likely to sample outlier points and produce ill-fitting cuboids.
With regards to the mean OA distance of covered points, $C=6$ and $C=9$ appear to be roughly on par.
In addition, we also show the performance of both variants as a function of actual cuboids $|\set{M}|$ per image, using a fixed number of hypotheses $|\set{H}|=4096$, in Fig.~\ref{fig:graphs_mss:per_cuboid}.
When limited to the same number of extracted cuboids, both variants behave very similarly.
Image coverage and AUC are roughly on par, and while both variants exhibit high standard deviations \wrt the OA distance of covered points for different values of $|\set{M}|$, we consider these results qualitatively similar.
We conclude that our method is more efficient with $C=6$ than $C=9$, as it is less likely to sample outliers and thus more likely to extract a larger number of cuboids $|\set{M}|$, while the quality of the extracted cuboids is roughly on par.

\subsection{Occlusion-Aware Inlier Counting}
\label{subsec:eval_oai}
{
\label{par:ablation_oai}
In order to demonstrate the impact of our proposed occlusion-aware (OA) inlier counting, we selectively enabled or disabled it for the following parts of our method:
\begin{compactitem}
    \item Training: sampling and hypothesis selection.
    \item Training: loss computation.
    \item Inference: sampling and hypothesis selection.
\end{compactitem}
When disabling the occlusion-aware inlier counting for one component, we use regular inlier counting based on the minimal $L_2$ distance instead.
We then evaluate these variants using ground truth depth input on the NYU~\cite{silberman2012indoor} validation set, and provide the quantitative results in Tab.~\ref{tab:ablation_results_oai}.
We show additional qualitative examples comparing cuboid based scene abstractions we attain with and without occlusion-aware inlier counting in Fig.~\ref{fig:oai_examples}.

As these results show, performance \wrt the occlusion-aware $L_2$ distance degrades severely when disabling the occlusion-aware inlier counting during inference, regardless of whether it was used during training (cf. rows 1-4 in Tab.~\ref{tab:ablation_results_oai}).
While the number of primitives, as well as the coverage, increase significantly, AUC percentages drop to single digits, and the mean distances for covered points increase from around $\SI{6.4}{\centi\meter}$ to roughly $\SI{1.3}{\meter}$.
This shows that the recovered cuboids cover more points, but create significantly more occlusions doing so, which results in less reasonable scene abstractions.
However, when we look at the impact of occlusion-aware inlier counting during training (cf. rows 5-8 in Tab.~\ref{tab:ablation_results_oai}), the differences are not as clear-cut.
Using occlusion-aware inlier counting for both sampling and loss during training still yields the highest AUC values, albeit with slimmer margins, between $0.2$ and $1.0$ percentage points.
Similarly, the mean OA distance of covered points is largely unaffected, while the mean OA distance of all points decreases only slightly ($1.5-\SI{2.0}{\centi\meter}$) which can be explained by the increase in number of cuboids ($0.2-0.8$) and coverage ($0.3-0.5$ percentage points).
   
\begin{table*}

\renewcommand{\arraystretch}{1.1} 
	\begin{center}
	\begin{tabular}{|c|c|c||ccc|ccc|cc|cc|cc|cc|cc|}
	\cline{1-17}
 	\multicolumn{3}{|c||}{occlusion-aware inlier} & 
	\multicolumn{6}{c|}{} &
	\multicolumn{8}{c|}{occlusion-aware $L_2$-distance} \\
	\multicolumn{2}{|c|}{training} & \multicolumn{1}{c||}{inference} & 
	\multicolumn{2}{c}{{\# primitives}} &\multirow{ 2}{*}{$\downarrow$}  &
	\multicolumn{2}{c}{image area} & \multirow{ 2}{*}{$\uparrow$} &
	\multicolumn{4}{c|}{mean (cm) $\downarrow$} &
	\multicolumn{4}{c|}{AUC (\%) $\uparrow$} \\
        sampling & loss & sampling & 
	\multicolumn{2}{c}{$|\set{M}|$} & &
	\multicolumn{2}{c}{covered (\%)} & &
	\multicolumn{2}{c|}{covered points}      & 
	\multicolumn{2}{c|}{all points} &
	\multicolumn{2}{c|}{$@\SI{20}{\centi\meter}$} & 
	\multicolumn{2}{c|}{${@\SI{5}{\centi\meter}}$}  \\
	\hline

	\xmark & \xmark & \xmark
	& $ 8.9 $ & \multicolumn{2}{c|}{$\scriptstyle \pm 0.21 $}
	& $ 99.8 $ & \multicolumn{2}{c|}{$\scriptstyle \pm \mathbf{0.02} $}
	& $ 135.1 $ & $\scriptstyle \pm 6.00 $
	& $ 135.0 $ & $\scriptstyle \pm 6.01 $
	& $ 6.2 $ & $\scriptstyle \pm 0.45 $
	& $ 2.1 $ & $\scriptstyle \pm 0.19 $ \\

	\xmark & \cmark & \xmark
	& $ 8.7 $ & \multicolumn{2}{c|}{$\scriptstyle \pm \mathbf{0.11} $}
	& $\mathbf{99.9} $ & \multicolumn{2}{c|}{$\scriptstyle \pm \mathbf{0.02}$}
	& $ 136.6 $ & $\scriptstyle \pm 3.18 $
	& $ 136.5 $ & $\scriptstyle \pm 3.18 $
	& $ 6.1 $ & $\scriptstyle \pm \mathbf{0.35} $
	& $ 2.1 $ & $\scriptstyle \pm \mathbf{0.18} $ \\
	\cmark & \xmark & \xmark
	& $ 9.1 $ & \multicolumn{2}{c|}{$\scriptstyle \pm 0.21 $}
	& $ 99.8 $ & \multicolumn{2}{c|}{$\scriptstyle \pm \mathbf{0.02} $}
	& $ 130.9 $ & $\scriptstyle \pm 5.65 $
	& $ 130.8 $ & $\scriptstyle \pm 5.66 $
	& $ 6.5 $ & $\scriptstyle \pm 0.53 $
	& $ 2.3 $ & $\scriptstyle \pm 0.21 $ \\
	\cmark & \cmark & \xmark
	& $ 9.2 $ & \multicolumn{2}{c|}{$\scriptstyle \pm 0.28 $}
	& $ 99.8 $ & \multicolumn{2}{c|}{$\scriptstyle \pm 0.03 $}
	& $ 129.8 $ & $\scriptstyle \pm 8.27 $
	& $ 129.7 $ & $\scriptstyle \pm 8.28 $
	& $ 6.7 $ & $\scriptstyle \pm 0.82 $
	& $ 2.3 $ & $\scriptstyle \pm 0.33 $ \\

	\xmark & \xmark & \cmark
	& $ 6.1 $ & \multicolumn{2}{c|}{$\scriptstyle \pm 0.34 $}
	& $ 65.6 $ & \multicolumn{2}{c|}{$\scriptstyle \pm 0.69 $}
	& $ \mathbf{6.4} $ & $\scriptstyle \pm \mathbf{0.29} $
	& $ 31.7 $ & $\scriptstyle \pm \mathbf{1.31} $
	& $ 56.3 $ & $\scriptstyle \pm 0.54 $
	& $ 29.5 $ & $\scriptstyle \pm 0.47 $ \\

	\xmark & \cmark & \cmark
	& $ \mathbf{5.8} $ & \multicolumn{2}{c|}{$\scriptstyle \pm 0.26 $}
	& $ 65.5 $ & \multicolumn{2}{c|}{$\scriptstyle \pm 0.97 $}
	& $ 6.6 $ & $\scriptstyle \pm 0.34 $
	& $ 31.8 $ & $\scriptstyle \pm 2.47 $
	& $ 55.8 $ & $\scriptstyle \pm 0.68 $
	& $ 29.0 $ & $\scriptstyle \pm 0.46 $ \\

	\cmark & \xmark & \cmark
	& $ 6.4 $ & \multicolumn{2}{c|}{$\scriptstyle \pm 0.52 $}
	& $ 65.7 $ & \multicolumn{2}{c|}{$\scriptstyle \pm 2.65 $}
	& $ \mathbf{6.4} $ & $\scriptstyle \pm 0.33 $
	& $ 32.2 $ & $\scriptstyle \pm 5.16 $
	& $ 56.3 $ & $\scriptstyle \pm 2.57 $
	& $ 29.7 $ & $\scriptstyle \pm 1.44 $ \\

	\cmark & \cmark & \cmark
	& $ 6.6 $ & \multicolumn{2}{c|}{$\scriptstyle \pm 0.36 $}
	& $ {66.0} $ & \multicolumn{2}{c|}{$\scriptstyle \pm 1.17 $}
	& $ \mathbf{6.4} $ & $\scriptstyle \pm 0.52 $ 
	& $ \mathbf{30.2} $ & $\scriptstyle \pm 1.57 $
	& $ \mathbf{56.8} $ & $\scriptstyle \pm 1.19 $
	& $ \mathbf{29.9} $ & $\scriptstyle \pm 0.87 $ \\
	\hline
	\end{tabular}
    \end{center}
	\caption{\textbf{Ablation Study:} 
	We analyse the influence of occlusion-aware inlier counting. We enable or disable it for hypothesis sampling and selection during training and inference, and for loss computation during training. We evaluate on {the validation set of} NYU Depth v2~\cite{silberman2012indoor} for depth input. {We present the number of primitives $|\set{M}|$, image coverage, mean of the occlusion-aware (OA) distances for covered as well as for all points, and AUC values for two upper bounds of the OA $L_2$ distance.} See Sec.~\ref{par:ablation_oai} for details.
	} 
	\label{tab:ablation_results_oai}
\end{table*}

\begin{figure*}
	\includegraphics[width=\linewidth]{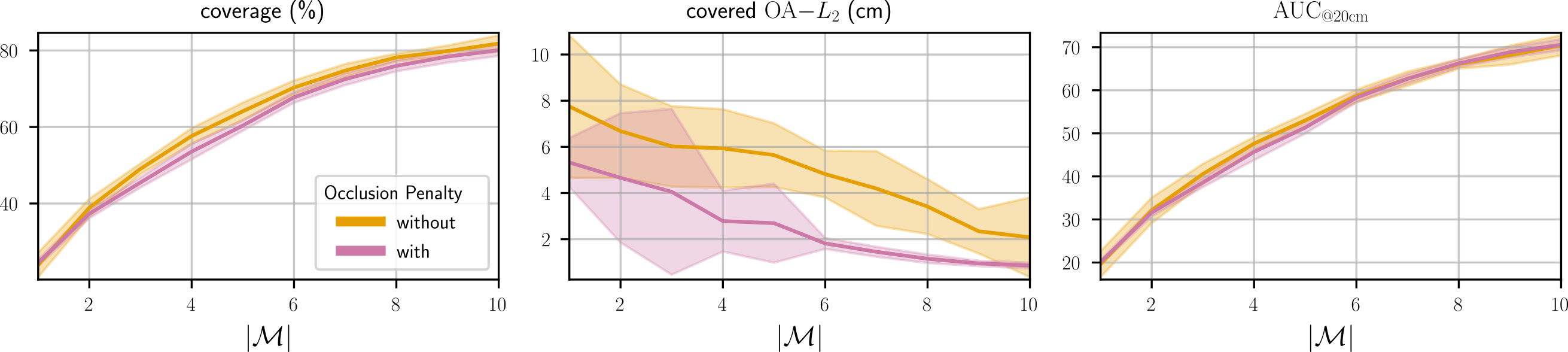}
	\caption{{
	        \textbf{Occlusion Penalty:} 
	        We compare the performance of our method with and without the additional penalty for occluded points (cf. Sec.~\ref{par:occlusion_penalty}).
			We plot image coverage, occlusion-aware (OA) distance of covered points and AUC$_{@\SI{20}{\centi\meter}}$ as a function of the number of extracted cuboids $|\set{M}|$.
			A smaller number of cuboids indicates a more parsimonious abstraction.
			If more than $|\set{M}|$ cuboids were extracted for a scene, we only consider the first $|\set{M}|$ cuboids for evaluation.
                Shaded areas indicate the $\pm \sigma$ interval. 
			See Sec.~\ref{par:ablation_oai} for a detailed discussion of the results.
			}}
	\label{fig:graphs_op}
\end{figure*}
\paragraph{Occlusion Penalty}
As described in Sec.~\ref{par:occlusion_penalty}, we propose a new leaky occlusion function which can penalise occlusions more strongly than before.
We set the transition point to $\tau_c = 2\tau$ and compare it to the non-leaky inlier function without an additional occlusion penalty, which is equivalent to $\tau_c \rightarrow \infty$.
In Fig.~\ref{fig:graphs_op} we show three evaluation metrics as a function of the number of cuboids $|\set{M}|$ per scene.
While image coverage and AUC are on par for both variants, the mean occlusion-aware distance of covered points is measurably lower on average when we utilise the occlusion penalty, with a margin of roughly $1-\SI{3}{\centi\meter}$.
Using this additional penalty, we are thus able to select cuboids which fit the scene more accurately and provide a more sensible abstraction.
}

\begin{table*}
\renewcommand{\arraystretch}{1.1} 
	\begin{center}
	\begin{tabular}{|c|c|c||ccc|ccc|cc|cc|cc|cc|cc|}
 
	\hline
        \multicolumn{3}{|c||}{regularisation} & 
	\multicolumn{2}{c}{{\# primitives}} &\multirow{ 2}{*}{$\downarrow$}  &
	\multicolumn{2}{c}{image area} & \multirow{ 2}{*}{$\uparrow$} &
	\multicolumn{4}{c|}{OA-$L_2$ mean (cm) $\downarrow$} &
	\multicolumn{4}{c|}{OA-$L_2$ AUC (\%) $\uparrow$} \\
        $\ell_{\text{im}}$ & $\ell_{\text{corr}}$ & $\ell_{\text{entropy}}$ &
	\multicolumn{2}{c}{$|\set{M}|$} & &
	\multicolumn{2}{c}{covered (\%)} & &
	\multicolumn{2}{c|}{covered points}      & 
	\multicolumn{2}{c|}{all points} &
	\multicolumn{2}{c|}{$@\SI{20}{\centi\meter}$} & 
	\multicolumn{2}{c|}{${@\SI{5}{\centi\meter}}$}  \\
	\hline

	\xmark & \xmark & \xmark
	& $ \mathbf{3.5} $ & \multicolumn{2}{c|}{ $\scriptstyle \pm 0.16 $}
	& $ 46.6 $ & \multicolumn{2}{c|}{$\scriptstyle \pm 4.95 $}
	& $ 11.2 $ & $\scriptstyle \pm 0.91 $ 
	& $ 71.0 $ & $\scriptstyle \pm 13.42 $
	& $ 35.2 $ & $\scriptstyle \pm 3.90 $
	& $ 17.6 $ & $\scriptstyle \pm 2.25 $ \\
	\xmark & \xmark & \cmark
	& $ 3.9 $ & \multicolumn{2}{c|}{$\scriptstyle \pm 0.48 $}
	& $ 52.2 $ & \multicolumn{2}{c|}{$\scriptstyle \pm 2.78 $}
	& $ 9.4 $ & $\scriptstyle \pm 1.20 $ 
	& $ 58.4 $ & $\scriptstyle \pm 5.28 $
	& $ 41.2 $ & $\scriptstyle \pm 3.40 $
	& $ 21.0 $ & $\scriptstyle \pm 1.86 $ \\
	\xmark & \cmark & \xmark
	& $ \mathbf{3.5} $ & \multicolumn{2}{c|}{$\scriptstyle \pm 0.28 $}
	& $ 51.0 $ & \multicolumn{2}{c|}{$\scriptstyle \pm 3.14 $}
	& $ 9.5 $ & $\scriptstyle \pm 1.55 $ 
	& $ 61.4 $ & $\scriptstyle \pm 4.02 $
	& $ 40.0 $ & $\scriptstyle \pm 3.49 $
	& $ 20.4 $ & $\scriptstyle \pm 1.87 $ \\
	\xmark & \cmark & \cmark
	& $ 3.9 $ & \multicolumn{2}{c|}{$\scriptstyle \pm \mathbf{0.10} $}
	& $ 54.7 $ & \multicolumn{2}{c|}{$\scriptstyle \pm 2.40 $}
	& $ 8.9 $ & $\scriptstyle \pm 0.87 $ 
	& $ 54.7 $ & $\scriptstyle \pm 5.35 $
	& $ 43.2 $ & $\scriptstyle \pm 2.05 $
	& $ 22.0 $ & $\scriptstyle \pm 1.13 $ \\
	\cmark & \xmark & \xmark
	& $ 3.7 $ & \multicolumn{2}{c|}{$\scriptstyle \pm 0.18 $}
	& $ 55.5 $ & \multicolumn{2}{c|}{$\scriptstyle \pm 1.00 $}
	& $ 8.7 $ & $\scriptstyle \pm 0.91 $ 
	& $ 53.9 $ & $\scriptstyle \pm 2.97 $
	& $ 45.0 $ & $\scriptstyle \pm 1.00 $
	& $ 22.7 $ & $\scriptstyle \pm 0.67 $ \\
	\cmark & \xmark & \cmark
	& $ 3.7 $ & \multicolumn{2}{c|}{$\scriptstyle \pm 0.20 $}
	& $ 54.2 $ & \multicolumn{2}{c|}{$\scriptstyle \pm 1.46 $}
	& $ 8.7 $ & $\scriptstyle \pm 1.55 $ 
	& $ 57.5 $ & $\scriptstyle \pm 2.88 $
	& $ 44.2 $ & $\scriptstyle \pm 1.05 $
	& $ 22.4 $ & $\scriptstyle \pm 0.69 $ \\
	\cmark & \cmark & \xmark
	& $ 6.5 $ & \multicolumn{2}{c|}{$\scriptstyle \pm {0.27} $}
	& $ 65.7 $ & \multicolumn{2}{c|}{$\scriptstyle \pm \mathbf{0.92} $}
	& $ 6.5 $ & $\scriptstyle \pm \mathbf{0.33} $ 
	& $ 31.0 $ & $\scriptstyle \pm \mathbf{1.38} $
	& $ 56.4 $ & $\scriptstyle \pm \mathbf{0.71} $
	& $ 29.6 $ & $\scriptstyle \pm \mathbf{0.53} $ \\
	\cmark & \cmark & \cmark
	& $ 6.6 $ & \multicolumn{2}{c|}{$\scriptstyle \pm 0.36 $}
	& $ \mathbf{66.0} $ & \multicolumn{2}{c|}{$\scriptstyle \pm 1.17 $}
	& $ \mathbf{6.4} $ & $\scriptstyle \pm 0.52 $ 
	& $ \mathbf{30.2} $ & $\scriptstyle \pm 1.57 $
	& $ \mathbf{56.8} $ & $\scriptstyle \pm 1.19 $
	& $ \mathbf{29.9} $ & $\scriptstyle \pm 0.87 $ \\
\hline
	\end{tabular}
    \end{center}
	\caption{\textbf{Ablation Study:}
	We evaluate our approach with and without each of the three regularisation loss terms described in Sec.~\ref{sec:training_regularisation}: sampling weight map correlation ($\kappa_{\text{corr}}$), selection probability entropy ($\kappa_{\text{entropy}}$), and inlier masking regularisation ($\kappa_{\text{im}}$).
	Results are computed using depth input, numerical solver and without refinement on the NYU validation set. 
 {See Sec.~\ref{par:ablation_reg} for a discussion.}
 \vspace{-0.25em}
	}
	\label{tab:ablation_results_regularisation}
\end{table*}

\begin{table*}
\renewcommand{\arraystretch}{1.1} 
	\begin{center}
	\begin{tabular}{|c|c||ccc|ccc|cc|cc|cc|cc|}
	\hline
        \multicolumn{1}{|c|}{fine} & 
        \multicolumn{1}{c||}{\multirow{2}{*}{pre-trained}} & 
	\multicolumn{2}{c}{{primitives}} &\multirow{ 2}{*}{}  &
	\multicolumn{2}{c}{image area} & \multirow{ 2}{*}{$\uparrow$} &
	\multicolumn{4}{c|}{OA-$L_2$ mean (cm) $\downarrow$} &
	\multicolumn{4}{c|}{OA-$L_2$ AUC (\%) $\uparrow$} \\
        tune & &
	\multicolumn{2}{c}{$|\set{M}|$  $\downarrow$} & &
	\multicolumn{2}{c}{covered (\%)} & &
	\multicolumn{2}{c|}{covered pts.}      & 
	\multicolumn{2}{c|}{all points} &
	\multicolumn{2}{c|}{$@\SI{20}{\centi\meter}$} & 
	\multicolumn{2}{c|}{${@\SI{5}{\centi\meter}}$}  \\
	\hline
	\multicolumn{16}{|c|}{\rule{0pt}{3ex}depth input + neural solver} \\
	\xmark & all
	& $ {3.8} $ & \multicolumn{2}{c|}{$\scriptstyle \pm 0.16 $}
	& $ 63.4 $ & \multicolumn{2}{c|}{$\scriptstyle \pm 1.49 $}
	& $ {10.1} $ & $\scriptstyle \pm 0.43 $ 
	& $ 42.1 $ & $\scriptstyle \pm 2.61 $
	& $ 46.1 $ & $\scriptstyle \pm 1.17 $
	& $ 20.7 $ & $\scriptstyle \pm 0.53 $ \\

        \cmark & \new{none}
        & $ \mathbf{1.0} $ & \multicolumn{2}{c|}{$\scriptstyle \pm 0.00 $}
	& $ 27.1 $ & \multicolumn{2}{c|}{$\scriptstyle \pm 3.78 $}
	& $ 50.1 $ & $\scriptstyle \pm 4.06 $
	& $ 145.3 $ & $\scriptstyle \pm 8.10 $
	& $ 4.9 $ & $\scriptstyle \pm 0.85 $
	& $ 1.2 $ & $\scriptstyle \pm 0.23 $ \\

        \cmark & \new{sampling network}
 	& $ \mathbf{1.0} $ & \multicolumn{2}{c|}{$\scriptstyle \pm 0.03 $}
	& $ 31.2 $ & \multicolumn{2}{c|}{$\scriptstyle \pm 4.79 $}
	& $ 51.2 $ & $\scriptstyle \pm 8.38 $
	& $ 129.3 $ & $\scriptstyle \pm 18.16 $
	& $ 6.1 $ & $\scriptstyle \pm 1.23 $
	& $ 1.6 $ & $\scriptstyle \pm 0.47 $ \\

        \cmark & \new{neural solver}
	& $ 4.0 $ & \multicolumn{2}{c|}{$\scriptstyle \pm 0.22 $}
	& $ 64.4 $ & \multicolumn{2}{c|}{$\scriptstyle \pm 1.09 $}
	& $ \mathbf{9.8} $ & $\scriptstyle \pm 0.47 $
	& $ \mathbf{39.1} $ & $\scriptstyle \pm 2.13 $
	& $ \mathbf{47.5} $ & $\scriptstyle \pm 0.98 $
	& $ \mathbf{21.9} $ & $\scriptstyle \pm 0.56 $ \\

	\cmark & all
	& $ 3.9 $ & \multicolumn{2}{c|}{$\scriptstyle \pm 0.09 $}
	& $ \mathbf{65.9} $ & \multicolumn{2}{c|}{$\scriptstyle \pm 0.72 $}
	& $ 10.2 $ & $\scriptstyle \pm 0.24 $ 
	& $ {39.2} $ & $\scriptstyle \pm 1.73 $
	& $ {47.3} $ & $\scriptstyle \pm 0.53 $
	& $ {21.3} $ & $\scriptstyle \pm 0.31 $ \\
 
	\multicolumn{16}{|c|}{\rule{0pt}{3ex}RGB input + numerical solver} \\
	\xmark & all
	& $ \mathbf{6.9} $ & \multicolumn{2}{c|}{$\scriptstyle \pm 0.30 $}
	& $ 66.7 $ & \multicolumn{2}{c|}{$\scriptstyle \pm 1.29 $}
	& $ 10.1 $ & $\scriptstyle \pm 0.44 $ 
	& $ 29.2 $ & $\scriptstyle \pm 1.88 $
	& $ 47.2 $ & $\scriptstyle \pm 0.93 $
	& $ 17.9 $ & $\scriptstyle \pm 0.50 $ \\

	\cmark  & all
	& $ 7.1 $ & \multicolumn{2}{c|}{$\scriptstyle \pm 0.10 $}
	& $ \mathbf{68.5} $ & \multicolumn{2}{c|}{$\scriptstyle \pm 0.46 $}
	& $ \mathbf{9.7} $ & $\scriptstyle \pm 0.14 $ 
	& $ \mathbf{27.1} $ & $\scriptstyle \pm 0.84 $
	& $ \mathbf{48.9} $ & $\scriptstyle \pm 0.31 $
	& $ \mathbf{18.9} $ & $\scriptstyle \pm 0.22 $ \\
\multicolumn{16}{|c|}{\rule{0pt}{3ex}RGB input + neural solver} \\

	\xmark  & all
	& $ \mathbf{4.1} $ & \multicolumn{2}{c|}{$\scriptstyle \pm 0.18 $}
	& $ 63.7 $ & \multicolumn{2}{c|}{$\scriptstyle \pm 2.27 $}
	& $ 13.2 $ & $\scriptstyle \pm 0.40 $ 
	& $ 42.3 $ & $\scriptstyle \pm 4.18 $
	& $ 39.2 $ & $\scriptstyle \pm 1.16 $
	& $ 14.4 $ & $\scriptstyle \pm 0.49 $ \\

	\cmark & all
	& $ 4.3 $ & \multicolumn{2}{c|}{$\scriptstyle \pm 0.07 $}
	& $ \mathbf{66.9} $ & \multicolumn{2}{c|}{$\scriptstyle \pm 0.74 $}
	& $ \mathbf{13.0} $ & $\scriptstyle \pm 0.33 $ 
	& $ \mathbf{37.4} $ & $\scriptstyle \pm 1.25 $
	& $ \mathbf{41.0} $ & $\scriptstyle \pm 0.37 $
	& $ \mathbf{15.1} $ & $\scriptstyle \pm 0.23 $ \\
	
	\hline

	\end{tabular}
    \end{center}
	\caption{{\textbf{Ablation Study:} 
 We evaluate the impact of fine-tuning the neural networks for depth estimation, sampling weight estimation and cuboid fitting in an end-to-end manner on the NYU validation set. 
  See Sec.~\ref{par:ablation_finetune} for a discussion of the results.
 }}
	\label{tab:ablation_results_refinement}
\end{table*}

\subsection{Additional Ablation Studies}
\label{subsec:more_ablation}

\paragraph{Regularisation}
\label{par:ablation_reg}
We analyse the effectiveness of the regularisation losses $\ell_{\text{corr}}$, $\ell_{\text{entropy}}$ and $\ell_{\text{im}}$, which we introduce in Sec.~\ref{sec:training_regularisation}.
These losses reduce the correlation coefficients between sampling weight maps $\vec{Q}$, maximise the entropy of selection probabilities $\vec{q}$, and reduce the probabilities of known inliers, respectively.
We selectively switch off one or multiple regularisation losses, \ie setting their corresponding weight ($\kappa_{\text{corr}}$, $\kappa_{\text{entropy}}$ or $\kappa_{\text{im}}$) to zero, in order to analyse their respective impact on performance.
As the results in Tab.~\ref{tab:ablation_results_regularisation} show, $\ell_{\text{im}}$ and $\ell_{\text{corr}}$ influence the results considerably.
Without either of these two regularisation losses, the number of primitives decreases by at least $40\%$, but the coverage in image space also decreases by more than $15\%$.
In addition, this yields measurably worse results \wrt all metrics based on the occlusion-aware distance. 
The mean distance of covered points increases by up to $72\%$.
By contrast, the entropy maximisation loss $\ell_{\text{entropy}}$ has little impact on the performance of our method. 
Especially when we compare the last two rows of Tab.~\ref{tab:ablation_results_regularisation}, \ie with $\ell_{\text{im}}$ and $\ell_{\text{corr}}$ enabled, we notice that the differences are within the standard deviations of each metric, and hence not significant.
We thus conclude that $\ell_{\text{im}}$ and $\ell_{\text{corr}}$ are necessary for optimal results, while $\ell_{\text{entropy}}$ may be disabled if the other two are enabled. 

\paragraph{End-to-End Network Fine-Tuning}
\label{par:ablation_finetune}
As we describe in Sec.~\ref{subsec:implementation_details}, all neural networks in our method -- \ie the depth estimator, the sampling weight estimator, and the neural network based solver -- are trained seperately. 
However, we subsequently fine-tune them jointly in an end-to-end manner.
In order to demonstrate the benefit of this additional training step, we compare the performance before and after fine-tuning for three different settings: depth input with neural solver, RGB input with numerical solver, and RGB input with neural solver.
\new{For depth input with neural solver, we additionally evaluate the influence of pre-training both networks or only one of the two networks before fine-tuning, or training both networks jointly from scratch.}
We list the results in Tab.~\ref{tab:ablation_results_refinement} and observe small but measurable improvements on almost all metrics after jointly fine-tuning all networks.
Only the number of cuboids $|\set{M}|$ increases slightly in all three settings.
\new{In addition, we observe that training the sampling network from scratch while using a pre-trained neural solver yields similar results to using pre-trained weights for both, \ie pre-training the sampling network is not necessary.
Pre-training the neural solver, however, is required, as training it from scratch jointly with the sampling network yields significantly worse accuracy.}
}

\paragraph{Number of Sampling Weight Maps}
\label{par:ablation_num_q}
To demonstrate the efficacy of predicting multiple sets of sampling weights at once (cf. Sec.~\ref{par:sampling}), we trained our approach with varying numbers $Q$ of sampling weight sets.
We evaluated these variants with depth input and present the results in Tab.~\ref{tab:ablation2_results}.
Predicting multiple sampling weight sets ($Q=4$) performs significantly better than predicting just one ($Q=1$) or two ($Q=2$) sets
{
\wrt coverage and occlusion-aware distances. 
While $Q=1$ yields the smallest number of cuboids, indicating a more parsimonious abstraction, similar results can be obtained with $Q=4$ by sampling fewer hypotheses (cf. Fig.~\ref{fig:graphs_mss:time}) or deliberately extracting fewer cuboids (cf. Fig.~\ref{fig:graphs_mss:per_cuboid}), which would be more efficient.
Besides that, the results obtained with the faster neural solver with $Q=4$ are also superior (cf. Tab.~\ref{tab:main_results}).
Increasing $Q$ further ($Q=8$) degrades the performance measurably, with an increased standard deviation of all performance metrics.
We hence conclude that $Q=4$ is the most sensible choice.
}

\begin{table}
\setlength\tabcolsep{.22em}
\renewcommand{\arraystretch}{1.1} 
	\begin{center}
	\begin{tabular}{|l|cc|cc|cc|cc|cc|}
	\cline{2-11}
	\multicolumn{1}{c|}{} &
	\multicolumn{2}{c|}{\# cuboids } &
	\multicolumn{2}{c|}{image area} & 
	\multicolumn{6}{c|}{occlusion-aware $L_2$-distance (cm)} \\
	\multicolumn{1}{c|}{} & 
	\multicolumn{2}{c|}{$|\set{M}|$ } & 
	\multicolumn{2}{c|}{covered (\%)} & 
	\multicolumn{2}{c|}{cov. mean} &
	\multicolumn{2}{c|}{mean} &
	\multicolumn{2}{c|}{AUC$_{@\SI{20}{\centi\meter}}$} 
	\\
	\hline
	$Q = 1$
	& $ \mathbf{3.7} $ & $\scriptstyle \pm \mathbf{0.24} $
	& $ 53.6 $ & $\scriptstyle \pm {2.27} $
	& $ 11.2 $ & $\scriptstyle \pm 3.32 $ 
	& $ 67.9 $ & $\scriptstyle \pm 11.68 $
	& $ 41.8 $ & $\scriptstyle \pm {2.12} $ \\
	$Q = 2$
	& $ 4.4 $ & $\scriptstyle \pm 0.58 $
	& $ 59.6 $ & $\scriptstyle \pm 3.07 $
	& $ 8.0 $ & $\scriptstyle \pm {0.85} $ 
	& $ 43.6 $ & $\scriptstyle \pm {5.77} $
	& $ 49.2 $ & $\scriptstyle \pm 3.44 $\\
	$Q = 4$
	& $ 6.6 $ & $\scriptstyle \pm 0.36 $
	& $ \mathbf{66.0} $ & $\scriptstyle \pm \mathbf{1.17} $
	& $ \mathbf{6.4} $ & $\scriptstyle \pm \mathbf{0.52} $ 
	& $ \mathbf{30.2} $ & $\scriptstyle \pm \mathbf{1.57} $
	& $ \mathbf{56.8} $ & $\scriptstyle \pm \mathbf{1.19} $\\
	$Q = 8$
	& $ 6.5 $ & $\scriptstyle \pm 2.21 $
	& $ 63.4 $ & $\scriptstyle \pm 8.61 $
	& $ 8.1 $ & $\scriptstyle \pm 2.69 $ 
	& $ 36.0 $ & $\scriptstyle \pm 17.61 $
	& $ 52.8 $ & $\scriptstyle \pm 10.24 $\\
	    \hline

	\end{tabular}
    \end{center}
\caption{\textbf{Ablation Study:} We evaluate our approach on the NYU validation set with varying numbers $Q$ of sampling weight sets with depth input, numerical solver and without refinement. {(cf. Sec.~\ref{par:ablation_num_q})}
	}
	\label{tab:ablation2_results}
\end{table}	

\begin{figure}	
		\centering
		\begin{subfigure}[t]{\linewidth}
			\centering
			\includegraphics[width=0.242\linewidth]{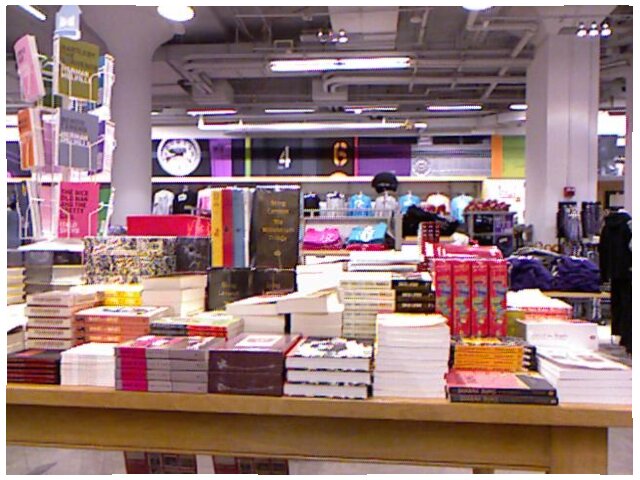}
			\includegraphics[width=0.242\linewidth]{fig/fail/final_0106_image.jpg}
			\includegraphics[width=0.242\linewidth]{fig/fail/final_0049_image.jpg}
			\includegraphics[width=0.242\linewidth]{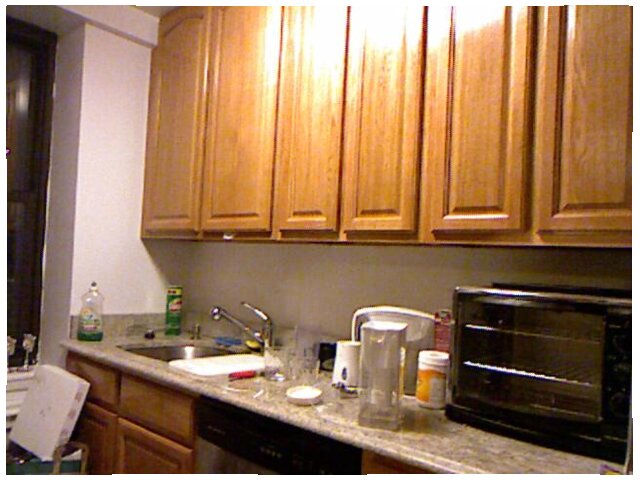}
		\end{subfigure}
		\begin{subfigure}[t]{\linewidth}
			\centering
			\includegraphics[width=0.242\linewidth]{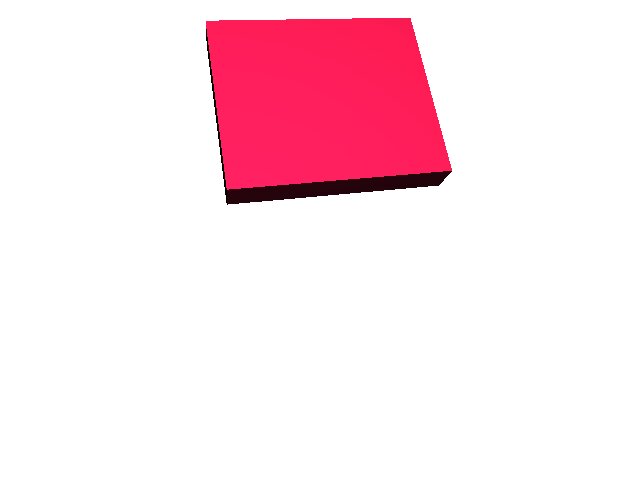}
			\includegraphics[width=0.242\linewidth]{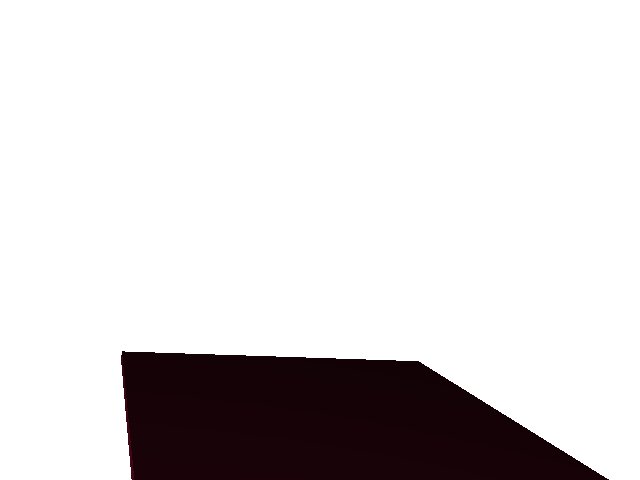}
			\includegraphics[width=0.242\linewidth]{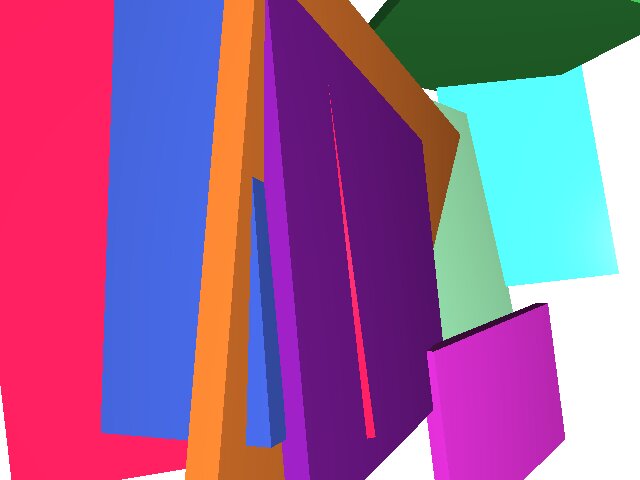}
			\includegraphics[width=0.242\linewidth]{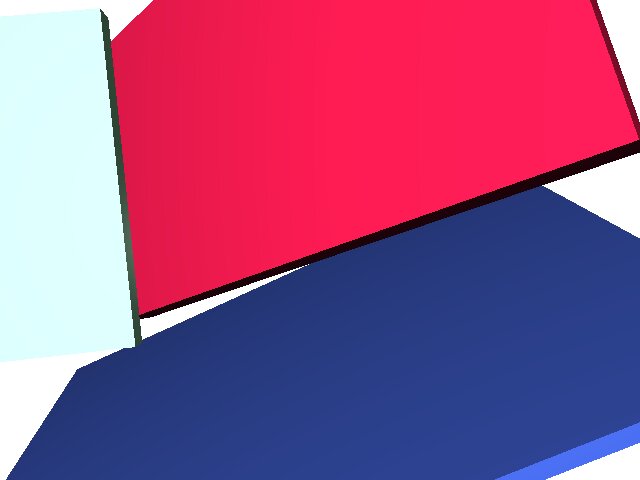}
		\end{subfigure}
  
		\caption{\new{\textbf{Failure Cases: } Examples that demonstrate limitations of our method using depth input and our neural solver (cf. Sec.~\ref{sec:limitations}).}}
		\label{fig:failure_cases}
\end{figure}

\new{
\section{Limitations and Future Work}
\label{sec:limitations}
We provide failure cases of our method, using depth input and our neural solver, in Fig.~\ref{fig:failure_cases}.
As the first two examples show, our method may not find a sufficient number of cuboids to abstract scenes with a cluttered geometry that is poorly approximated using only a few primitives.
In such cases, our method terminates early when it does not sample a cuboid hypothesis with a sufficiently large number of inliers, yielding a very small coverage of the scene. 
This could be mitigated by switching from a sequential multi-model fitting approach -- finding one cuboid after another -- to a parallel approach~\cite{kluger2024parsac}.
The last two examples in Fig.~\ref{fig:failure_cases} exhibit a different issue: sometimes our method predicts cuboids that are poorly aligned with the geometry of the scene.
This can result in over-segmentation (third example) when the method superimposes multiple ill-fitting cuboids in order to increase the number of inliers, but may also result in under-segmentation (fourth example) if a large ill-fitting cuboid prohibits adding other cuboids in the same area.
We predict cuboid parameters using minimal sets of points, which only contain limited and local information about the scene.
If these points are not sampled optimally, the resulting cuboid may thus be poorly aligned with the overall scene.
In future work, we would thus like to investigate cuboid solver approaches which incorporate additional scene context but do not suffer from the problems induced estimating cuboids from larger-than-minimal sets of points (cf. Sec.~\ref{subsec:minimal_set_size}).
Adding Manhattan-world constraints may also improve the alignment, as demonstrated recently by~\cite{vavilala2023convex}.
Lastly, while we demonstrated that our method can be applied to different domains -- real-world indoor vs. synthetic outdoor -- this does require re-training of the neural networks and determining suitable hyper-parameters. 
}

\section{Conclusion}
We present a 3D scene parser which abstracts complex real-world scenes into ensembles of simpler volumetric primitives.
It builds upon a learning-based robust estimator, which we extend in order to recover cuboids from RGB images.
To this end, we propose an occlusion-aware distance metric which enables us to correctly handle opaque scenes.
We facilitate end-to-end training by circumventing backpropagation through our {numerical optimisation based cuboid} solver, deriving the gradient of primitive parameters \wrt the input features analytically.
{In addition, we devise a new neural network based cuboid solver which is significantly faster, fully differentiable, and provides meaningful scene abstractions.}
Our algorithm neither requires known ground truth primitive parameters nor any other costly annotations. 
It can thus be straight-forwardly applied to other datasets which lack this information.
{We provide detailed empirical analyses of our method, of its individual components and their hyperparameters, using a diverse set of evaluation metrics.}
Results on the challenging real-world NYU Depth v2 dataset \new{and the synthetic outdoor SMH dataset} demonstrate that the proposed method successfully parses and abstracts complex 3D scenes.

\paragraph{Acknowledgements}
This work was supported by the BMBF grant \emph{LeibnizAILab} (01DD20003), by the DFG grant \emph{COVMAP} (RO 2497/12-2), by the DFG Cluster of Excellence \emph{PhoenixD} (EXC 2122), and by the Center for Digital Innovations (ZDIN).

\appendix
\begin{appendices}


We provide additional implementation details for our method, including descriptions of the feature extraction network (Sec.~\ref{subsec:feature_net}), sampling weight network (Sec.~\ref{subsec:sampling_net}) and neural cuboid solver network (Sec.~\ref{subsec:solver_net}) architectures.
\new{Sec.~\ref{subsec:refinement} provides the description and evaluation of an optional Expectation-Maximisation based cuboid refinement.}
In Sec.~\ref{sec:oasq}, we establish the occlusion-aware distance metric for superquadrics, which we need for our quantitative evaluation. 
\new{We show additional qualitative results for our method on the NYU Depth v2~\cite{silberman2012indoor} and Synthetic Metropolis Homographies~\cite{kluger2024parsac} datasets in Sec.~\ref{sec:qualitative}.}
In Sec.~\ref{sec:tulsiani}, we provide examples obtained using the cuboid based approach of~\cite{tulsiani2017learning}.

\section{Implementation Details}
\label{sec:implementation}

\begin{table}
\setlength\tabcolsep{.25em}
    \centering
    \begin{tabular}{c|lc|c|c|c|c}
        \multicolumn{3}{c|}{}   & \multicolumn{2}{c|}{NYU } & \multicolumn{2}{c}{SMH} \\      
        \multicolumn{3}{c|}{}                           & initial & fine- & initial & fine- \\
        \multicolumn{3}{c|}{}                           & training & tuning & training & tuning \\
        \hline
        \parbox[t]{2.5mm}{\multirow{11}{*}{\rotatebox[origin=c]{90}{training}}}
        &  learning rate            &               & $10^{-5}$ & $10^{-7}$ & $10^{-6}$ & $10^{-8}$ \\
        &  softmax scale factor               & $\alpha$      & $10$ & $1000$ & $10$ & $1000$ \\
        &  IMR weight               & $\kappa_{\text{im}}$      & $0.01$ & $0$  & $0.01$ & $0$ \\
        &  correlation weight       & $\kappa_{\text{corr}}$      & $1.0$ & $0$& $1.0$ & $0$\\
        &  entropy weight           & $\kappa_{\text{entropy}}$      & $1.0$ & $0$& $0.1$ & $0$\\
        &  max. epochs                   &  $N_e$             & \multicolumn{2}{c|}{$25$} & \multicolumn{2}{c}{$100$} \\
        &  number of instances      & $|\set{M}|$           & \multicolumn{4}{c}{$8$}  \\
        &  batch size               & $B$           & \multicolumn{4}{c}{$2$}   \\
        &  sample count             & $K$           & \multicolumn{4}{c}{$2$}  \\
        &  primitive samples  & $|\set{H}|$   & \multicolumn{4}{c}{$32$} \\
        &  {occlusion penalty threshold}         & $\tau_c$         & \multicolumn{4}{c}{ $\frac{\text{epoch}}{N_e} (\tau - 1) + 1$ }  \\
        \hline
        \parbox[t]{2.5mm}{\multirow{8}{*}{\rotatebox[origin=c]{90}{both}}}
         &  inlier threshold         & $\tau$        & \multicolumn{2}{c|}{$0.004$}   & \multicolumn{2}{c}{$0.04$}  \\
         &  minimum cuboid size     & $a_{\mathrm{min}}$           & \multicolumn{2}{c|}{$10^{-3}$}  & \multicolumn{2}{c}{$2.0$}  \\
         &  maximum cuboid size     & $a_{\mathrm{max}}$           & \multicolumn{2}{c|}{$2.0$}  & \multicolumn{2}{c}{$30.0$}  \\
         &  $f_h$ (numerical) learning rate      &               & \multicolumn{2}{c|}{$0.01$}  & \multicolumn{2}{c}{$0.5$}  \\
         &  $f_h$ (numerical) iterations         &               & \multicolumn{4}{c}{$50$}  \\
         &  minimal set size         & $C$        & \multicolumn{4}{c}{6}  \\
         & inlier softness          & $\beta$       & \multicolumn{4}{c}{5} \\
         &  sampling weight sets     & $Q$           & \multicolumn{4}{c}{$4$}  \\
        \hline
        \parbox[t]{2.5mm}{\multirow{2}{*}{\rotatebox[origin=c]{90}{test}}}
        &  primitive samples  &$|\set{H}|$    & \multicolumn{4}{c}{$4096$}  \\
         & occlusion penalty threshold         & $\tau_c$        & \multicolumn{4}{c}{$2 \cdot \tau$}  \\
    \end{tabular}
    \caption{{\textbf{User definable parameters} of our approach and the values we chose for our experiments. {We distinguish between values used either for training, testing, or both. For training, we further distinguish between values used either during the initial training or during the end-to-end fine-tuning.} \new{As the SMH dataset contains a significantly larger number of scenes, we defined one epoch to contain $500$ randomly samples scenes.} }}
    \label{tab:parameters}
\end{table}

\subsection{Feature Extraction Network}
\label{subsec:feature_net}
We employ the \emph{Big-to-Small} (BTS)~\cite{lee2019big} depth estimation CNN as our feature extraction network.
We use a variant of their approach using a DenseNet-161~\cite{huang2017densely} as the base network.
It is pre-trained on NYU Depth v2~\cite{silberman2012indoor} and achieves state-of-the-art results for monocular depth estimation.
Please refer to~\cite{lee2019big} for details.

\subsection{Sampling Weight Network}
\label{subsec:sampling_net}
\begin{figure*}	
\centering
\includegraphics[width=0.99\linewidth]{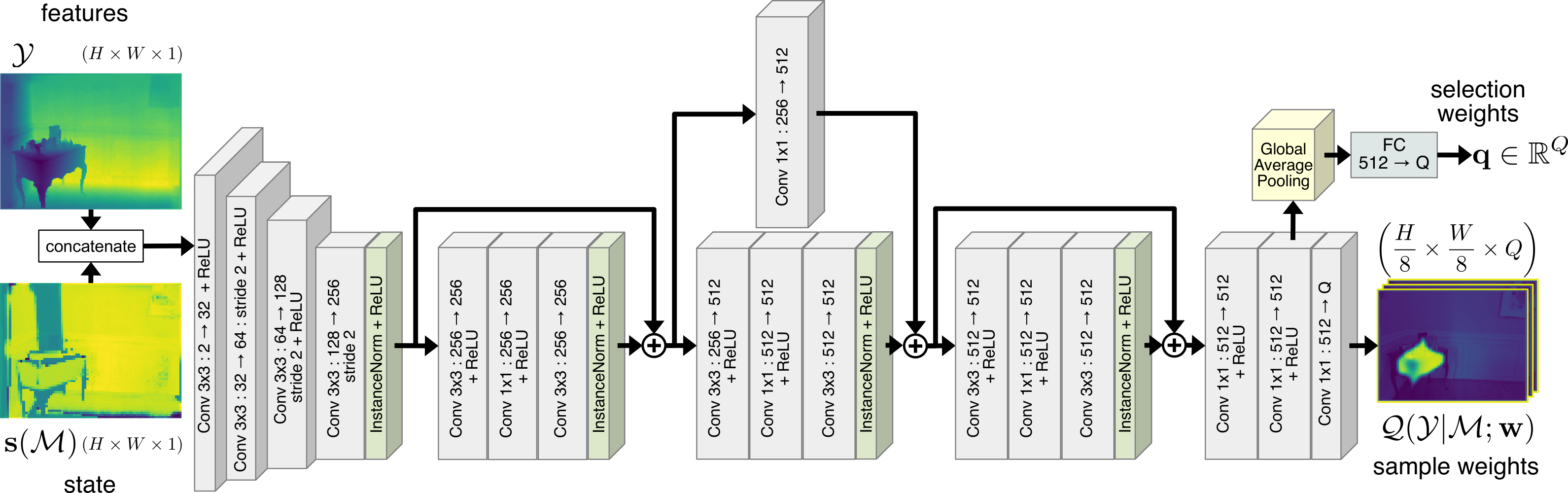}
\caption{ \textbf{Sampling Weight Network Architecture:}
We stack features $\set{Y}$ and state $\vec{s}$ into a tensor of size $H \times W \times 2$, with image width $W$ and height $H$. We feed this tensor into a neural network consisting of $3\times3$ and $1\times1$ convolutional layers, instance normalisation layers~\cite{ulyanov2016instance}, and ReLU activations~\cite{he2015delving} arranged as residual blocks~\cite{he2016deep}. The last convolutional layer predicts the sampling weight sets $\set{Q}(\set{Y}|\set{M}; \vec{w})$. We apply global average pooling and a fully-connected layer to the output of the penultimate convolutional layer in order to predict the selection weights $\vec{q}$. This architecture is based on~\cite{KluBra2020, Brachmann2019:NGRANSAC}.
}
\label{fig:sampling_net}
\end{figure*} 
For prediction of sampling weights, we use a neural network based on the architecture for scene coordinate regression used in~\cite{Brachmann2019:NGRANSAC}. 
Refer to Fig.~\ref{fig:sampling_net} for an overview.
The input of the network is a concatenation of features $\set{Y}$ and state $\vec{s}$ of size $H \times W \times 2$, with image width $W$ and height $H$.
When using ground truth depth as input, we normalise $\set{Y}$ by the mean and variance of the training set.
When using the features predicted by the feature extraction network, we add a batch normalisation~\cite{ioffe2015batch} layer between the two networks in order to take care of input normalisation.
We augmented the network of~\cite{Brachmann2019:NGRANSAC} with instance normalisation~\cite{ulyanov2016instance} layers, which proved crucial for the ability of the network to segment distinct structures in the scene.
The network thus consists of $3\times3$ and $1\times1$ convolutional layers, instance normalisation layers~\cite{ulyanov2016instance}, and ReLU activations~\cite{he2015delving} arranged as residual blocks~\cite{he2016deep}.
While most convolutions are applied with stride one, layers two to four use a stride of two, resulting in a final output spatially subsampled by a factor of eight \wrt the input.
We apply sigmoid activation to the last convolutional layer to predict the sampling weight sets $\set{Q}(\set{Y}|\set{M}; \vec{w})$ with size $\frac{H}{8} \times \frac{W}{8} \times Q$, \ie $Q$ sets of sampling weights for each input.
Additionally, we apply global average pooling and a fully-connected layer with sigmoid activation to the output of the penultimate convolutional layer in order to predict the selection weights $\vec{q}$.

\subsection{Neural Cuboid Solver}
\label{subsec:solver_net}
\begin{figure*}	
\centering
\includegraphics[width=\linewidth]{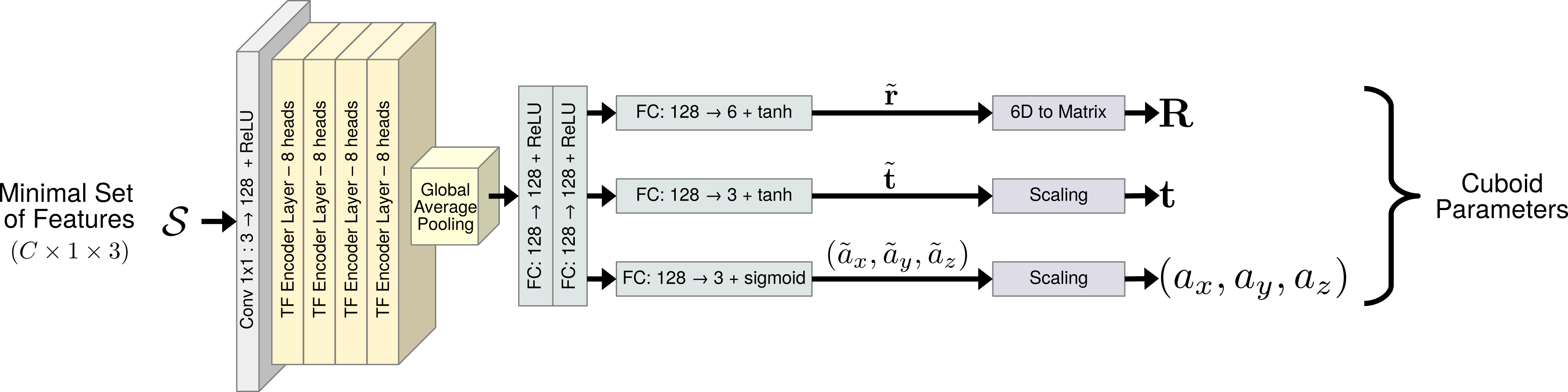}
\caption{{ \textbf{Neural Cuboid Solver Network Architecture:}
We feed a minimal set of $C$ features (3D points) into a $1\times 1$ convolution layer, followed by a stack of four transformer encoder layers~\cite{vaswani2017attention}.
After reduction to a single feature vector via average pooling, we apply two additional fully connected layers.
Three separate fully connected layers then predict rotation, translation and size of the cuboid.
Size $(a_x, a_y, a_z)$ and translation $\vec{t}$ are then scaled according to the minimum and maximum cuboid size $(a_{\mathrm{min}}, a_{\mathrm{max}})$.
The network predicts the rotation as a 6D vector~\cite{zhou2019continuity}, which we then convert into a rotation matrix $\vec{R}$.
}}
\label{fig:solver_net}
\end{figure*} 

{
As described in Sec.~3.2.3, we optionally use a neural network to predict cuboid parameters from a minimal set of $C$ features (3D points) $\set{S} = \{\vec{y}_1, \dots, \vec{y}_C\} \subset \set{Y}$.
Fig.~\ref{fig:solver_net} shows an overview of the network architecture.
We first compute the mean of $\set{S}$:
\begin{equation}
    \mu_{\set{S}} = \frac{1}{|\set{S}|} \sum_{\vec{y} \in \set{S}} \vec{y} \, ,
\end{equation}
and subtract it from $\set{S}$ before feeding it into network architecture depicted in Fig.~\ref{fig:solver_net}.
After the last set of fully connected layers, we get tentative cuboid parameters $(\tilde{a}_x, \tilde{a}_y, \tilde{a}_z, \tilde{\vec{r}}, \tilde{\vec{t}})$.
We convert the 6D rotation vector $\tilde{\vec{r}}$ to a rotation matrix $\vec{R}$ as described in~\cite{zhou2019continuity}.
Due to the sigmoid activation function, size parameters $\tilde{a}$ are predicted in the range $[0, 1]$.
We thus scale them using minimum and maximum size values $a_{\mathrm{min}}$ and $a_{\mathrm{max}}$:
\begin{equation}
    {a} = \tilde{a} * (a_{\mathrm{max}} - a_{\mathrm{min}}) + a_{\mathrm{min}} \, .
\end{equation}
Similarly, we rescale the translation $\tilde{\vec{t}}$ using $a_{\mathrm{max}}$ and $\mu_{\set{S}}$:
\begin{equation}
    \vec{t} = \tilde{\vec{t}} * a_{\mathrm{max}} + \mu_{\set{S}} \, .
\end{equation}
}

\subsection{Neural Network Training}
We implement our method using PyTorch~\cite{paszke2017automatic} version 1.10.0. 
We use the Adam~\cite{KingmaB14} optimiser to train the neural networks.
In order to avoid divergence induced by bad hypothesis samples frequently occurring at the beginning of training, we clamp losses to an absolute maximum value of $0.3$.
{Training was performed using single RTX 3090 GPUs.}

{
\subsubsection{Sampling Weight Network}
First, we train the sampling weight network by itself using ground truth depth for up to $25$ epochs with a learning rate of $10^{-5}$ \new{(NYU) or for up to $100$ epochs with a learning rate of $10^{-6}$ (SMH). The SMH dataset contains a significantly larger number of scenes than NYU, hence we define one epoch to contain 500 randomly samples scenes.}
We use a batch size of $B=2$ and utilise the regularisation losses described in Sec.~3.5 of the main article.

\begin{table*}
\setlength\tabcolsep{.5em}
\renewcommand{\arraystretch}{1.1} 
	\begin{center}
	\begin{tabular}{|l|c|ccc|ccc|cc|cc|cc|cc|}
	\cline{2-16}
	  \multicolumn{1}{c|}{} & 
        mean &
	\multicolumn{2}{c}{{\# primitives}} &\multirow{ 2}{*}{$\downarrow$}  &
	\multicolumn{2}{c}{image area} & \multirow{ 2}{*}{$\uparrow$} &
	\multicolumn{4}{c|}{OA-$L_2$ mean (cm) $\downarrow$} &
	\multicolumn{4}{c|}{OA-$L_2$ AUC (\%) $\uparrow$} \\
        \multicolumn{1}{c|}{} &
        time (s) & 
	\multicolumn{2}{c}{$|\set{M}|$} & &
	\multicolumn{2}{c}{covered (\%)} & &
	\multicolumn{2}{c|}{covered points}      & 
	\multicolumn{2}{c|}{all points} &
	\multicolumn{2}{c|}{$@\SI{20}{\centi\meter}$} & 
	\multicolumn{2}{c|}{${@\SI{5}{\centi\meter}}$}  \\
	\hline
	with EM refinement
	& $9.02$
	& $ \mathbf{6.5} $ & \multicolumn{2}{c|}{$\scriptstyle \pm \mathbf{0.25} $}
	& $ \mathbf{67.9} $ & \multicolumn{2}{c|}{$\scriptstyle \pm \mathbf{1.12} $}
	& $ 6.6 $ & $\scriptstyle \pm \mathbf{0.37} $
	& $ 30.8 $ & $\scriptstyle \pm 1.61 $
	& $ \mathbf{57.7} $ & $\scriptstyle \pm \mathbf{0.96} $
	& $ \mathbf{35.1} $ & $\scriptstyle \pm \mathbf{0.72} $ \\
	
	w/o EM refinement
	& $\mathbf{7.63}$
	& $ 6.6 $ & \multicolumn{2}{c|}{$\scriptstyle \pm 0.36 $}
	& $ {66.0} $ & \multicolumn{2}{c|}{$\scriptstyle \pm {1.17} $}
	& $ \mathbf{6.4} $ & $\scriptstyle \pm 0.52 $ 
	& $ \mathbf{30.2} $ & $\scriptstyle \pm \mathbf{1.57} $
	& $ {56.8} $ & $\scriptstyle \pm {1.19} $
	& $ {29.9} $ & $\scriptstyle \pm {0.87} $ \\
	    \hline

	\end{tabular}
    \end{center}
\caption{{\textbf{Ablation Study:} We evaluate our method on the NYU dataset with an additional EM based refinement step (cf. Sec.~\ref{subsec:refinement}). 
	}}
	\label{tab:em_results}
\end{table*}	

\subsubsection{Neural Cuboid Solver}
In order to train the neural solver individually, we generate synthetic training data on the fly.
We create cuboids with random size parameters $a \sim \mathcal{U}(0.01, 2)$ and translation $\vec{t} \sim [\mathcal{U}(-5, 5), \mathcal{U}(-5, 5),\mathcal{U}(0.5, 10)]$ 
\new{for NYU, or $a \sim \mathcal{U}(2.0, 30)$ and $\vec{t} \sim [\mathcal{U}(-50, 50), \mathcal{U}(-50, 50),\mathcal{U}(1.0, 50)]$ for SMH. We sample }
 rotation in axis-angle notation $\vec{r} = \alpha \cdot \frac{\vec{d}}{\|\vec{d}\|_2}$, with angle $\alpha \sim \mathcal{U}(-\pi, \pi)$ and direction $\vec{d} = [d_x, d_y, d_z] \sim \mathcal{U}(0, 1)$.
We then randomly sample a minimal set $\set{S}$ of $C$ points on the three cuboid sides which face a virtual camera located at origin.
On each cuboid side, the point coordinates are sampled uniformly.
The probability of a point being sampled on one of the cuboid sides is proportional to the surface area of the side, multiplied by the cosine of the angle of incidence.
This way we can approximate the distribution of points one would expect from a depth sensor in a similar setting.
Using this data, we train the neural solver with $C=6$, batch size $4096$ and learning rate $10^{-4}$ for $150000$ iterations.
We define the loss as the mean squared distance between the surface of the cuboid with parameters $\vec{h}$ predicted by the network, and the points in $\set{S}$:
\begin{equation}
    \ell_{\text{solver}} = \frac{1}{|\set{S}|} \sum_{y \in \set{S}} d(\vec{h}, \vec{y})^2 \, .
\end{equation}

\subsubsection{Fine-tuning}
After training each network individually, we fine-tune the whole pipeline end-to-end with a reduced learning rate of $10^{-7}$ for up to $25$ epochs \new{(NYU) or with a learning rate of $10^{-8}$ for up to $100$ epochs (SMH).}
We disable all regularisation losses as they prove to be detrimental to overall performance if used during end-to-end fine-tuning.
}

{
\section{EM Refinement}
\label{subsec:refinement}
In the field of robust model fitting, it is common to perform additional refinement steps which utilise more than just a minimal set of data points~\cite{KluBra2020,barath2019progressive,barath2018multi,Brachmann2019:NGRANSAC,barath2018graph,Fischler1981:RANSAC,chum2003locally}.
After selecting a new cuboid, we thus optionally perform a refinement step which optimises the cuboid parameters, using all features $\set{Y}$.
Based on the Expectation-Maximisation (EM) algorithm~\cite{dempster1977maximum}, we assume that each feature $\vec{y} \in \set{Y}$ belongs either to exactly one of the cuboids $\vec{h} \in \set{M}$, or to none.
We describe the feature-to-cuboid associations via a hidden variable $\set{V}$, \ie $(\vec{h},\vec{y}) \in \set{V}$ if feature $\vec{y}$ belongs to cuboid $\vec{h}$.
These associations are not known beforehand.
We thus want to find a refined set of cuboids $\set{M}'$ which maximises the expected log-likelihood $Q$ of our features $\set{Y}$, given the current set of cuboids $\set{M}$, over all possible $\set{V}$:
\begin{equation}
    Q(\set{M}'|\set{M}) = \mathbb{E}_{\set{V}|\set{Y},\set{M}} [\log p(\set{Y},\set{V}|\set{M'})] \, .
    \label{eq:q_def}
\end{equation}
As we assume independence of all features in $\set{Y}$, the likelihood expands to:
\begin{equation}
   p(\set{Y},\set{V}|\set{M'}) = \prod_{\vec{y} \in \set{Y}} \left( \sum_{\vec{h} \in \set{M}'} p(\vec{y}|\vec{h}) \cdot \mathbbm{1}_{\set{V}}((\vec{h},\vec{y})) \right) \, ,
\end{equation}
and Eq.~\ref{eq:q_def} expands to:
\begin{equation}
    Q(\set{M}'|\set{M}) = \sum_{\vec{h} \in \set{M}'}\sum_{\vec{y} \in \set{Y}} \log(p(\vec{y}|\vec{h})) \cdot p((\vec{h},\vec{y}) \in \set{V}|\set{M})\, .
    \label{eq:q_expanded}
\end{equation}
We express the association probability via the posterior:
\begin{equation}
    p((\vec{h},\vec{y}) \in \set{V}|\set{M}) = \frac{p(\vec{y}|\vec{h})) \cdot p(\vec{h})}{\sum_{\vec{h}' \in \set{M}} p(\vec{y}|\vec{h'}) \cdot p(\vec{h'})} \, ,
\end{equation}
and define the likelihood as a normal distribution \wrt the point-to-cuboid distance:
\begin{equation}
    p(\vec{y}|\vec{h}) = \frac{1}{\sqrt{2\pi \sigma^2_{\vec{h}}}} \exp{\left(\frac{-d^2(\vec{h}, \vec{y})}{2\sigma_{\vec{h}}^2}\right)} \, .
\end{equation}
Starting from an initial set of values for $\set{M}$, $p(\vec{h})$ and $\sigma_{\vec{h}}$, we approximate $\set{M}' \in \argmax_{\hat{\set{M}}} Q(\hat{\set{M}}|\set{M})$ via iterative numerical optimisation, \ie gradient descent.
In this case, the E-step is equivalent to the forward pass, \ie computing the value of the $Q$-function, and the M-step is approximated by the backward pass.
Once the optimisation loop converges or reaches an iteration limit, we continue with robust fitting (Sec.~3.2 of the main article) for the next cuboid.

As the results in Tab.~\ref{tab:em_results} show, 
the refinement step improves the AUC$_{@\SI{5}{\centi\meter}}$ metric measurably by $17.3\%$, \ie the fit of the cuboids to close-by points is improved.
Its impact on the other metrics, however, is either small or slightly detrimental.
Moreover, the refinement step adds substantial computational overhead, increasing the computation time by $18.2\%$.
We thus conclude that it does not provide a tangible benefit for the scene abstraction task and is likely not worth the additional computation time.

}

\section{OA Distance for Superquadrics}
\label{sec:oasq}
The surface of a superellipsoid~\cite{barr1981superquadrics}, which is the type of superquadric used in~\cite{paschalidou2019superquadrics,Paschalidou2020:Decomposition}, can be described by its inside-outside function:
\begin{equation}
    f_{\text{sq}}(x,y,z) = \left( \left( \frac{x}{a_x} \right)^{ \frac{2}{\epsilon_2}} \!\! + \left( \frac{y}{a_y} \right)^{\frac{2}{\epsilon_2}} \right)^{\frac{\epsilon_1}{\epsilon_2}} \!\! + \left( \frac{z}{a_y} \right)^{\frac{2}{\epsilon_1}} \!\! -1 \, ,
\end{equation}
with $\epsilon_1, \epsilon_2$ describing the shape of the superquadric, and $a_x, a_y, a_z$ describing its extent along the canonical axes. 
If $f_{\text{sq}} = 0$, the point $(x,y,z)$ resides on the superquadric surface. 
For $f_{\text{sq}} > 0$ and $f_{\text{sq}} < 0$, it is outside or inside the superquadric, respectively. 
Alternatively, a point on the superquadric surface can be described by:
\begin{equation}
    \vec{p}(\eta, \omega) = 
    \begin{bmatrix}
    a_x \cos^{\epsilon_1}(\eta) \cos^{\epsilon_2}(\omega) \\
    a_y \cos^{\epsilon_1}(\eta) \sin^{\epsilon_2}(\omega) \\
    a_z \sin^{\epsilon_1}(\eta) 
    \end{bmatrix} \, ,
    \label{eq:sq_io}
\end{equation}
parametrised by two angles $\eta,\omega$. The surface normal at such a point is defined as:
\begin{equation}
    \vec{n}(\eta, \omega) = 
    \begin{bmatrix}
    a_x^{-1} \cos^{2-\epsilon_1}(\eta) \cos^{2-\epsilon_2}(\omega) \\
    a_y^{-1} \cos^{2-\epsilon_1}(\eta) \sin^{2-\epsilon_2}(\omega) \\
    a_z^{-1} \sin^{2-\epsilon_1}(\eta) 
    \end{bmatrix} \, .
    \label{eq:sq_normal}
\end{equation}
Unfortunately, no closed form solution exists for calculating a point-to-superquadric distance, which we would need in order to compute the occlusion-aware distance metric as described in Sec. 3.2. 
We therefore approximate it by sampling points and determining occlusion and self-occlusion using Eq.~\ref{eq:sq_io} and Eq.~\ref{eq:sq_normal}.

\subsection{Occlusion}
Given a 3D point $\vec{y} = \left[x,\,y,\,z\right]\tran$ and a camera centre $\vec{c} = \left[0,\,0,\,0\right]\tran$, we sample $L$ points uniformly on the line of sight:
\begin{equation}
    \set{Y}_{\text{los}} = \{\frac{1}{L} \vec{y}, \, \frac{2}{L} \vec{y}, \, \dots, \vec{y} \} = 
    \{\vec{y}_1, \, \vec{y}_2, \, \dots, \vec{y}_L \} \, .
\end{equation}
For a superquadric $\vec{h} = (\epsilon_1, \epsilon_2, a_x, a_y, a_z, \vec{R}, \vec{t})$, we transform all points in $\set{Y}_{\text{los}}$ into the superquadric-centric coordinate system:
\begin{equation}
    \vec{\hat{y}} = \vec{R}(\vec{y}-\vec{t}) \, ,
\end{equation}
and determine whether they are inside or outside of the superquadric:
\begin{equation}
    \set{F}_{\text{los}} = \{ f_{\text{sq}}(\vec{\hat{y}}_1), \dots, f_{\text{sq}}(\vec{\hat{y}}_L) \} \, .
\end{equation}
We then count the sign changes in $\set{F}_{\text{los}}$: if there are none, $\vec{y}$ is not occluded by the superquadric; otherwise, it is:
\begin{equation}
    \chi_{\text{sq}}(\vec{y}, \vec{h}) =
\begin{cases}
1 ~&\text{ if }~ \vec{h}  \text{ occludes }  \vec{y}, \\
0 ~&\text{ else } \, .
\end{cases}
\end{equation}

\subsection{Self-Occlusion}
Using the source code provided by the authors of~\cite{paschalidou2019superquadrics}, we sample $N$ points $\vec{p}_i$ uniformly on the surface of superquadric $\vec{h}$ and determine their corresponding surface normals $\vec{n}_i$. 
For each point, we compute the vector $\vec{v}_i = \vec{p}_i - \hat{\vec{c}}$, with $\vec{\hat{c}} = \vec{R}(\vec{c}-\vec{t})$ being the camera centre in the superquadric-centric coordinate system. 
If $\vec{v}_i \tran \vec{n}_i = 0$, \ie the two vectors are orthogonal, $\vec{p}_i$ lies on the rim of the superquadric, which partitions it into a visible and an invisible part~\cite{jaklic2000segmentation}. 
Assuming $\vec{n}_i$ points outward, it follows that $\vec{p}_i$ is invisible if $\vec{v}_i \tran \vec{n}_i > 0$ and $f_{\text{sq}}(\hat{\vec{c}}) > 0$, in which case we discard it. 
We denote the set of visible points of $\vec{h}$ as $\set{P}(\vec{h})$.

\subsection{Occlusion-Aware Distance}
We define the distance of a point $\vec{y}$ to one superquadric $\vec{h}$ as the minimum distance to any of its visible points $\set{P}$:
\begin{equation}
    d_{\text{sq}}(\vec{h}, \vec{y}) = \min_{\vec{p} \in \set{P}(\vec{h})} \| \vec{p}-\vec{y} \|_2 \, .
\end{equation}
Similarly to cuboids, we compute the distance of $\vec{y}$ to the most distant occluding superquadric, given a set of superquadrics $\set{M}$:
\begin{equation}
    d_{\text{o,sq}}(\set{M}, \vec{y}) = 
\max_{\vec{h} \in \set{M}} 
\left( \,
\chi_{\text{sq}}(\vec{y}, \vec{h}) \cdot d_{\text{sq}}(\vec{h}, \vec{y}) 
\, \right) \, ,
\end{equation}
and corresponding occlusion-aware distance:
\begin{equation}
    d_{\text{oa,sq}}(\set{M}, \vec{y}) = 
    \max 
    \left( 
    \min_{\vec{h} \in \set{M}} d_{\text{sq}}(\vec{h}, \vec{y}), \; d_{\text{o,sq}}(\set{M}, \vec{y}) 
    \right) \, . \nonumber
\end{equation}

\new{
\section{Qualitative Results}
\label{sec:qualitative}
\setlength{\imgwidth}{0.305\linewidth}
\begin{figure}	

\setlength\tabcolsep{0.1em}
\renewcommand{\arraystretch}{1.} 
	\begin{center}
        \begin{tabular}{cccc}
             &
             Ground Truth & 
             Neural Solver & 
             Numerical Solver  \\
 
\parbox[t]{3mm}{\rotatebox[origin=l]{90}{{\hspace{1.1em} RGB Image}}}
& \includegraphics[width=\imgwidth]{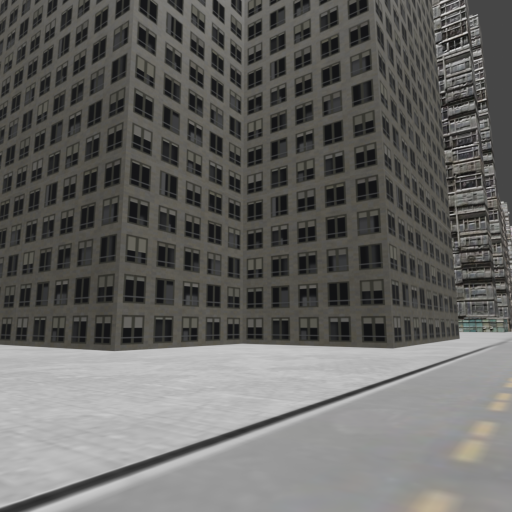} 
& \includegraphics[width=\imgwidth]{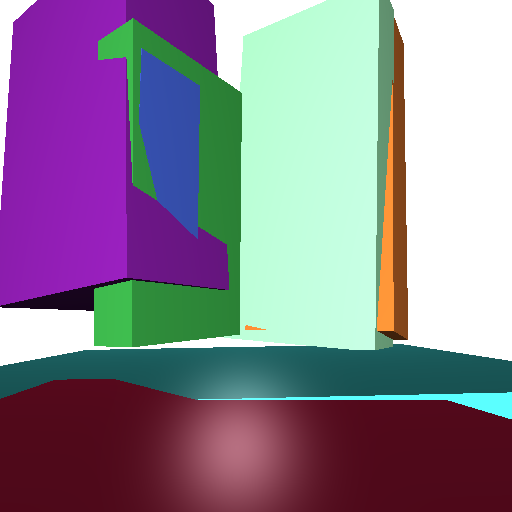} 
& \includegraphics[width=\imgwidth]{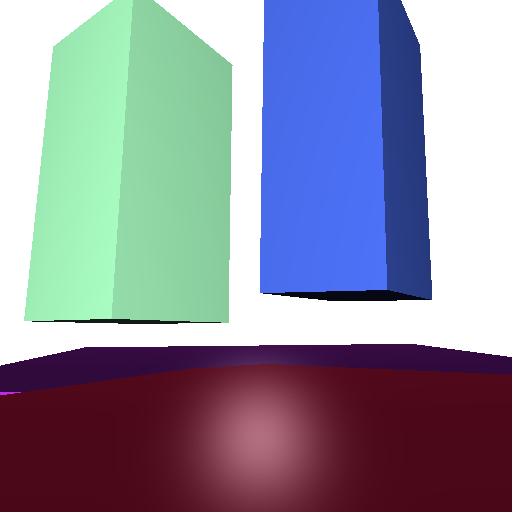} \\
\parbox[t]{3mm}{\rotatebox[origin=l]{90}{{\hspace{2em} Depth}}}
& \includegraphics[width=\imgwidth]{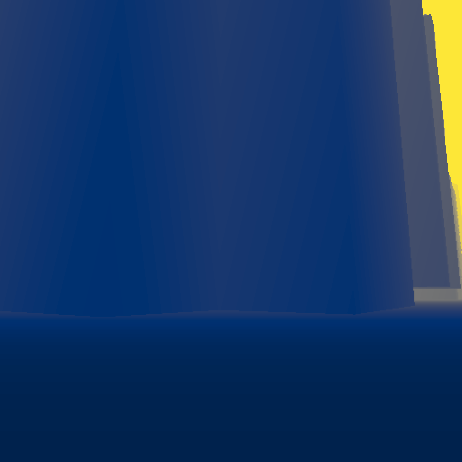} 
& \includegraphics[width=\imgwidth]{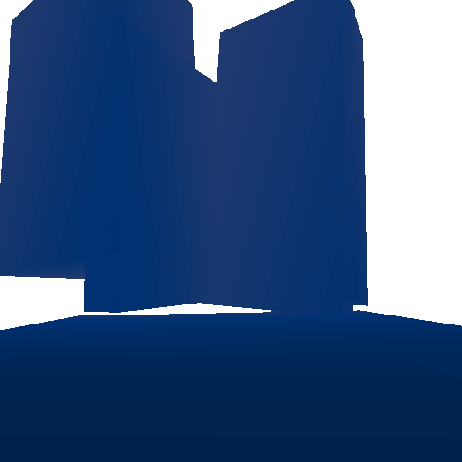} 
& \includegraphics[width=\imgwidth]{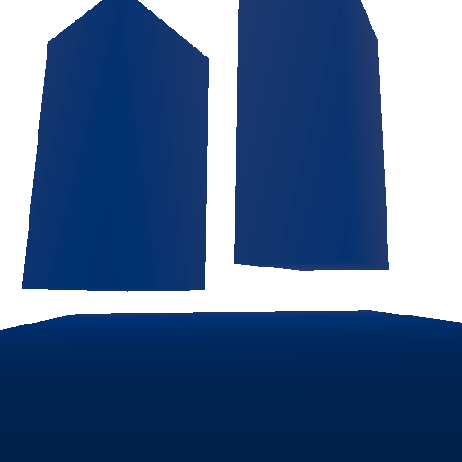} \\
 
& \includegraphics[width=\imgwidth]{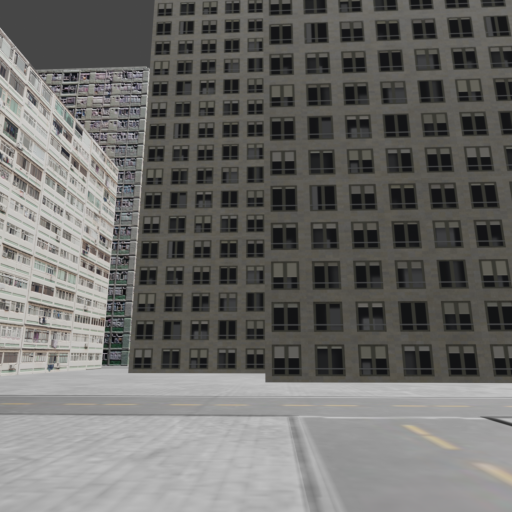} 
& \includegraphics[width=\imgwidth]{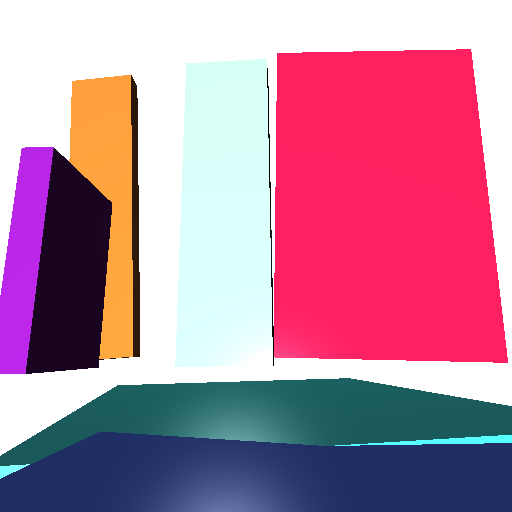} 
& \includegraphics[width=\imgwidth]{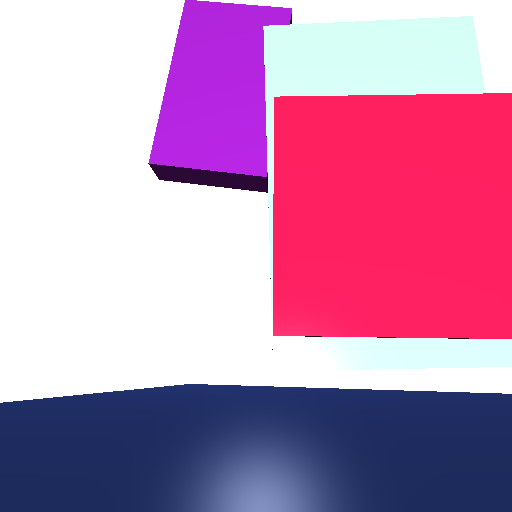} \\
& \includegraphics[width=\imgwidth]{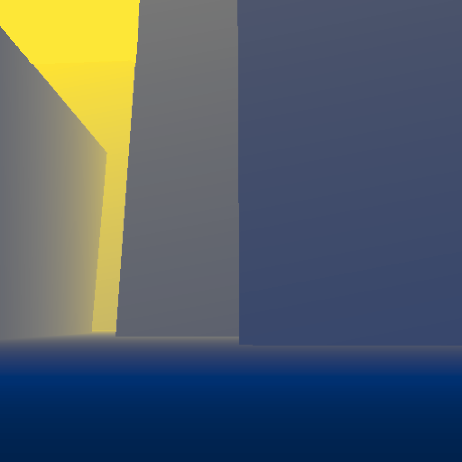} 
& \includegraphics[width=\imgwidth]{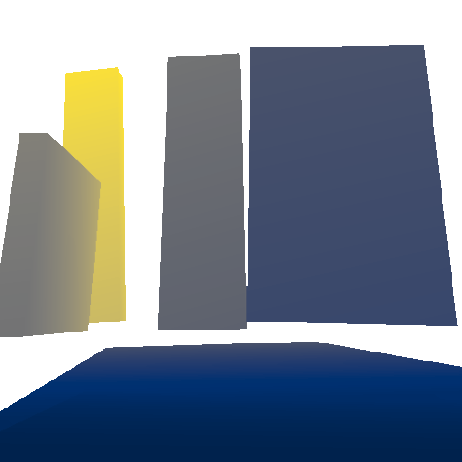} 
& \includegraphics[width=\imgwidth]{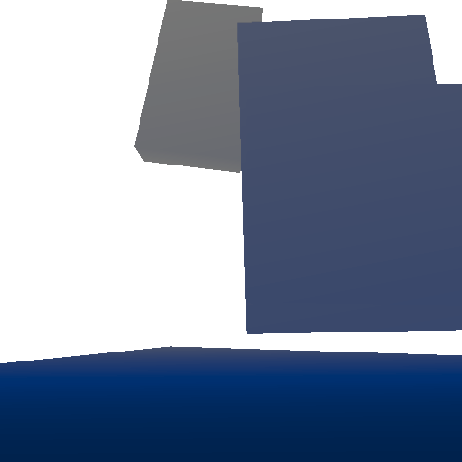} \\
 
& \includegraphics[width=\imgwidth]{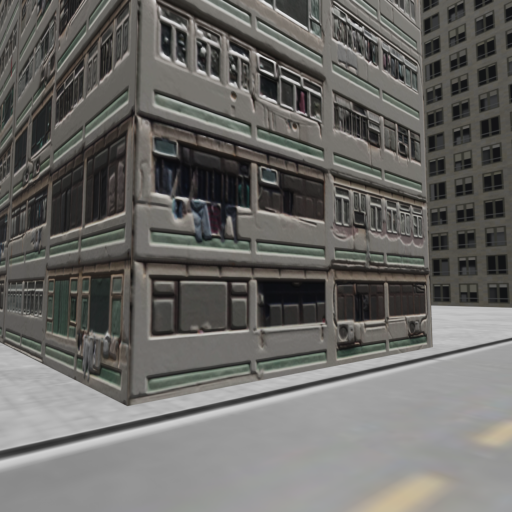} 
& \includegraphics[width=\imgwidth]{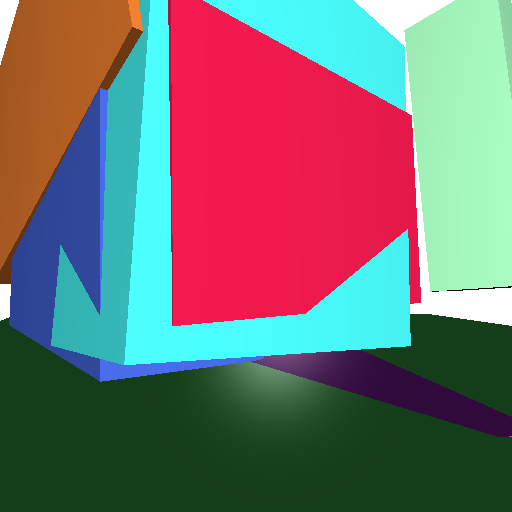} 
& \includegraphics[width=\imgwidth]{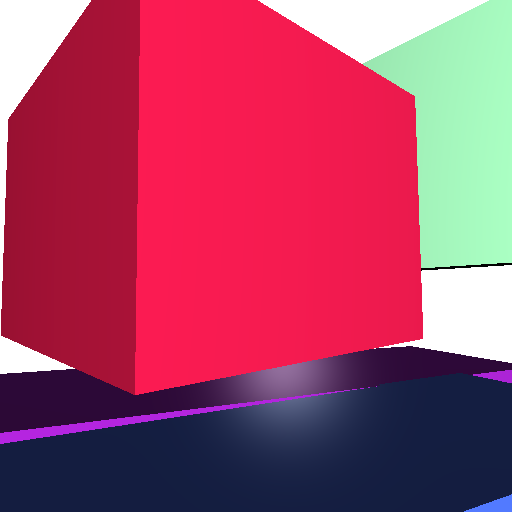} \\
& \includegraphics[width=\imgwidth]{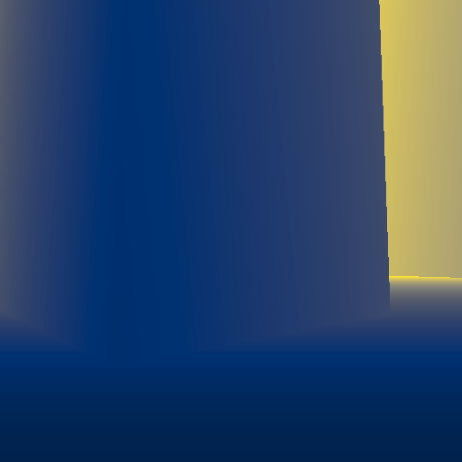} 
& \includegraphics[width=\imgwidth]{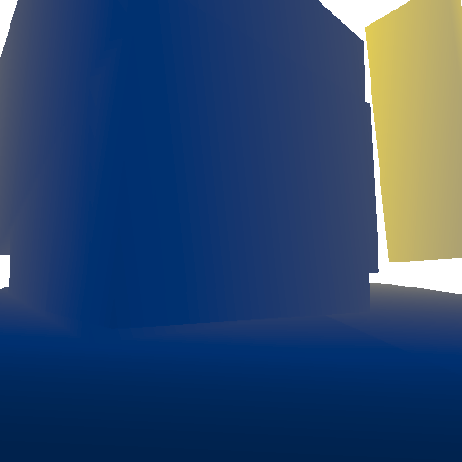} 
& \includegraphics[width=\imgwidth]{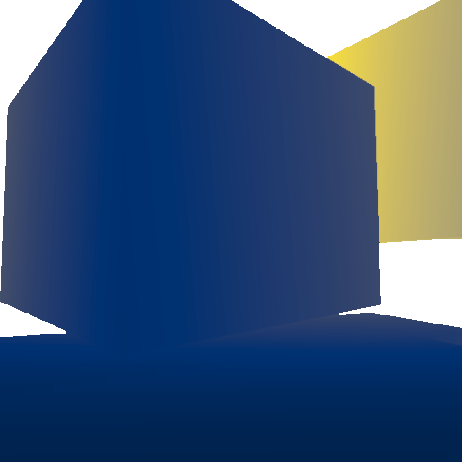} \\
 
& \includegraphics[width=\imgwidth]{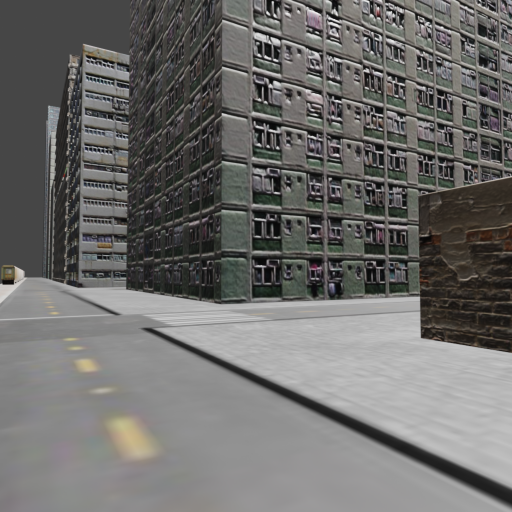} 
& \includegraphics[width=\imgwidth]{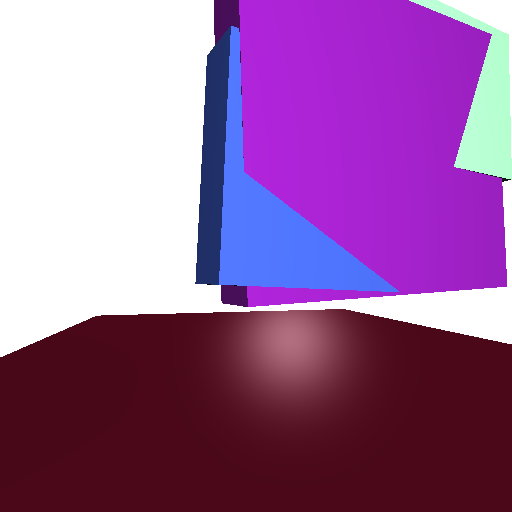} 
& \includegraphics[width=\imgwidth]{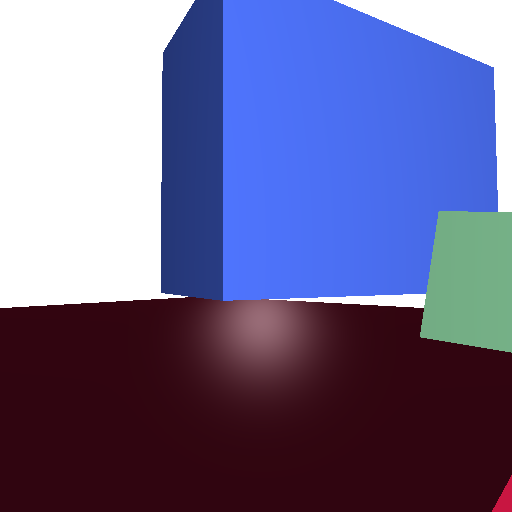} \\
& \includegraphics[width=\imgwidth]{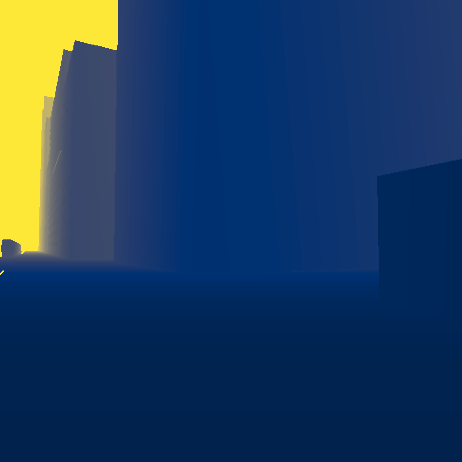} 
& \includegraphics[width=\imgwidth]{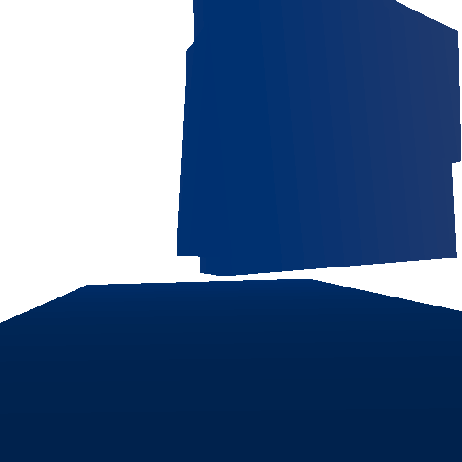} 
& \includegraphics[width=\imgwidth]{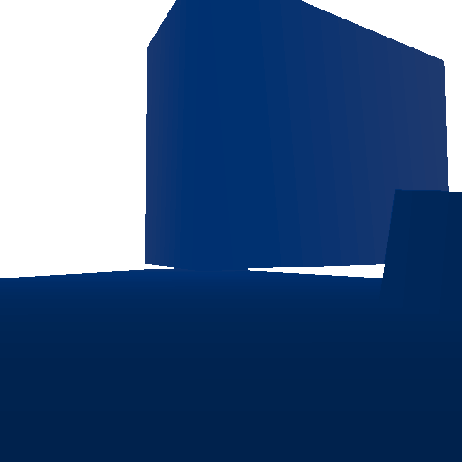} \\

        \end{tabular}
     
		\caption{\textbf{SMH: Qualitative Results.}
		\new{
		First column: Ground truth images and depth. 
		Columns 2-3: Cuboids obtained with our proposed methods, using ground truth depth as input. 
		}}
		\label{fig:smh_examples}
  \end{center}
\end{figure}   
\begin{figure*}	

\setlength\tabcolsep{0.1em}
\renewcommand{\arraystretch}{1.} 
	\begin{center}
        \begin{tabular}{ccoooobb}
             &
             & \multicolumn{4}{c}{\cellcolor{ours}{\textbf{Ours}}} 
             & \multicolumn{2}{c}{\cellcolor{baseline}{Baselines}} \\
             &
             & \multicolumn{2}{c}{\cellcolor{ours}{Depth Input}} 
             & \multicolumn{2}{c}{\cellcolor{ours}{RGB Input}} 
             & \multicolumn{2}{c}{\cellcolor{baseline}{Depth Input}} \\
             &
             Ground Truth & 
             Neural Solver & 
             Numerical Solver & 
             Neural Solver & 
             Numerical Solver & 
             CONSAC~\cite{KluBra2020} & 
             SQ-Parsing~\cite{paschalidou2019superquadrics} \\

                \parbox[t]{3mm}{{\rotatebox[origin=l]{90}{RGB Image}}}&
			\includegraphics[width=0.136\linewidth]{fig/examples_new/rgb/final_0642_image.png} &
			\includegraphics[width=0.136\linewidth]{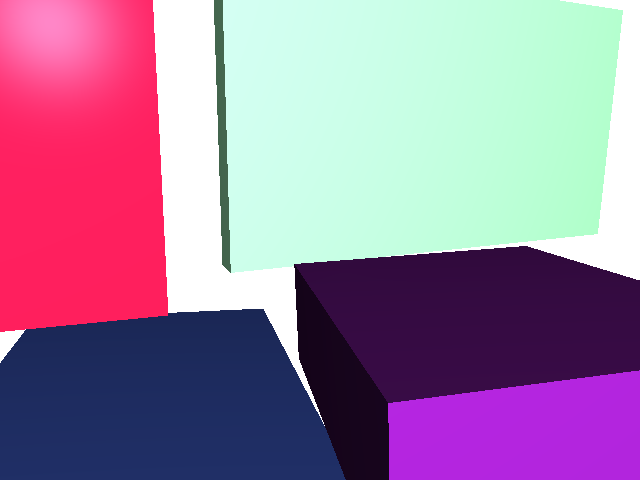} &
			\includegraphics[width=0.136\linewidth]{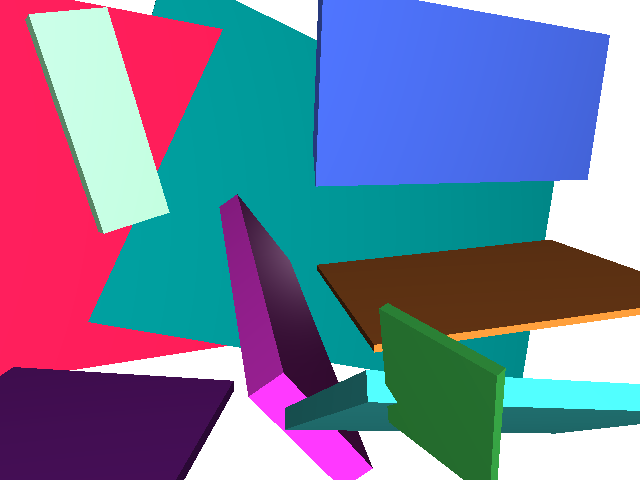} &
			\includegraphics[width=0.136\linewidth]{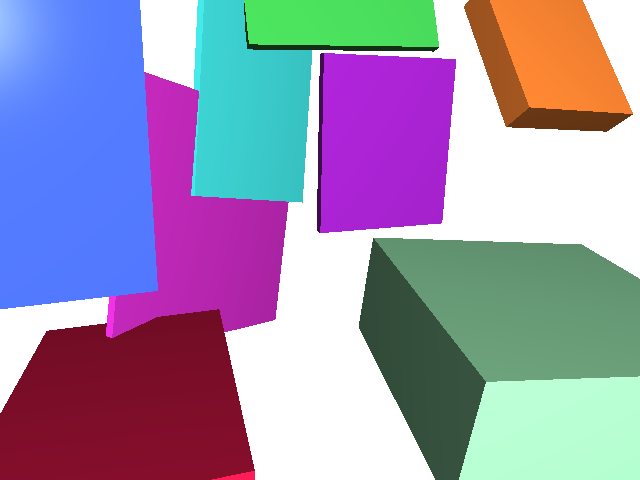} &
			\includegraphics[width=0.136\linewidth]{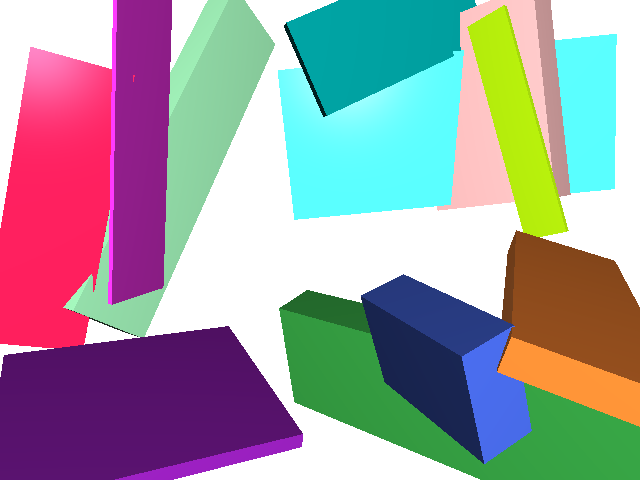} &
			\includegraphics[width=0.136\linewidth]{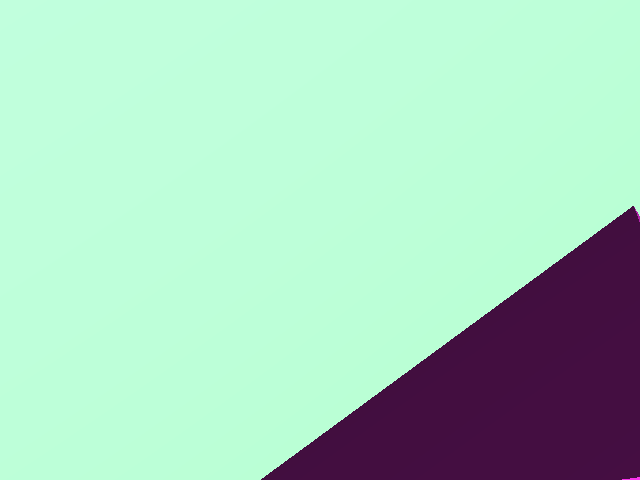} &
			\includegraphics[width=0.136\linewidth]{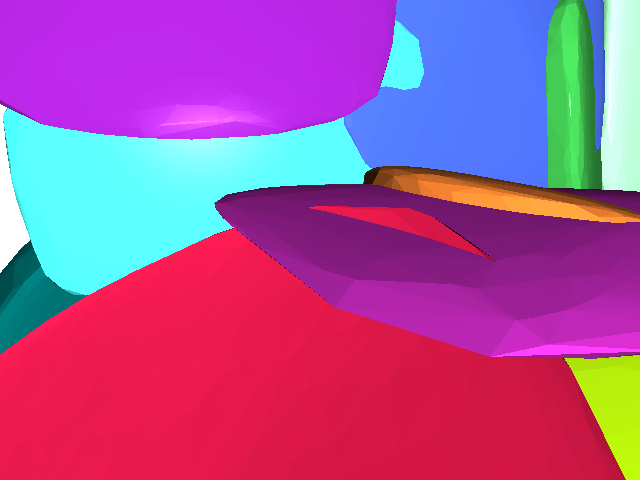} \\
   
                \parbox[t]{3mm}{{\rotatebox[origin=l]{90}{\hspace{1em} Depth}}} &
			\includegraphics[width=0.136\linewidth]{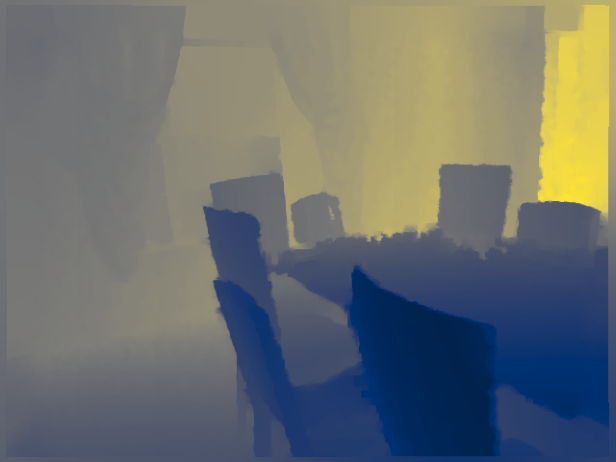} &
			\includegraphics[width=0.136\linewidth]{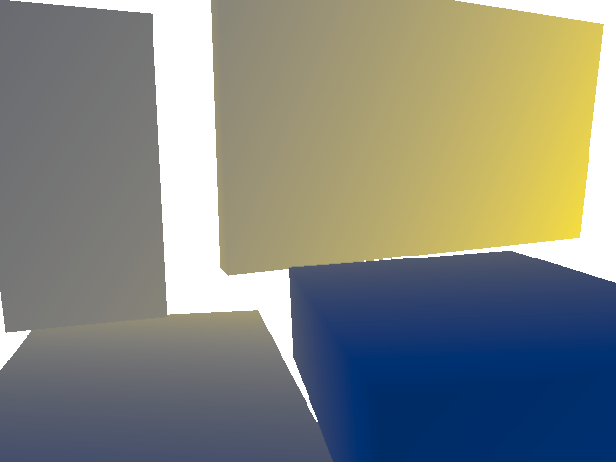} &
			\includegraphics[width=0.136\linewidth]{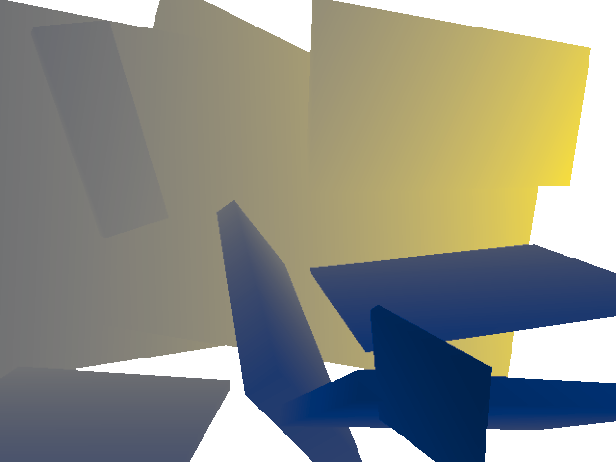} &
			\includegraphics[width=0.136\linewidth]{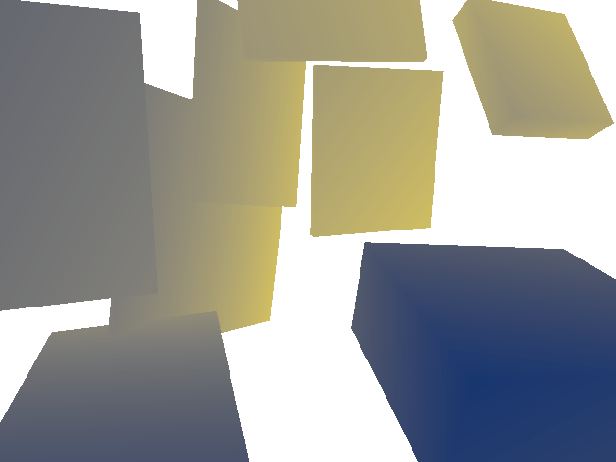} &
			\includegraphics[width=0.136\linewidth]{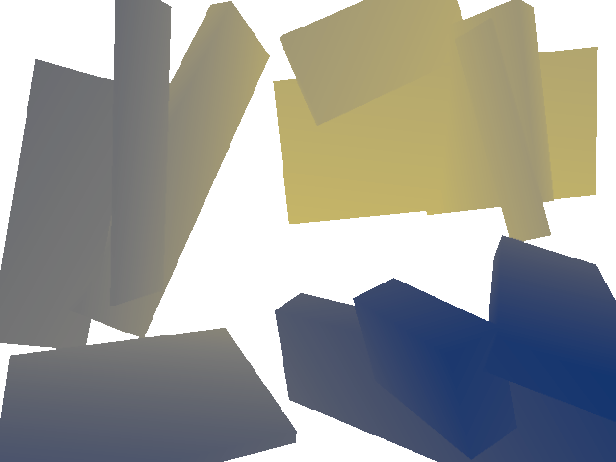} &
			\includegraphics[width=0.136\linewidth]{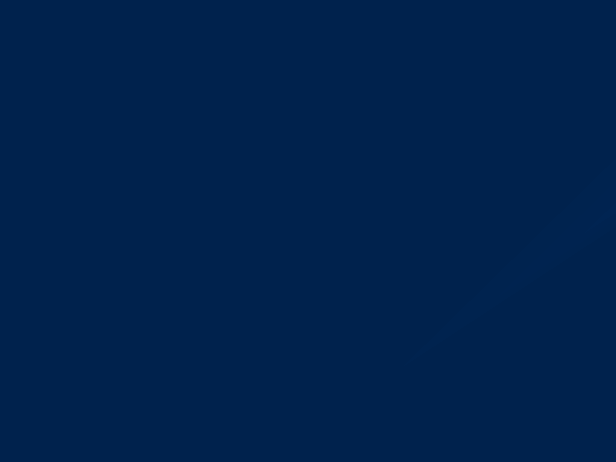} &
			\includegraphics[width=0.136\linewidth]{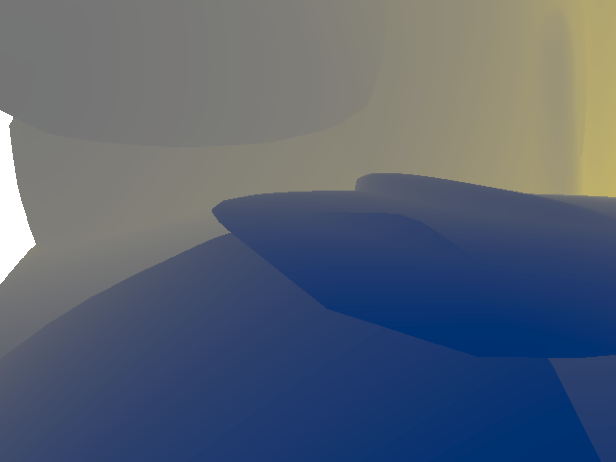} \\

                 &
			\includegraphics[width=0.136\linewidth]{fig/examples_new/rgb/final_0224_image.png} &
			\includegraphics[width=0.136\linewidth]{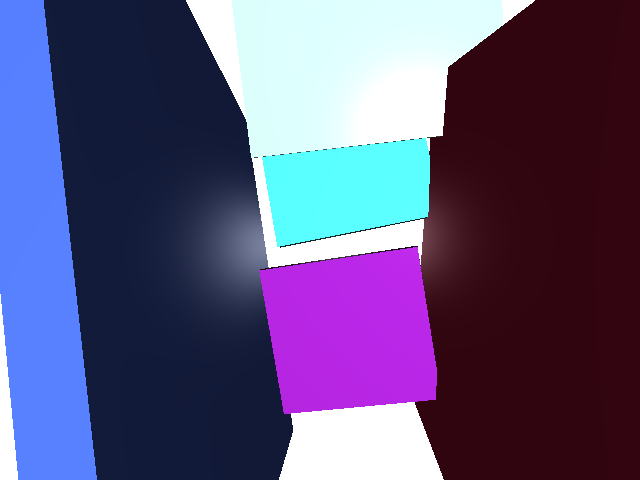} &
			\includegraphics[width=0.136\linewidth]{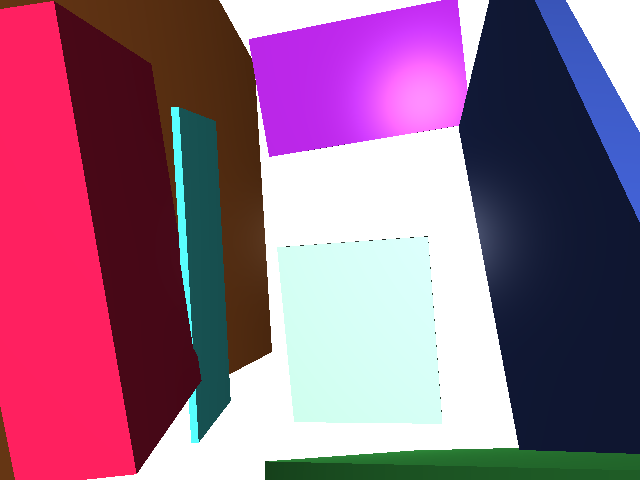} &
			\includegraphics[width=0.136\linewidth]{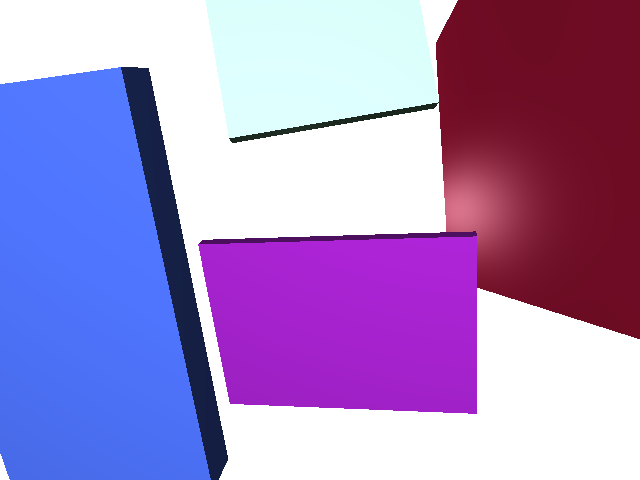} &
			\includegraphics[width=0.136\linewidth]{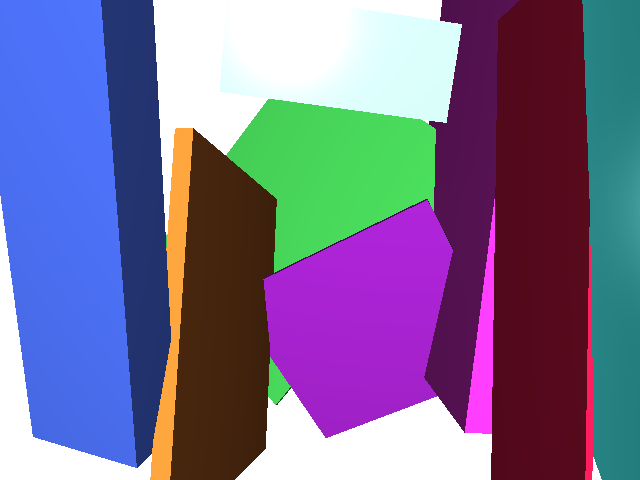} &
			\includegraphics[width=0.136\linewidth]{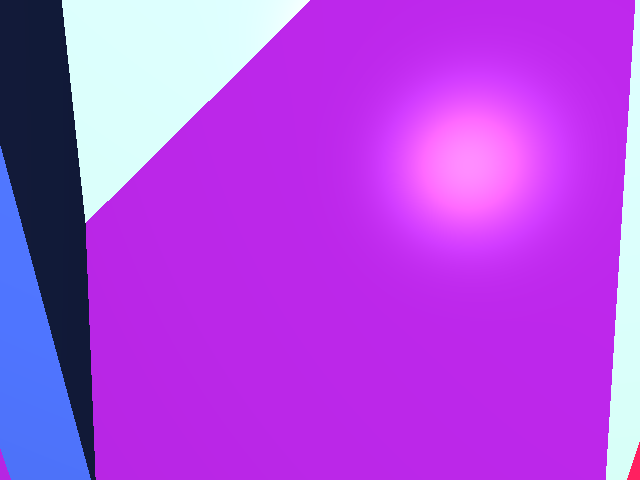} &
			\includegraphics[width=0.136\linewidth]{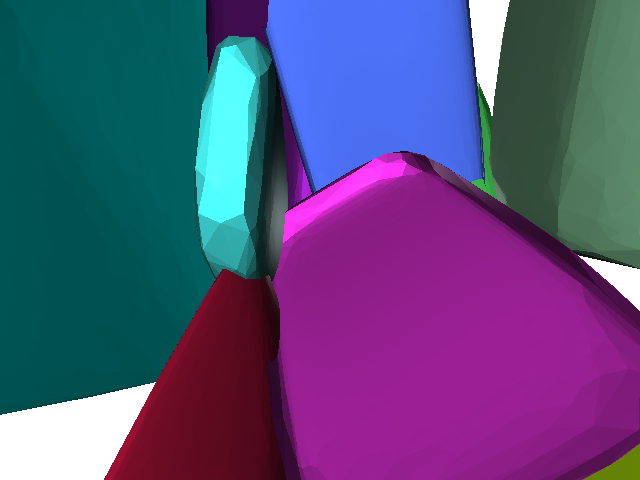} \\
   
                 &
			\includegraphics[width=0.136\linewidth]{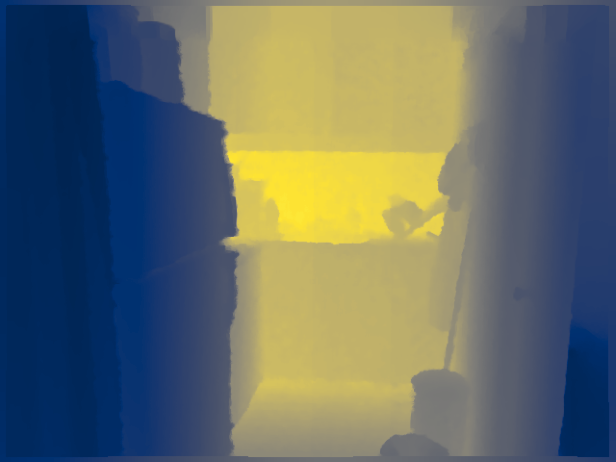} &
			\includegraphics[width=0.136\linewidth]{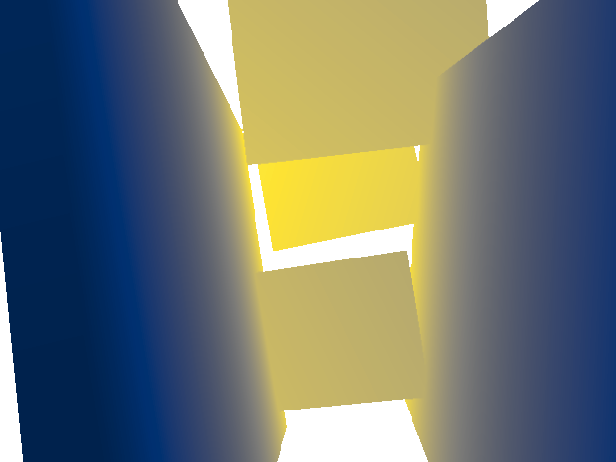} &
			\includegraphics[width=0.136\linewidth]{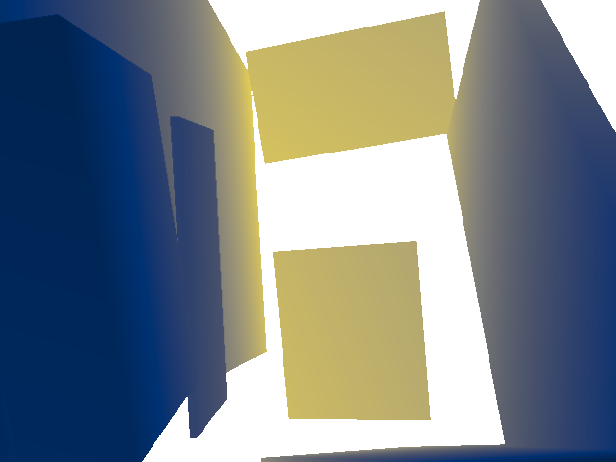} &
			\includegraphics[width=0.136\linewidth]{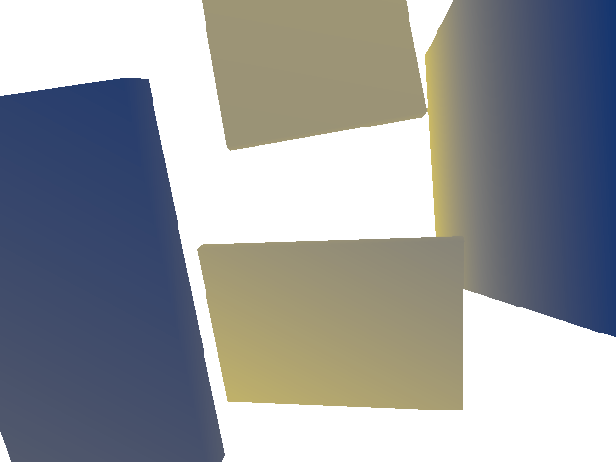} &
			\includegraphics[width=0.136\linewidth]{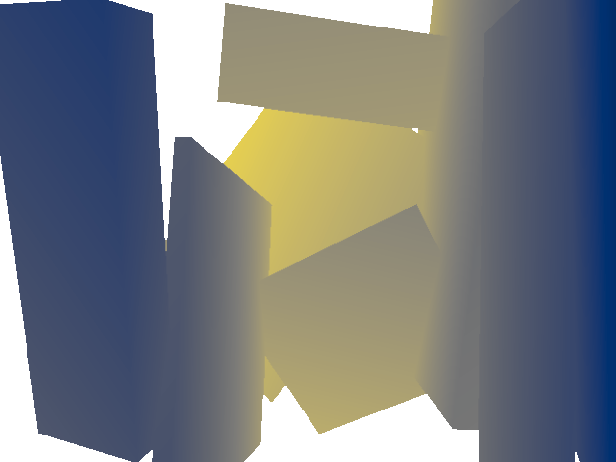} &
			\includegraphics[width=0.136\linewidth]{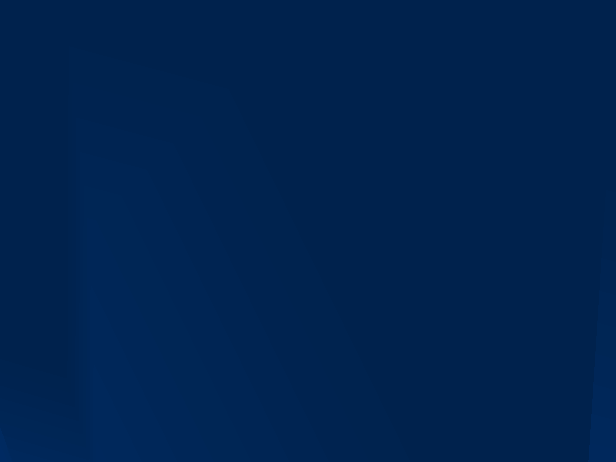} &
			\includegraphics[width=0.136\linewidth]{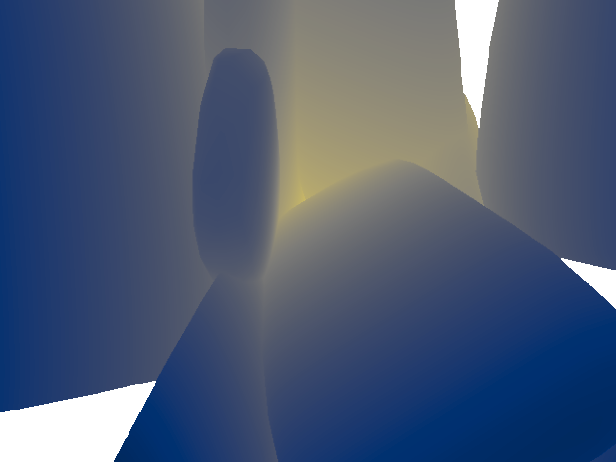} \\

                &
			\includegraphics[width=0.136\linewidth]{fig/examples_new/rgb/final_0281_image.png} &
			\includegraphics[width=0.136\linewidth]{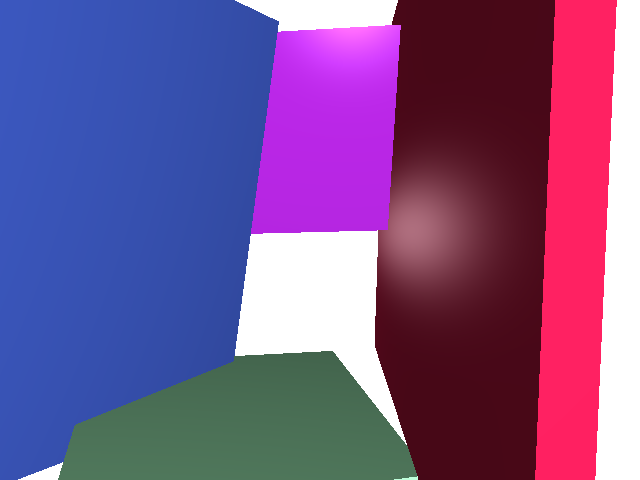} &
			\includegraphics[width=0.136\linewidth]{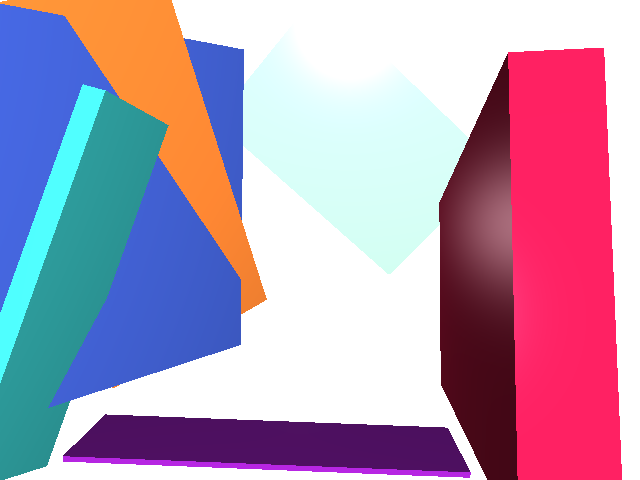} &
			\includegraphics[width=0.136\linewidth]{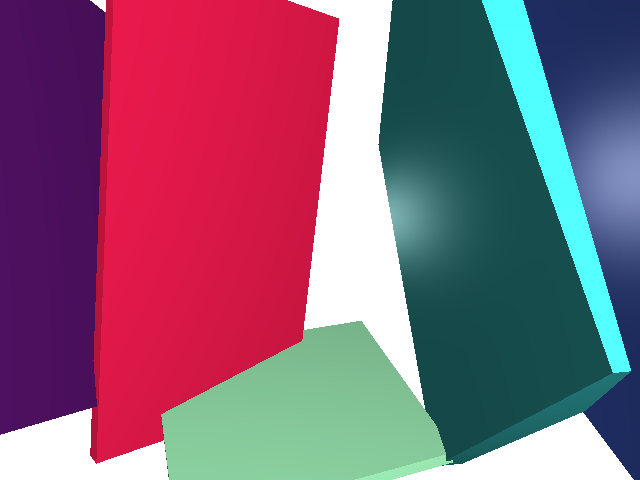} &
			\includegraphics[width=0.136\linewidth]{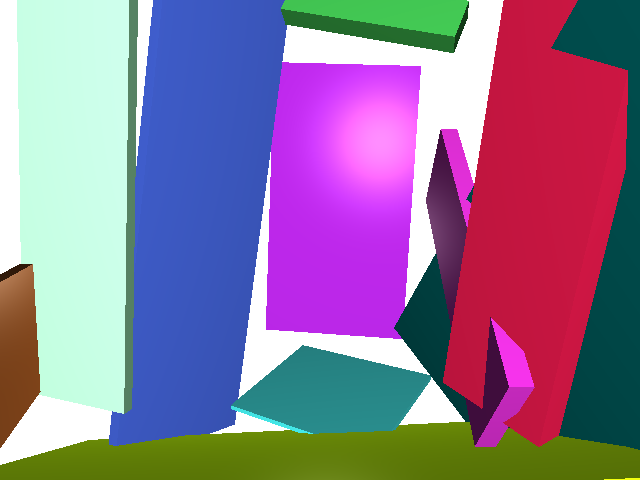} &
			\includegraphics[width=0.136\linewidth]{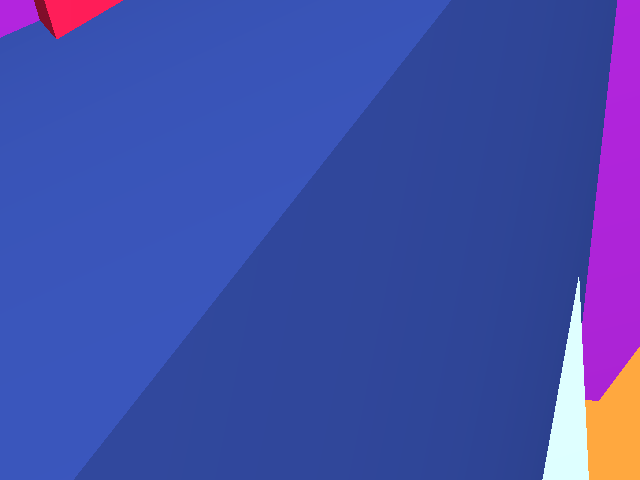} &
			\includegraphics[width=0.136\linewidth]{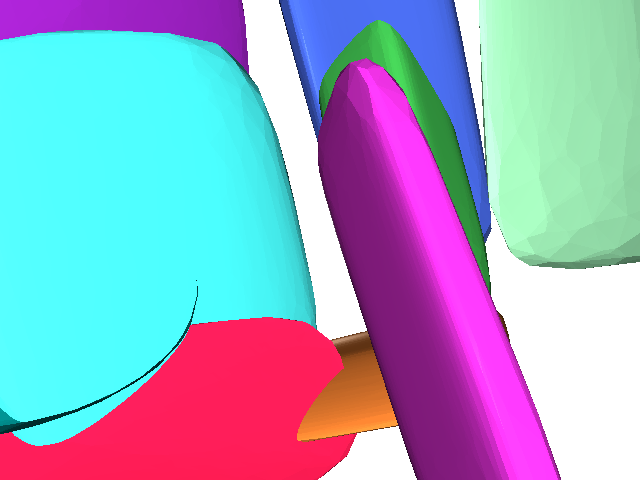} \\
   
                &
			\includegraphics[width=0.136\linewidth]{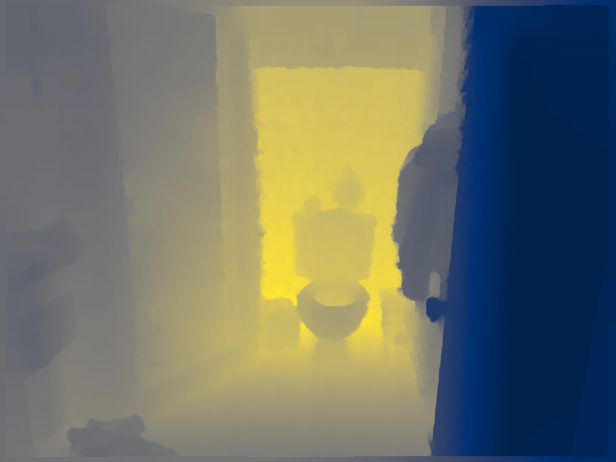} &
			\includegraphics[width=0.136\linewidth]{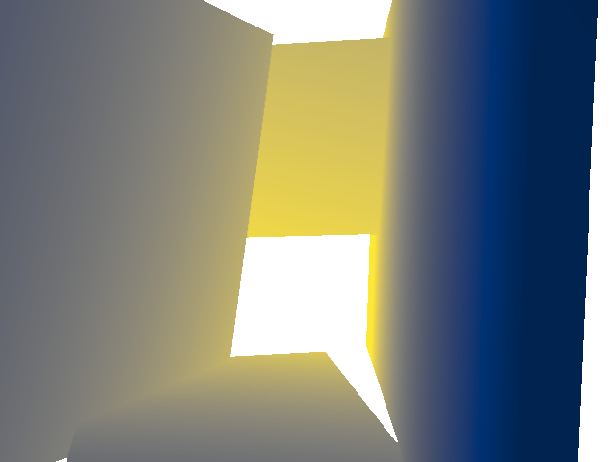} &
			\includegraphics[width=0.136\linewidth]{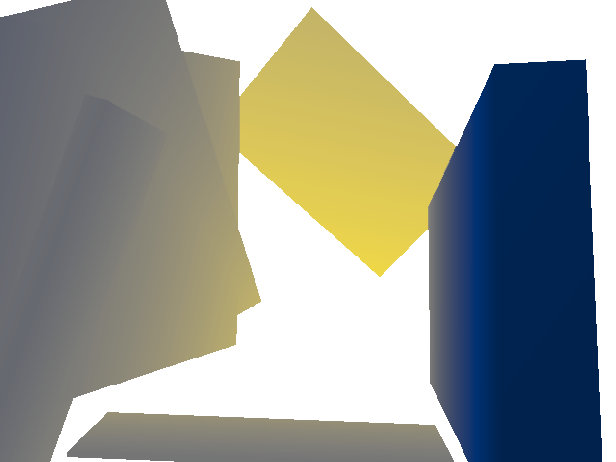} &
			\includegraphics[width=0.136\linewidth]{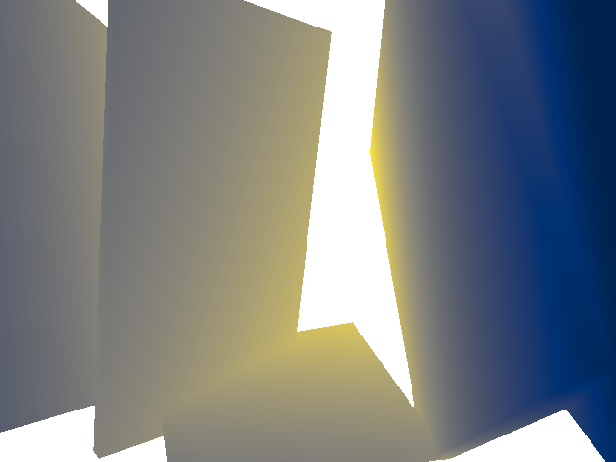} &
			\includegraphics[width=0.136\linewidth]{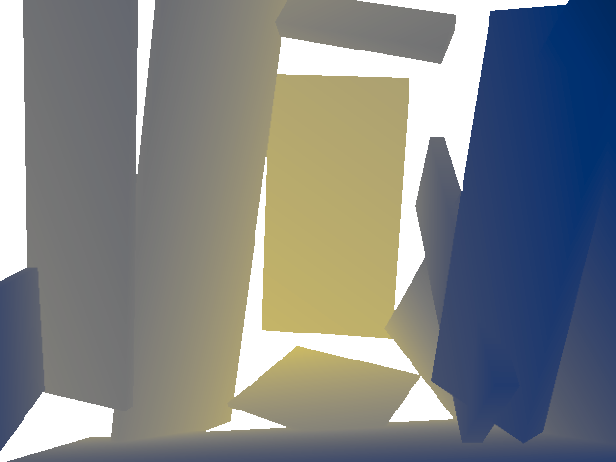} &
			\includegraphics[width=0.136\linewidth]{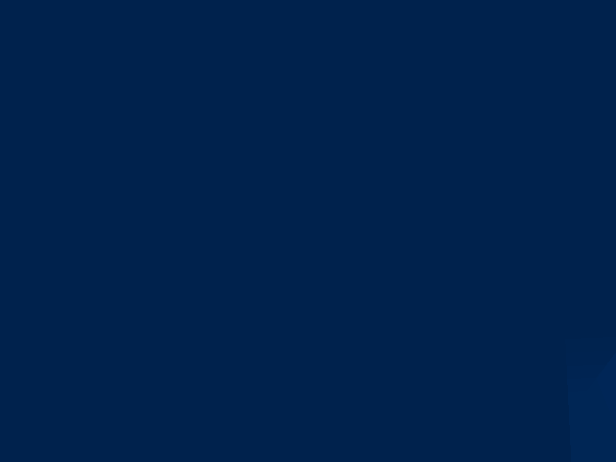} &
			\includegraphics[width=0.136\linewidth]{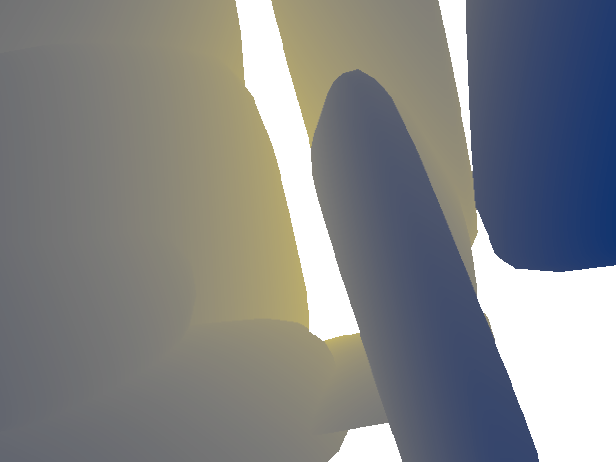} \\

                &
			\includegraphics[width=0.136\linewidth]{fig/examples_new/rgb/final_0422_image.png} &
			\includegraphics[width=0.136\linewidth]{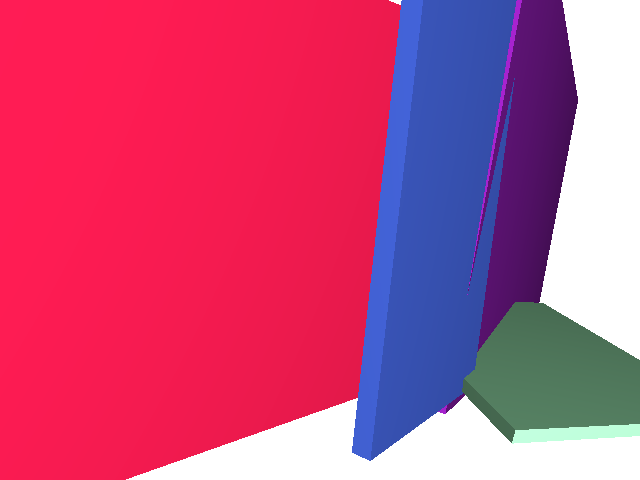} &
			\includegraphics[width=0.136\linewidth]{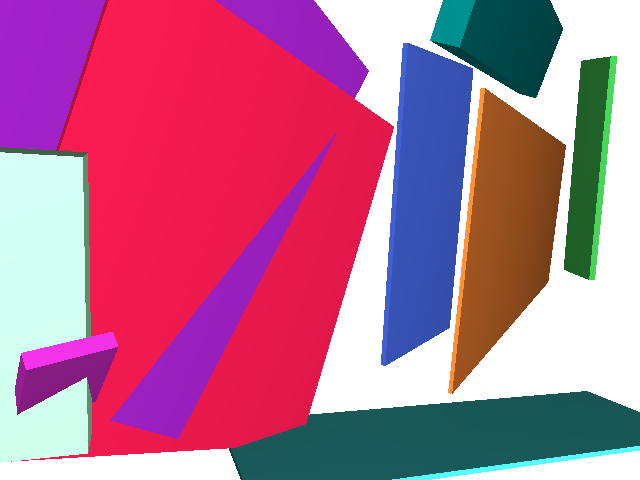} &
			\includegraphics[width=0.136\linewidth]{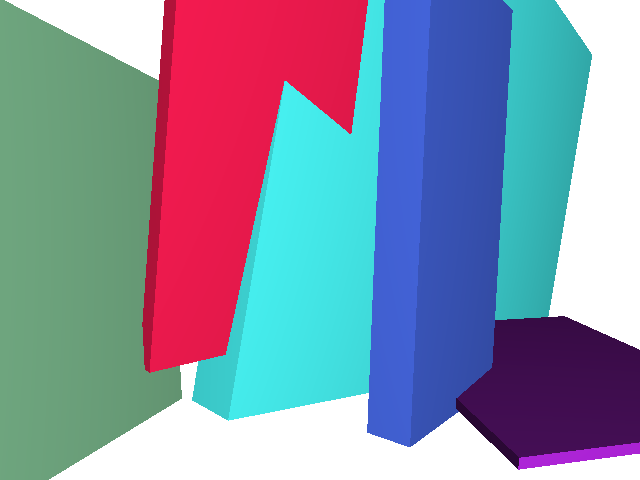} &
			\includegraphics[width=0.136\linewidth]{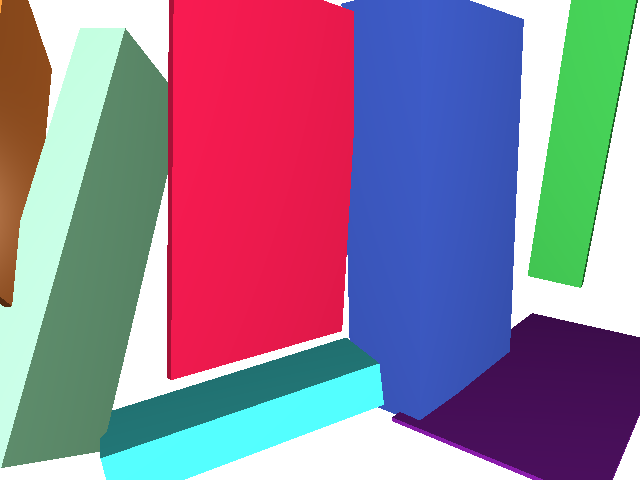} &
			\includegraphics[width=0.136\linewidth]{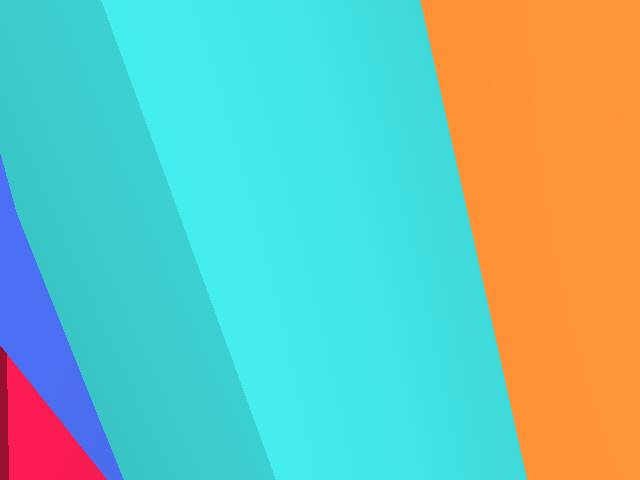} &
			\includegraphics[width=0.136\linewidth]{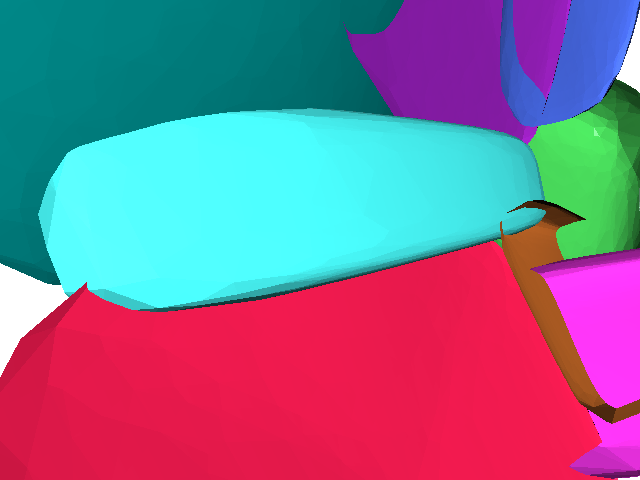} \\
   
                &
			\includegraphics[width=0.136\linewidth]{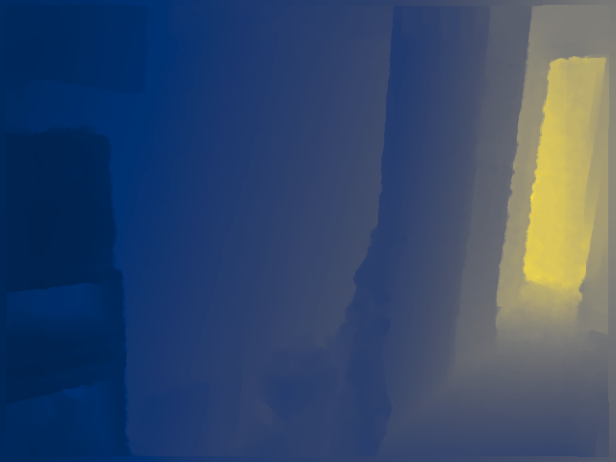} &
			\includegraphics[width=0.136\linewidth]{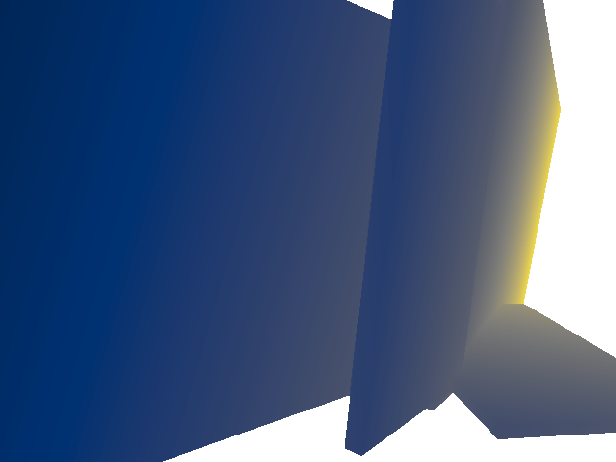} &
			\includegraphics[width=0.136\linewidth]{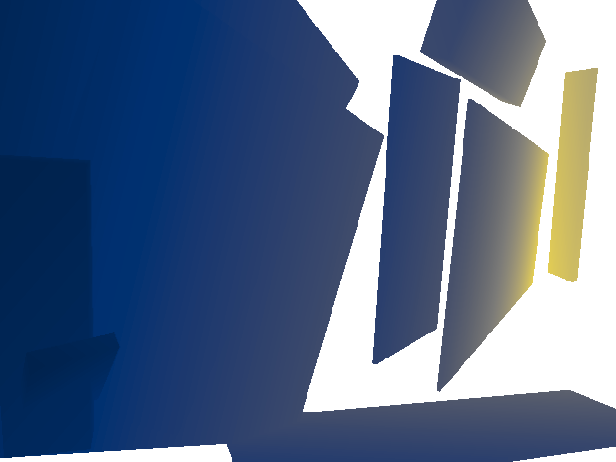} &
			\includegraphics[width=0.136\linewidth]{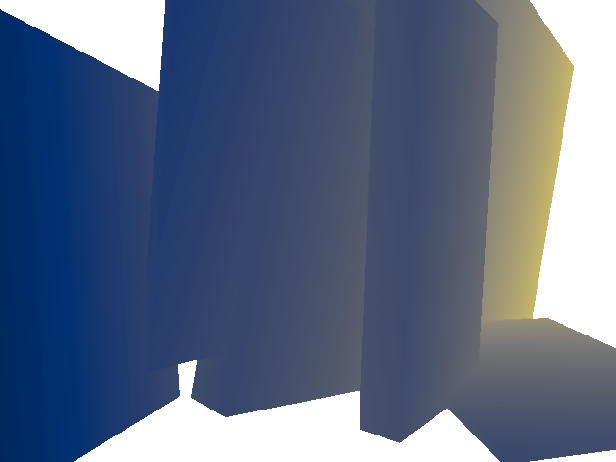} &
			\includegraphics[width=0.136\linewidth]{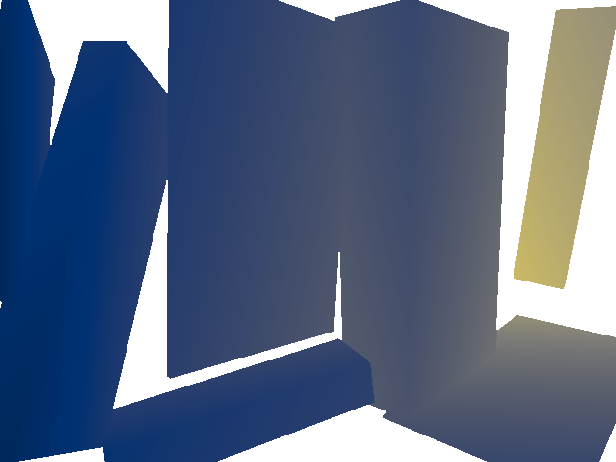} &
			\includegraphics[width=0.136\linewidth]{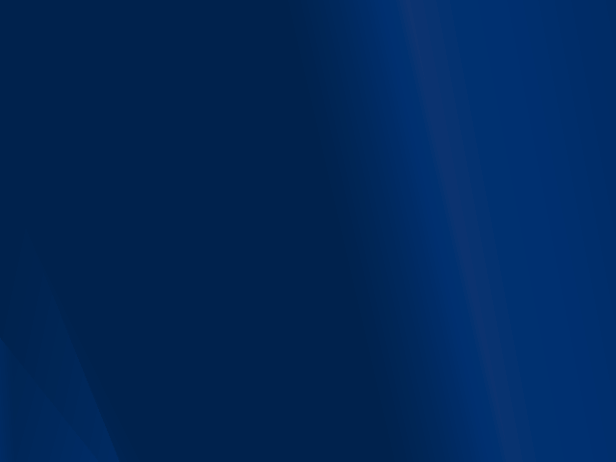} &
			\includegraphics[width=0.136\linewidth]{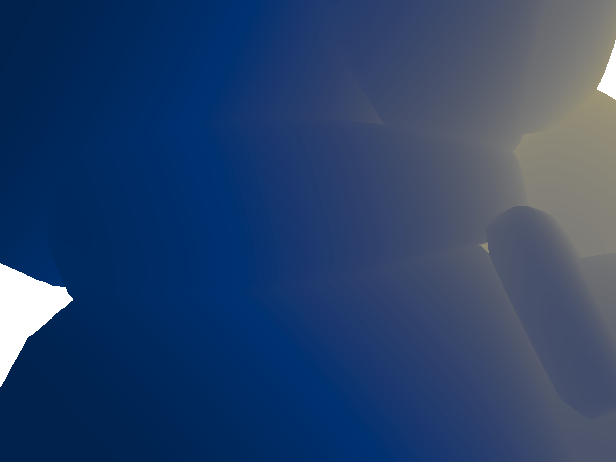} \\

                &
   			\includegraphics[width=0.136\linewidth]{fig/examples_new/rgb/final_0297_image.png} &
			\includegraphics[width=0.136\linewidth]{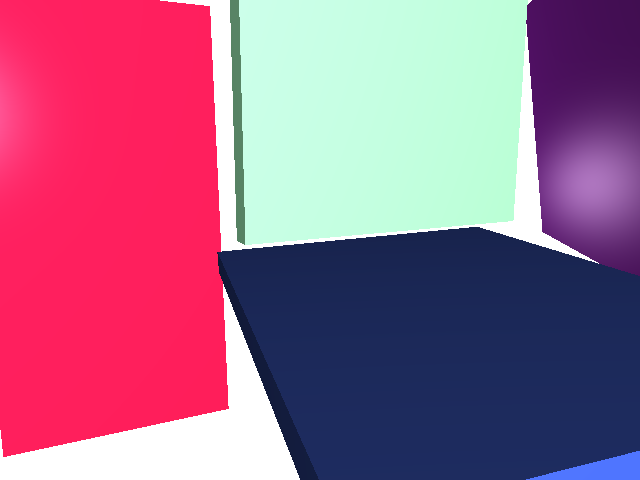} &
			\includegraphics[width=0.136\linewidth]{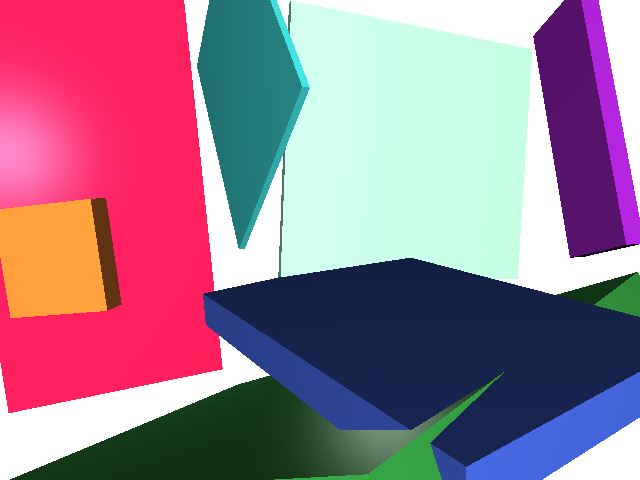} &
			\includegraphics[width=0.136\linewidth]{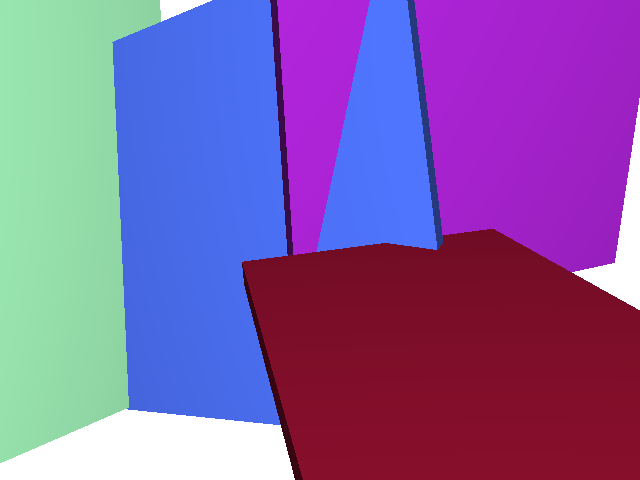} &
			\includegraphics[width=0.136\linewidth]{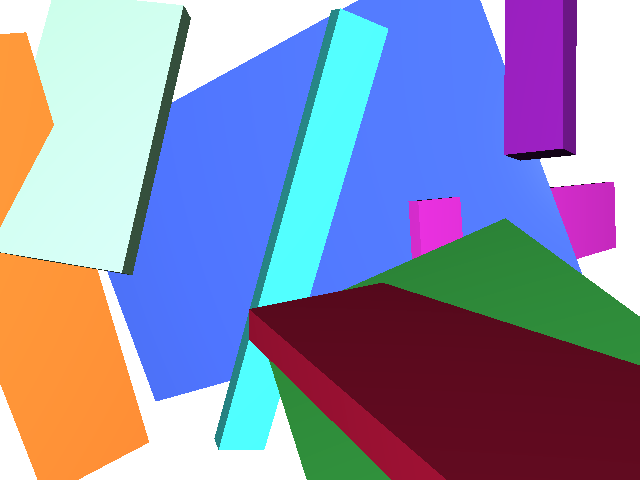} &
			\includegraphics[width=0.136\linewidth]{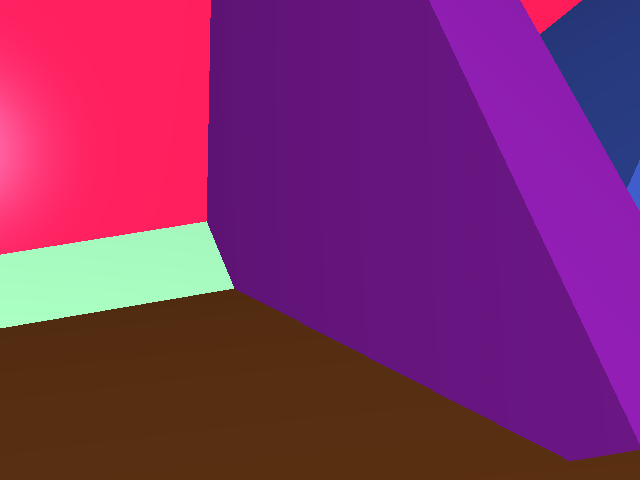} &
			\includegraphics[width=0.136\linewidth]{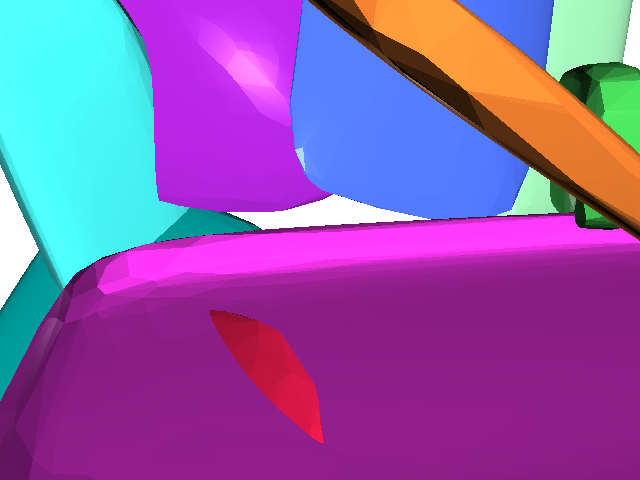} \\
   
                &
   			\includegraphics[width=0.136\linewidth]{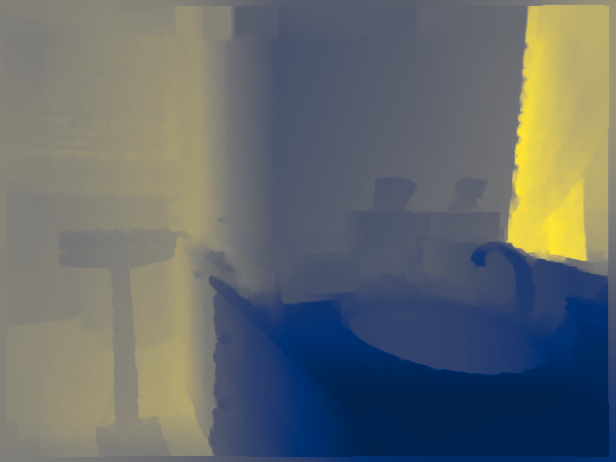} &
			\includegraphics[width=0.136\linewidth]{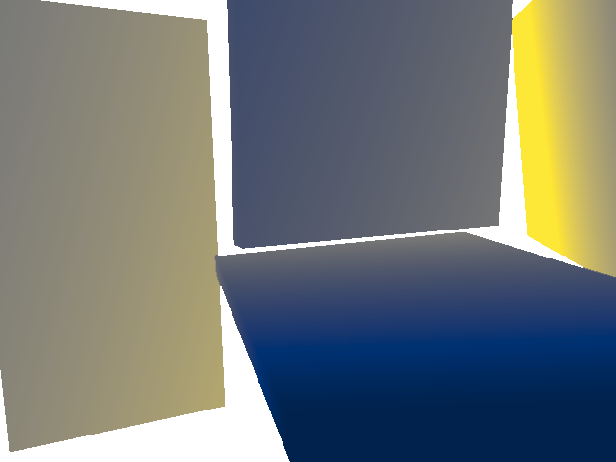} &
			\includegraphics[width=0.136\linewidth]{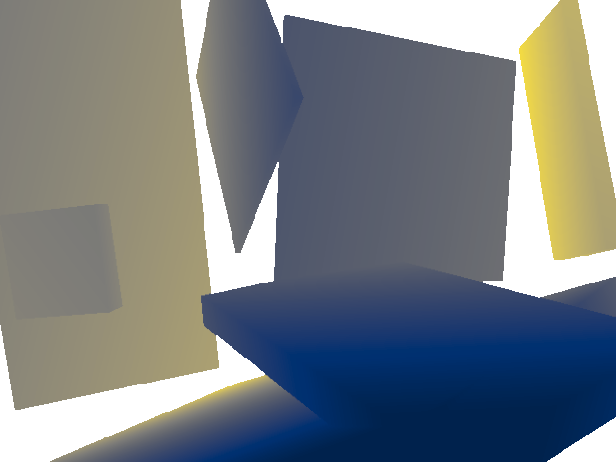} &
			\includegraphics[width=0.136\linewidth]{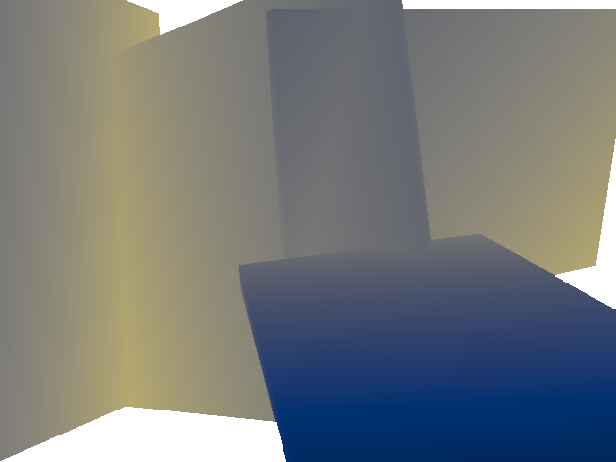} &
			\includegraphics[width=0.136\linewidth]{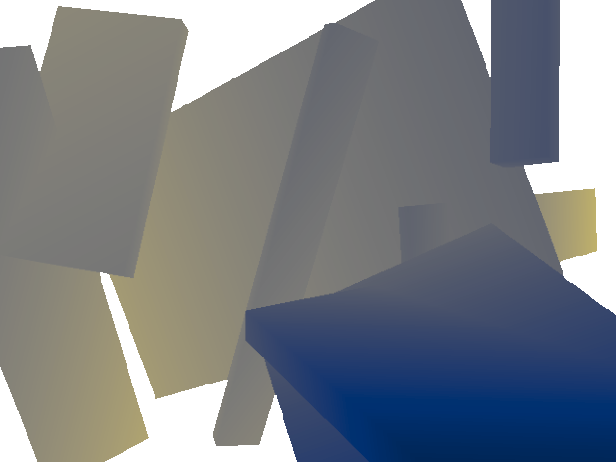} &
			\includegraphics[width=0.136\linewidth]{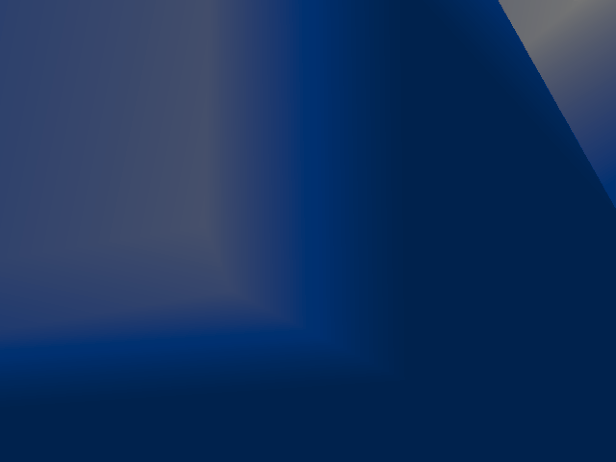} &
			\includegraphics[width=0.136\linewidth]{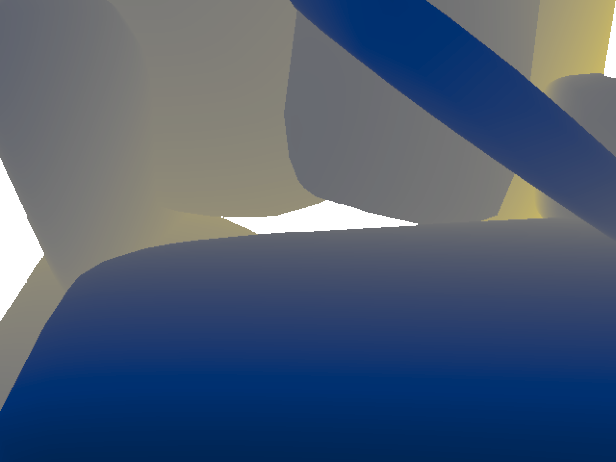} \\
        \end{tabular}
     
		\caption{\textbf{NYU: Qualitative Results.}
		\new{
		First column: Input RGB images. 
		Columns 2-5: Cuboids obtained with our proposed methods, using either ground truth depth or RGB images as input. 
		Column 6: Cuboids obtained using CONSAC~\cite{KluBra2020} adapted for cuboid fitting.
		Column 7: Superquadrics obtained with SQ-Parsing~\cite{paschalidou2019superquadrics}, trained on NYU~\cite{silberman2012indoor}.
		For our methods and for CONSAC, the colours convey the order in which the cuboids have been selected: \textcolor{c_red}{red}, \textcolor{c_blue}{blue}, \textcolor{c_green}{green}, \textcolor{c_purple}{purple}, \textcolor{c_cyan}{cyan}, \textcolor{c_orange}{orange}.
		}}
		\label{fig:examples}
  \end{center}
\end{figure*}   

In Figs.~\ref{fig:smh_examples} and~\ref{fig:examples}, we provide additional qualitative results on the Synthetic Metropolis Homographies and NYU Depth v2 datasets.
}

\section{Cuboid Parsing Baseline}
\label{sec:tulsiani}
\begin{figure*}	
		\centering
		\begin{subfigure}[t]{0.99\linewidth}
			\centering
			\includegraphics[width=0.16\linewidth]{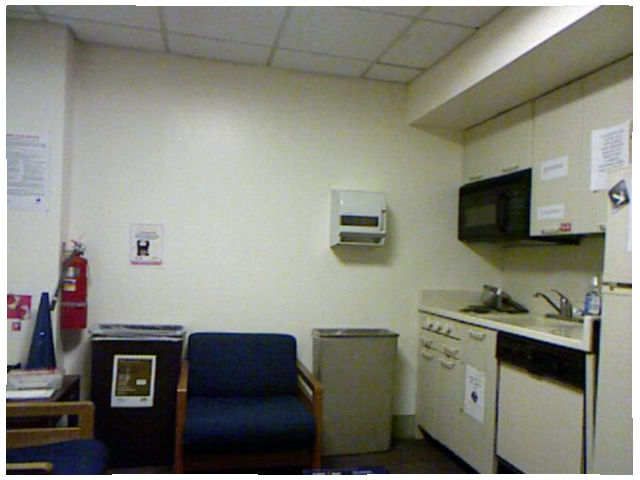}
			\includegraphics[width=0.16\linewidth]{fig/volprim/0010.jpg}
			\includegraphics[width=0.16\linewidth]{fig/volprim/0020.jpg}
			\includegraphics[width=0.16\linewidth]{fig/volprim/0836.jpg}
			\includegraphics[width=0.16\linewidth]{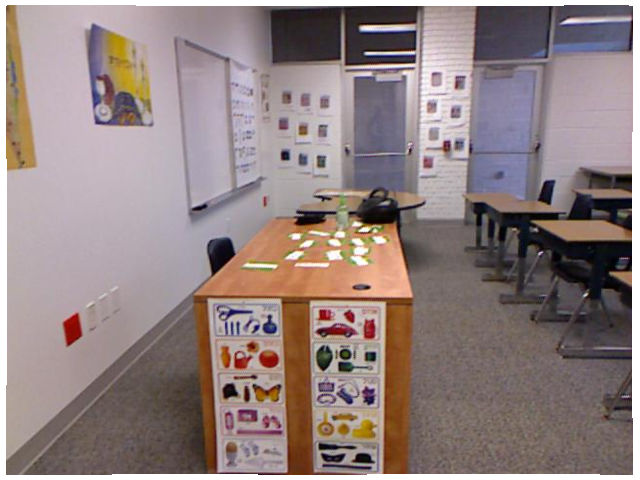}
			\includegraphics[width=0.16\linewidth]{fig/volprim/0570.jpg}
			\caption{Original images}
		\end{subfigure}
		\begin{subfigure}[t]{0.99\linewidth}
			\centering
			\includegraphics[width=0.16\linewidth]{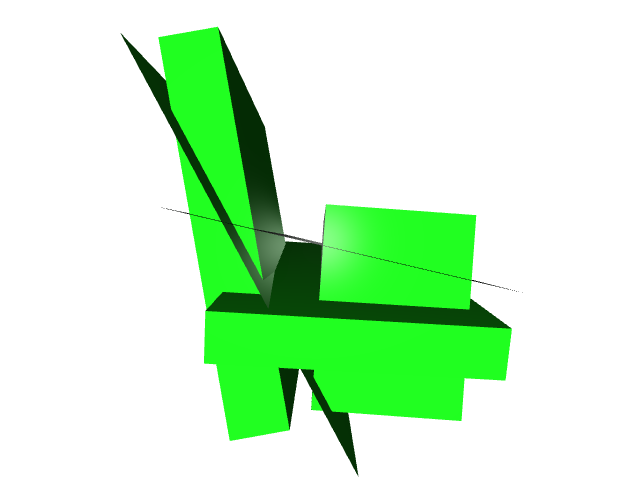}
			\includegraphics[width=0.16\linewidth]{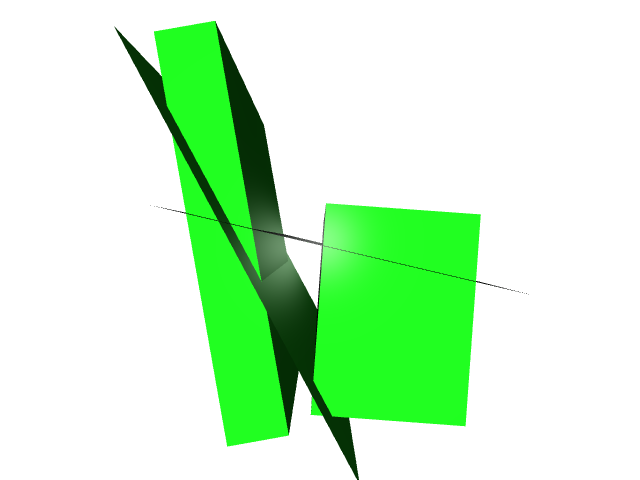}
			\includegraphics[width=0.16\linewidth]{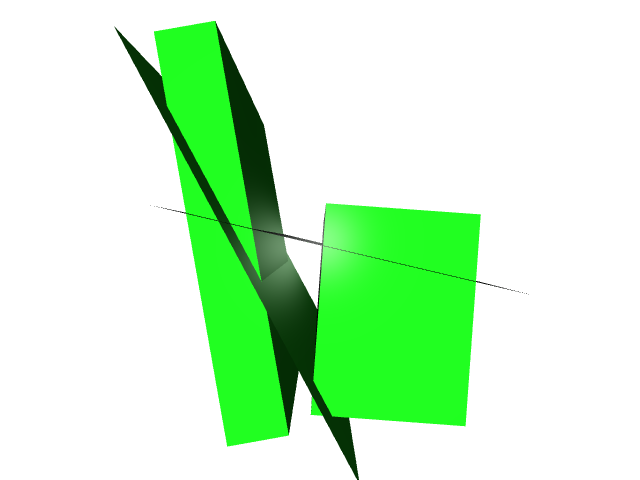}
			\includegraphics[width=0.16\linewidth]{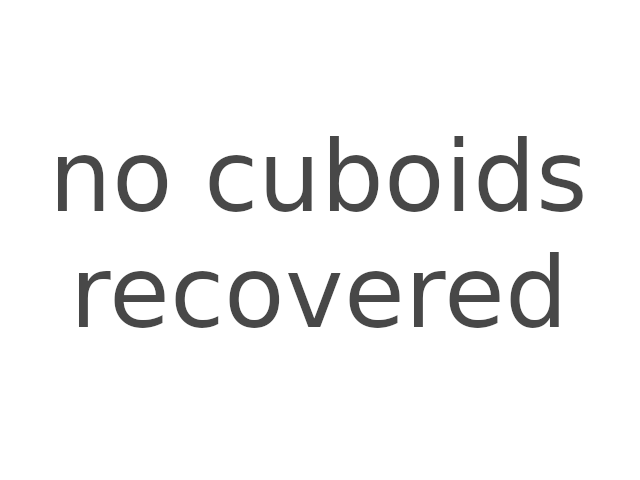}
			\includegraphics[width=0.16\linewidth]{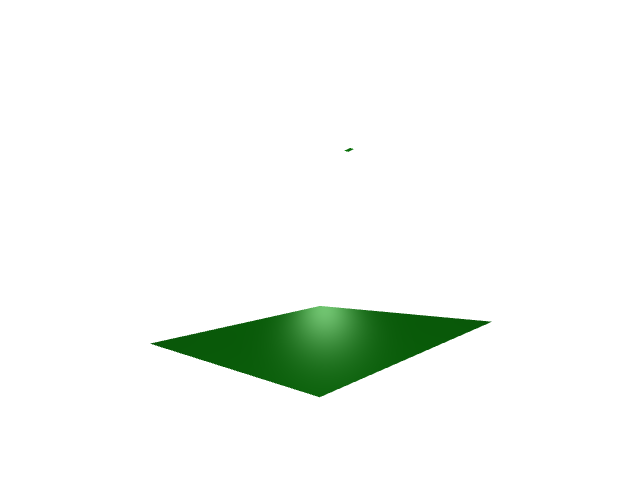}
			\includegraphics[width=0.16\linewidth]{fig/volprim/no_cuboids.png}
			\caption{Recovered cuboids using~\cite{tulsiani2017learning}, first training run}
		\end{subfigure}
		\begin{subfigure}[t]{0.99\linewidth}
			\centering
			\includegraphics[width=0.16\linewidth]{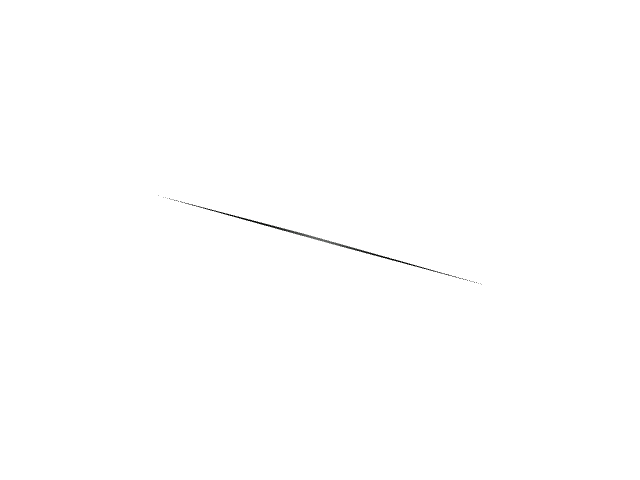}
			\includegraphics[width=0.16\linewidth]{fig/volprim/no_cuboids.png}
			\includegraphics[width=0.16\linewidth]{fig/volprim/no_cuboids.png}
			\includegraphics[width=0.16\linewidth]{fig/volprim/no_cuboids.png}
			\includegraphics[width=0.16\linewidth]{fig/volprim/no_cuboids.png}
			\includegraphics[width=0.16\linewidth]{fig/volprim/no_cuboids.png}
			\caption{Recovered cuboids using~\cite{tulsiani2017learning}, second training run}
		\end{subfigure}
		\begin{subfigure}[t]{0.99\linewidth}
			\centering
			\includegraphics[width=0.16\linewidth]{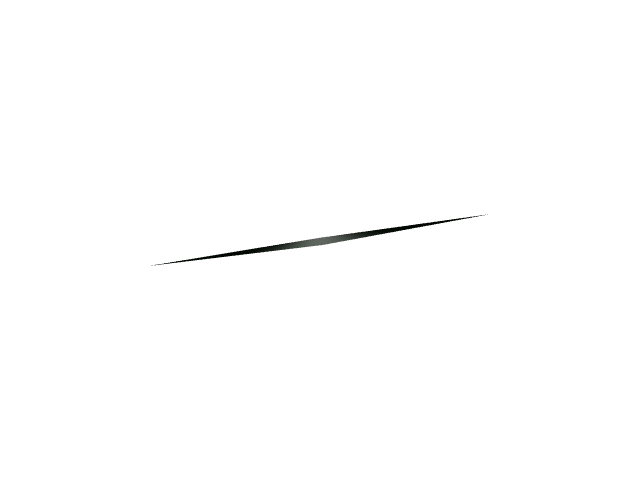}
			\includegraphics[width=0.16\linewidth]{fig/volprim/no_cuboids.png}
			\includegraphics[width=0.16\linewidth]{fig/volprim/no_cuboids.png}
			\includegraphics[width=0.16\linewidth]{fig/volprim/no_cuboids.png}
			\includegraphics[width=0.16\linewidth]{fig/volprim/no_cuboids.png}
			\includegraphics[width=0.16\linewidth]{fig/volprim/no_cuboids.png}
            \caption{Recovered cuboids using~\cite{tulsiani2017learning}, third training run}
		\end{subfigure}
		
		\caption{\textbf{Cuboid Parsing Baseline:} We show qualitative results for the cuboid based approach of~\cite{tulsiani2017learning}. Row (a) shows the original images from the NYU dataset. Rows (b)-(d) show corresponding cuboid predictions by~\cite{tulsiani2017learning} for three different training runs using ground truth depth as input. As these examples show, the method is unable to recover sensible cuboid configurations for these real-world indoor scenes. See Sec.~\ref{sec:tulsiani} for details. }
		\label{fig:volprim_examples}
\end{figure*}

As mentioned in the main paper, we were unable to obtain sensible results with the cuboid based approach of~\cite{tulsiani2017learning} for the NYU dataset.
We trained their approach on NYU using ground truth depth input, following their instructions published together with their source code, using the same input data as we did for the superquadrics approach of~\cite{paschalidou2019superquadrics}, multiple times with different random seeds, but to no avail.
Their approach mostly predicts the same or very similar cuboid configurations for different images.
Often, no cuboids are recovered at all.
We show a couple of examples for three different training runs in Fig.~\ref{fig:volprim_examples}.

\end{appendices}

\bibliography{paper}{}
\bibliographystyle{IEEEtran}

\end{document}